%% file: main.tex
\pdfoutput=1
\documentclass{article}

\usepackage{microtype}
\usepackage{graphicx}
\usepackage{subcaption}
\usepackage{booktabs}
\usepackage{array}
\usepackage{multirow}
\usepackage{hyperref}

\usepackage[preprint]{icml2026}

\usepackage{amsmath}
\usepackage{amssymb}
\usepackage{mathtools}
\usepackage{amsthm}
\usepackage{bm}
\usepackage{tikz}
\usetikzlibrary{decorations.pathreplacing,arrows.meta,positioning}
\usepackage{enumitem}
\allowdisplaybreaks
\usepackage[capitalize,noabbrev]{cleveref}

\theoremstyle{plain}
\newtheorem{theorem}{Theorem}[section]
\newtheorem{proposition}[theorem]{Proposition}
\newtheorem{lemma}[theorem]{Lemma}

\theoremstyle{definition}
\newtheorem{definition}[theorem]{Definition}

\theoremstyle{remark}
\newtheorem{remark}{Remark}[section]

\input{macros}

\icmltitlerunning{Improved Regret in Adversarial MDPs via Model-Free Self-Normalized PO}

\begin{document}

\twocolumn[
  \icmltitle{Improved Regret in Adversarial MDPs\\via Model-Free Self-Normalized Policy Optimization}

  \icmlsetsymbol{equal}{*}

  \begin{icmlauthorlist}
    \icmlauthor{Orin Levy}{yyy,comp,equal}
    \icmlauthor{Aviv Rosenberg}{comp,equal}
    \icmlauthor{Alon Cohen}{comp,sch}
    \icmlauthor{Yishay Mansour}{yyy,comp}
  \end{icmlauthorlist}

  \icmlaffiliation{yyy}{School of Computer Science and AI, Tel Aviv University}
  \icmlaffiliation{comp}{Google Research}
  \icmlaffiliation{sch}{School of Electrical Engineering, Tel Aviv University}

  \icmlcorrespondingauthor{Orin Levy}{orinlevy96@gmail.com}
  \icmlcorrespondingauthor{Aviv Rosenberg}{avivros007@gmail.com}
  \icmlcorrespondingauthor{Alon Cohen}{alonco@tauex.tau.ac.il}
  \icmlcorrespondingauthor{Yishay Mansour}{mansour.yishay@gmail.com}

  \icmlkeywords{Reinforcement Learning, Adversarial MDPs, Policy Optimization, Bandit Feedback, Linear MDPs, ICML}

  \vskip 0.3in
]

\printAffiliationsAndNotice{\icmlEqualContribution}

\begin{abstract}
We study episodic adversarial Markov decision processes (AMDPs) with unknown transitions and bandit feedback. For tabular AMDPs with $S$ states, $A$ actions, horizon $H$ and $K$ episodes, a regret bound with the optimal $\sqrt S$ dependence was obtained only very recently, by an algorithm that uses transition confidence sets. We introduce \alg{}, a simpler, model-free algorithm with better bounds. Each state keeps its action distribution fixed over a block of episodes whose length is inversely proportional to its visitation frequency, and then divides its accumulated returns by the number of visits it received. 
%This self-normalized estimate is unbiased, up to a negligible term, when the other states do not change their policies, and otherwise has only a small relative bias, which a mild discount of the returns absorbs. 
\alg{} has regret $\tO(H^{7/4}\sqrt{SAK})$ with high probability, improving the previous $\tO(H^{5/2}\sqrt{SAK}+H^4S^2A)$ in the horizon and removing the additive term, with $O(SA)$ memory and $O(HA)$ amortized runtime per episode. \alg{} also gives the first bound with $\sqrt S$ dependence under aggregate feedback with unknown transitions, $\tO(H^{11/4}\sqrt{SAK})$,
%and,with Gram matrices in place of visit counts, 
and when applied to $d$-dimensional linear AMDPs, it improves the best bound for efficient algorithms to $\tO((d^{3/2}H^{11/4}+d^{7/4}H^{9/4})\sqrt{K})$.
\end{abstract}

\input{sections/1_intro}
\input{sections/2_setting}
\input{sections/3_algorithm}
\input{sections/4_analysis}
\input{sections/5_aggregate}
\input{sections/6_linear}

\input{sections/7_related_discussion}

%\clearpage
\section*{Impact Statement}
This paper presents theoretical work whose goal is to advance the foundations of Reinforcement Learning and Machine Learning. There are many potential societal consequences of our work, none of which we feel must be specifically highlighted here.

\section*{Acknowledgment}
This project is supported by the European Research Council (ERC) under the European Union’s
Horizon 2020 research and innovation program (grant agreement No. 882396), by the Israel Science
Foundation (1357/24), the Blavatnik foundation at Tel Aviv University and by a grant from the
Tel Aviv University Center for AI and Data Science (TAD). 
OL is also supported by the Google PhD Fellowship Award
(2025). AC is supported by
the Israel Science Foundation (ISF, grant number 2250/22).

\bibliography{refs}
\bibliographystyle{icml2026}

\newpage
\appendix
\onecolumn
\crefalias{section}{appendix}
\crefalias{subsection}{subappendix}

\input{appendix/0_overview}

\input{appendix/N_related}
\input{appendix/A_notation}
\input{appendix/B_identity}
\input{appendix/C_omd}
\input{appendix/D_drift}
\input{appendix/E_selfnorm}
\input{appendix/F_classification}
\input{appendix/G_events}
\input{appendix/H_local}
\input{appendix/I_main_proof}
\input{appendix/J_extensions}
\input{appendix/K_horizon}
\input{appendix/L_aggregate}
\input{appendix/M_linear_block}

\end{document}

%% file: macros.tex
\newcommand{\Sset}{\mathcal S}
\newcommand{\Aset}{\mathcal A}
\newcommand{\F}{\mathcal F}
\newcommand{\G}{\mathcal G}
\newcommand{\Fagg}{\mathcal F^{\mathrm{agg}}}
\newcommand{\Gagg}{\mathcal G^{\mathrm{agg}}}
\newcommand{\Omagg}{\Omega_{\mathrm{agg}}}

\newcommand{\cC}{\mathcal C}

\newcommand{\cK}{\mathcal K}

\newcommand{\E}{\mathbb E}
\newcommand{\Prob}{\mathbb P}
\newcommand{\R}{\mathbb R}

\newcommand{\ind}{\mathbf 1}
\newcommand{\Lmax}{L_{\max}}
\newcommand{\Unif}{\operatorname{Unif}}
\newcommand{\SN}{\mathrm{SN}}
\newcommand{\Ecl}{\mathcal E_{\mathrm{cl}}}
\newcommand{\good}{\mathcal E}
\newcommand{\Eclagg}{\mathcal E_{\mathrm{cl}}^{\mathrm{agg}}}
\newcommand{\Eagg}{\mathcal E_{\mathrm{agg}}}
\newcommand{\etail}{\varepsilon_{\mathrm{tail}}}
\newcommand{\ip}[2]{\langle #1,#2\rangle}
\newcommand{\KL}{\operatorname{KL}}
\newcommand{\Reg}{\mathrm{Reg}_K}
\newcommand{\aeta}{\mathsf a_\eta}
\newcommand{\beeta}{\mathsf b_\eta}
\newcommand{\agam}{\mathsf a_\gamma}
\newcommand{\bgam}{\mathsf b_\gamma}
\newcommand{\taeta}{\tilde{\mathsf a}_\eta}
\newcommand{\tbeeta}{\tilde{\mathsf b}_\eta}
\newcommand{\tagam}{\tilde{\mathsf a}_\gamma}
\newcommand{\tbgam}{\tilde{\mathsf b}_\gamma}
\newcommand{\kagg}{\kappa_{\mathrm{agg}}}
\newcommand{\Rb}{\mathsf R}
\newcommand{\pre}{\mathrm{pre}}
\newcommand{\lrn}{\mathrm{lrn}}
\newcommand{\cmp}{\mathrm{cmp}}
\newcommand{\Var}{\operatorname{Var}}
\DeclareMathOperator*{\argmin}{arg\,min}

\DeclareMathOperator{\poly}{poly}
\newcommand{\tO}{\widetilde O}
\newcommand{\tTheta}{\widetilde\Theta}
\newcommand{\tOmega}{\widetilde\Omega}

\newcommand{\what}{\widehat w}
\newcommand{\Qhat}{\widehat Q}

\newcommand{\Qtil}{\widetilde Q}
\newcommand{\Sigmahat}{\widehat\Sigma}
\newcommand{\phat}{\widehat p}
\newcommand{\alg}{\textup{SN-PO}}
\newcommand{\aggalg}{\textup{SN-PO-Agg}}
\newcommand{\linblkalg}{\textup{SN-PO-Lin}}
\newcommand{\acc}{\mathsf{acc}}

%% file: sections/1_intro.tex
\section{Introduction}\label{sec:intro}

Adversarial Markov decision processes (AMDPs) model reinforcement learning problems in which the cost function may change arbitrarily between episodes while the transition kernel stays fixed \citep{even2009online,yu2009markov,neu2010online}. A learner interacts with the environment for $K$ episodes of $H$ steps and is compared with the best fixed policy in hindsight. We study the case of unknown transitions and \emph{bandit} feedback, where the learner observes only the costs of the state-action pairs it visits, with costs fixed in advance by an oblivious adversary.

%\paragraph{The $\sqrt S$ gap.}
For tabular AMDPs with $S$ states and $A$ actions, \citet{jin2020learning} proved the first $\sqrt K$ regret bound in this setting, $\tO(HS\sqrt{AK})$, and asked whether the $\sqrt S$ gap to the lower bound $\Omega(H\sqrt{SAK})$ can be closed. Later algorithms, including policy optimization methods \citep{luo2021policy,li2026closing}, kept the linear dependence on $S$. The regret decomposes into regrets of the individual states, each weighted by how often the \emph{comparator} visits the state, so a state has to account for all $K$ episodes although it observes a return only when the learner visits it. A return observed at $s$ under the current policy $\pi^k$ is therefore reweighted by the inverse of the probability $q^k(s)$ that $\pi^k$ reaches $s$, which depends on the unknown transitions. These algorithms replace $q^k(s)$ by an optimistic value computed from a confidence set of transition kernels, whose uncertainty causes the linear dependence on $S$.

\citet{kulkarni2026introducing} recently closed the gap in this setting. Their key insight is that a policy can be evaluated from \emph{future} episodes without freezing it, since freezing the whole policy while samples are collected worsens the dependence on $K$. Their algorithm attains $\tO(H^{5/2}\sqrt{SAK}+H^4S^2A)$ regret, and the same idea gives $\tO(d^3H^4\sqrt K)$ in linear AMDPs with $d$-dimensional features, where efficient algorithms had been limited to $K^{3/4}$ regret \citep{liu2024towards}. Independently, \citet{liu2026computationally} obtained $\sqrt K$ expected regret for linear AMDPs with an efficient algorithm that uses only the total cost of each episode.

%\paragraph{This paper.}
In this work, we simplify the freezing idea and prove that freezing \emph{locally} is sufficient: in our method, each state freezes only its own action distribution, over a block of episodes adapted to its visitation frequency, while the other states update on their own schedules. The resulting estimator, \emph{self-normalization over blocks}, is new to AMDPs (its closest antecedent, for ergodic average-reward MDPs, is that of \citealt{wei2020model}) and gives the $\sqrt S$ rate without a transition model, with better bounds, and also under aggregate feedback. We further apply this idea in linear MDPs to obain improved bound as stated next. 

\begin{table}[t]
\caption{Regret bounds for episodic AMDPs with unknown transitions; $S$ is the total number of states over all layers, and bounds stated for per-layer counts are converted. Bounds hold with high probability, except that $^\star$ holds in expectation, for an algorithm that needs only aggregate feedback. The linear block lists efficient algorithms. Lower bounds: \citet{jin2018q,domingues2021episodic}, \citet{lancewicki2025near}, \citet{zhou2021nearly}.}
\label{tab:comparison}
\begin{center}
\begin{footnotesize}
\renewcommand{\arraystretch}{1.1}
\setlength{\tabcolsep}{6pt}
\resizebox{\columnwidth}{!}{%
\begin{tabular}{@{}ll@{}}
\toprule
Algorithm & Regret\\
\midrule
\multicolumn{2}{@{}l}{\emph{Tabular, bandit feedback}}\\
\citet{jin2020learning} & $\tO(HS\sqrt{AK})$\\
\citet{li2026closing} & $\tO(HS\sqrt{AK}+S^3A)$\\
\citet{kulkarni2026introducing} & $\tO(H^{5/2}\sqrt{SAK}+H^4S^2A)$\\
\textbf{\alg{} (ours)} & $\tO(H^{7/4}\sqrt{SAK})$\\
Lower bound & $\Omega(H\sqrt{SAK})$\\
\midrule
\multicolumn{2}{@{}l}{\emph{Tabular, aggregate feedback}}\\
\citet{lancewicki2025near} & $\tO(H^2S\sqrt{AK}+HS^3A)$\\
\textbf{\aggalg{} (ours)} & $\tO(H^{11/4}\sqrt{SAK})$\\
Lower bound & $\Omega(H^{3/2}\sqrt{SAK})$\\
\midrule
\multicolumn{2}{@{}l}{\emph{Linear, bandit feedback}}\\
\citet{liu2026computationally}$^\star$ & $\tO(d^5H^{9/2}\sqrt{K})$\\
\citet{kulkarni2026introducing} & $\tO(d^3H^4\sqrt{K})$\\
\textbf{\linblkalg{} (ours)} & $\tO((d^{3/2}H^{11/4}{+}d^{7/4}H^{9/4})\sqrt{K})$\\
Lower bound & $\Omega(dH^{3/2}\sqrt{K})$\\
\bottomrule
\end{tabular}}
\end{footnotesize}
\end{center}
\vskip -0.1in
\end{table}

\paragraph{Contributions.}
\cref{tab:comparison} lists our bounds and earlier ones.
\begin{itemize}[leftmargin=*,nosep]
\item \textbf{Tabular AMDPs, bandit feedback.} \alg{} has regret $\tO(H^{7/4}\sqrt{SAK})$ (\cref{thm:main}). To our knowledge it is the first algorithm with $\sqrt S$ dependence that needs no transition model, and it uses $O(SA)$ memory and $O(HA)$ amortized time per episode. Its bound has no additive term: it holds in this form for every $K$, while the only other bound with $\sqrt S$ dependence in this setting, $\tO(H^{5/2}\sqrt{SAK}+H^4S^2A)$, takes its leading-order form only for $K\ge H^3S^3A$. The horizon dependence improves by $H^{3/4}$.
\item \textbf{Aggregate feedback.} When only the total cost of each episode is observed, a variant with a surcharge on the recorded returns has regret $\tO(H^{11/4}\sqrt{SAK})$ (\cref{thm:agg_main}), to our knowledge the first bound with $\sqrt S$ dependence in this setting when the transitions are unknown.
\item \textbf{Linear AMDPs.} Block self-normalization extends to linear function approximation: Gram matrices replace visit counts, and the feature regions of \citet{kulkarni2026introducing} replace states. The resulting efficient algorithm improves the previous $\tO(d^3H^4\sqrt K)$ to $\tO((d^{3/2}H^{11/4}+d^{7/4}H^{9/4})\sqrt K)$ for all $d$ and $H$ (\cref{thm:linear_block}).
\item \textbf{New techniques.} A bound on the bias of self-normalized estimates under multiplicative drift, which is relative to the estimated quantity (\cref{lem:sn_informal}), and optimism as a multiplicative discount inside the sampled returns, which places the mean of the estimate between the charges of the learner and of the comparator (\cref{sec:identity}).
\end{itemize}

\paragraph{Comparison with \citet{kulkarni2026introducing}.}\label{sec:related}
Like their algorithm, whose windows of future episodes inspired our blocks, \alg{} evaluates a state over a window of length inversely proportional to its visitation probability, needs the policies to change slowly, and leaves rarely visited states to optimism. The estimator is new to AMDPs, and its analysis is new. They estimate the inverse visitation probability by geometric resampling \citep{neu2016importance}, update in every episode, treat the pending estimates as delayed feedback, and use transition confidence sets. We normalize by the realized visit count of a block, prove that the resulting bias is relative, and pay for it with a multiplicative discount inside the sampled returns. As a result, \alg{} needs no transition model, its bound has no additive term, and the $\sqrt S$ rate extends to aggregate feedback. The horizon dependence improves by $H^{3/4}$: the returns fed to the learners stay in $[0,H]$, and the drift within a block is controlled at the level of trajectory laws (\cref{sec:analysis}; details in \cref{app:compare}).

%% file: sections/2_setting.tex
\section{Problem Setting}\label{sec:setting}

%\paragraph{AMDPs.}
We consider episodic AMDPs with horizon $H\ge 1$ in the standard layered form \citep{neu2010online,jin2020learning}. The state space $\Sset=\Sset_1\cup\dots\cup\Sset_H$ consists of $H$ disjoint nonempty layers with $\Sset_1=\{s_1\}$, and $h(s)\in[H]$ denotes the layer of $s$. The action set $\Aset$ has $A\ge 2$ elements. From $(s,a)$ with $h(s)<H$ the next state is drawn from an unknown kernel $p(\cdot\mid s,a)\in\Delta(\Sset_{h(s)+1})$. For $g:\Sset\to\R$ we write $(pg)(s,a):=\sum_{s'}p(s'\mid s,a)g(s')$, and $(pg)(s,a):=0$ on $\Sset_H\times\Aset$. In \emph{tabular} AMDPs, $S:=|\Sset|=\sum_{h=1}^H|\Sset_h|\ge H$ is the total number of states. A non-layered episodic MDP with state set $\Sset_{\mathrm{base}}$ and a fixed initial state becomes layered by taking $\Sset_1=\{(s_1,1)\}$ and $\Sset_h=\Sset_{\mathrm{base}}\times\{h\}$ for $h\ge 2$, so bounds convert through $S\le H|\Sset_{\mathrm{base}}|$.

\paragraph{Interaction protocol.}
An oblivious adversary fixes cost functions $c^1,\dots,c^K:\Sset\times\Aset\to[0,1]$ before the first episode, with $K\ge 2$. In episode $k\in[K]$ the learner chooses a Markov policy $\pi^k:\Sset\to\Delta(\Aset)$ based on its observations from episodes $1,\dots,k-1$, starts at $s_1^k=s_1$, and for $h=1,\dots,H$ draws $a_h^k\sim\pi^k(\cdot\mid s_h^k)$ and $s_{h+1}^k\sim p(\cdot\mid s_h^k,a_h^k)$. It always observes the trajectory $\tau^k=(s_1^k,a_1^k,\dots,s_H^k,a_H^k)$. Under \emph{bandit feedback} it also observes the visited costs $c^k(s_h^k,a_h^k)$, $h\in[H]$; under \emph{aggregate feedback} it observes only their sum $C^k$. We write $\F_k$ for the $\sigma$-algebra generated by the first $k$ episodes, including any internal randomization of the learner, and $\ind_k[s,a]:=\ind\{s_{h(s)}^k=s,\,a_{h(s)}^k=a\}$. The learner knows $H$, $A$, $K$, the confidence level $\delta$ and $S$ (or $d$).

\paragraph{Values, occupancies and regret.}
For a policy $\pi$ and a cost function $c$, define, backward from layer $H$, $Q^\pi(s,a;c) := c(s,a)+(pV^\pi(\cdot\,;c))(s,a)$ and $V^\pi(s;c) := \ip{\pi(\cdot\mid s)}{Q^\pi(s,\cdot\,;c)}$, where $\ip{u}{v}$ is the inner product on $\R^{\Aset}$ (or on $\R^d$). The occupancy of $\pi$ is $q^\pi(s):=\Prob_{p,\pi}(s_{h(s)}=s)$ and $q^\pi(s,a):=q^\pi(s)\pi(a\mid s)$, so that $\sum_{s\in\Sset_h}q^\pi(s)=1$ for every layer $h$. With $\pi^\star\in\argmin_\pi\sum_{k=1}^K V^\pi(s_1;c^k)$, the regret is $\Reg:=\sum_{k=1}^K\bigl(V^{\pi^k}(s_1;c^k)-V^{\pi^\star}(s_1;c^k)\bigr)$. We abbreviate $Q^k:=Q^{\pi^k}(\cdot\,;c^k)$, $V^k:=V^{\pi^k}(\cdot\,;c^k)$, $q^k:=q^{\pi^k}$ and $q^\star:=q^{\pi^\star}$. We write $\tO$, $\tOmega$ and $\tTheta$ to hide factors polylogarithmic in $S,A,H,K,d$ and $1/\delta$.

%% file: sections/3_algorithm.tex
\section{Self-Normalized Policy Optimization}\label{sec:algorithm}

This section presents \alg{} (\cref{alg:snpo}) for tabular AMDPs with bandit feedback. As in other policy optimization methods \citep{shani2020optimistic,luo2021policy}, every state runs its own online learner over the actions.

\subsection{The Algorithm}\label{sec:snpo_design}

\paragraph{Intuition.} The rationale for local freezing is as follows.
Suppose that a state $s$ keeps its action distribution $\pi(\cdot\mid s)$ fixed during a block of $\Lambda$ consecutive episodes and is visited $N$ times. Summing the returns that follow action $a$ over these visits, dividing by $N\pi(a\mid s)$ and multiplying by $\Lambda$ estimates the \emph{sum} over the block of the cost-to-go of $(s,a)$, with the realized visit frequency $N/\Lambda$ in place of the unknown $q^k(s)$. If the other states do not change their policies during the block, the visits are exchangeable and the estimate is unbiased up to the probability of an empty block, however the costs vary within the block and however few visits it contains. A block therefore needs only logarithmically many visits, and the bias is governed by how much the other policies move during the block, not by $N^{-1/2}$. Three ideas turn this observation into an algorithm.

\emph{Visitation-adapted blocks.} A state visited with probability $q$ uses blocks of length about $M/q$ for a logarithmic $M$. Block lengths are powers of two and blocks are aligned, so the blocks of different states are nested (\cref{fig:dyadic}). States that are too rare to fill a block stay \emph{inactive} and are not estimated.

\emph{Relative bias from multiplicative stability.} Other states update while a state waits for its block to end, so $q^k(s)$ and the cost-to-go drift inside the block. We update with log-barrier mirror descent, which changes every action probability by a small multiplicative factor, and prove that the bias of the self-normalized estimate is then a small fraction of the estimated quantity itself (\cref{lem:sn_informal}).

\emph{Optimism as a discount inside the returns.} We pay for this bias, and for the states that are not estimated, by making the sampled returns optimistic: every future cost is discounted at each step, and a return is cut at the first unestimated state, as in R-max \citep{brafman2002rmax}. No bonus function and no confidence set is needed.

\paragraph{Periods and blocks.}
The $K$ episodes are split into periods of $\Lmax$ episodes. All policies start uniform, and every state is \emph{inactive} during the first period. At the start of each period $j\ge 2$, every state $s$ looks at its number of visits $N_{j-1}(s)$ during period $j-1$. If $N_{j-1}(s)<M$, for a logarithmic number $M$, the state is inactive in period $j$: it keeps its action distribution and forms no estimate. Otherwise the state is \emph{active} and divides period $j$ into blocks of length
\begin{equation}\label{eq:blocklength}
  \Lambda_j(s) := \min\bigl\{2^i : i\in\mathbb Z_{\ge 0},\ 2^i N_{j-1}(s) \ge M\Lmax\bigr\}.
\end{equation}
Since $N_{j-1}(s)/\Lmax$ estimates the visitation probability of $s$, an active state expects about $M$ to $2M$ visits per block. It keeps $\pi(\cdot\mid s)$ fixed inside each block and updates it when the block ends. Since $\Lambda_j(s)$ is a power of two that divides $\Lmax$ and blocks end at multiples of $\Lambda_j(s)$, the blocks of any two states are nested or disjoint (\cref{fig:dyadic}).

\input{sections/fig_dyadic}

\paragraph{The block estimate.}
Let $s\in\Sset_h$ be active and let $b$ be one of its blocks, with length $\Lambda_b=\Lambda_j(s)$ and action distribution $\pi_b$. A visit to $s$ in episode $k$ records a return $Y_h^k$: the cost that the episode incurs from layer $h$ on, discounted as specified in \eqref{eq:sample_returns} below. During the block, the state counts its visits $N_b:=\sum_{k\in b}\ind\{s_h^k=s\}$ and accumulates the returns that follow each action, $\Sigma_b(a):=\sum_{k\in b}\ind_k[s,a]\,Y_h^k$. Importance weighting would estimate the sum over the block of the discounted cost-to-go of $(s,a)$ by $\sum_{k\in b}\ind_k[s,a]Y_h^k/(q^k(s)\pi_b(a))$, which requires the unknown probability $q^k(s)$ that episode $k$ visits $s$. \alg{} replaces $q^k(s)$ by the visit frequency $N_b/\Lambda_b$ of the block itself:
\begin{equation}\label{eq:plain}
  \frac{\Lambda_b}{N_b}\cdot\frac{\Sigma_b(a)}{\pi_b(a)} .
\end{equation}
Two properties of this estimate carry the analysis. First, if no other state changed its policy during the block, the visits would be exchangeable, and \eqref{eq:plain} would be unbiased for the block sum up to the probability of an empty block, although the costs change from episode to episode (\cref{eq:frozen} below). Second, its average under $\pi_b$ is at most $H\Lambda_b$ on every sample path, however rarely $s$ is visited, because the returns are averaged over the visits; importance weights are of order $1/q^k(s)$.

\begin{table}[t]
\caption{Parameters of \alg{} in \cref{thm:main}, up to universal constants; the exact values are in \cref{tab:params} (\cref{app:notation}).}
\label{tab:params_main}
\begin{center}
\begin{footnotesize}
\setlength{\tabcolsep}{4pt}
\begin{tabular}{@{}lll@{}}
\toprule
Symbol & Value & Role\\
\midrule
$J$ & $\log(64SAK^3/\delta)$ & logarithmic factor\\
$D$ & $\min\{H,\sqrt{HJ}\}$ & drift of trajectory laws\\
$M$ & $\Theta(J)$ & visits per block\\
$\eta$ & $\Theta\bigl(\sqrt{A/(HDSK)}\bigr)$ & OMD step size\\
$\gamma_0$ & $\Theta\bigl(\sqrt{H/(ASK)}\bigr)$ & implicit exploration\\
$a_0$ & $\Theta(\eta HD)$ & margin per unit block length\\
$\kappa$ & $a_0+A\gamma_0$ & discount per unit block length\\
$\Lmax$ & largest $2^m\le 1/(4\kappa)$ & period length\\
\bottomrule
\end{tabular}
\end{footnotesize}
\end{center}
\end{table}

\paragraph{Optimism inside the returns.}
Other states do update during a block, and inactive states are not estimated. \alg{} pays for both by recording returns that are smaller than the actual cost-to-go. A block $b$ uses a margin $\epsilon_b:=a_0\Lambda_b$ and an exploration parameter $\gamma_b:=\gamma_0\Lambda_b$, both proportional to its length (\cref{tab:params_main} gives $a_0$ and $\gamma_0$), and we write $f_b:=1-\epsilon_b$ and $\chi_b(a):=\pi_b(a)/(\pi_b(a)+\gamma_b)$. With $b$ the block of $s$ containing episode $k$, each pair $(s,a)$ receives the \emph{survival weight} (its form is explained after \eqref{eq:omd_update})
\begin{equation}\label{eq:w}
  w^k(s,a) := \begin{cases}
  f_b^2\,\chi_b(a) & \text{if $s$ is active},\\
  0 & \text{if $s$ is inactive},
  \end{cases}
\end{equation}
and the return recorded at layer $h$ of episode $k$ is
\begin{equation}\label{eq:sample_returns}
  Y_h^k := \sum_{h'=h}^{H}\Bigl(\prod_{m=h+1}^{h'}w^k(s_m^k,a_m^k)\Bigr)\,c^k(s_{h'}^k,a_{h'}^k),
\end{equation}
that is, $Y_h^k=c^k(s_h^k,a_h^k)+w^k(s_{h+1}^k,a_{h+1}^k)\,Y_{h+1}^k$: the weight of a pair discounts everything from that pair on. The returns lie in $[0,H]$. At an active state the weights average at least $1-2\epsilon_b-A\gamma_b$ under $\pi_b$, and the discount grows with the block length, like an exploration bonus that scales inversely with the visitation probability \citep{luo2021policy}. At an inactive state the weight is zero, so the returns of all earlier layers stop just before this state: the region that the algorithm has not learned is treated as free of cost, which is the optimism of R-max \citep{brafman2002rmax}.

\paragraph{Policy Update.}
When a block ends, the state forms the estimate
\begin{equation}\label{eq:estimator}
  \widehat\ell_b(a) := f_b\,\frac{\Lambda_b}{N_b}\,\frac{\Sigma_b(a)}{\pi_b(a)+\gamma_b}\qquad(\widehat\ell_b:=0\text{ if }N_b=0),
\end{equation}
which is \eqref{eq:plain} with implicit exploration \citep{neu2015explore} and a factor $f_b$, and performs one step of online mirror descent (OMD) with the log-barrier regularizer \citep{foster2016learning} $\psi(\pi):=-\sum_{a}\log\pi(a)$ and its Bregman divergence $D_\psi$:
\begin{equation}\label{eq:omd_update}
  \pi(\cdot\mid s) \leftarrow \argmin_{\pi\in\Delta(\Aset)}\bigl\{\eta\ip{\pi}{\widehat\ell_b} + D_\psi(\pi,\pi_b)\bigr\}.
\end{equation}
The log-barrier turns the pathwise bound on the estimate into multiplicative stability: an update changes every action probability by a factor $e^{\pm O(\eta H\Lambda_b)}$ (\cref{lem:barrier}). The factors $f_b$ create the margins of the analysis. With $v^k(s,a)$ the mean of $Y_h^k$ given a visit to $(s,a)$, the regret decomposition of \cref{sec:identity} charges the learner and the comparator at $s$ on the two sides of the mean of the estimate,
\[
  \underbrace{f_b^2\chi_b\,v^k}_{\text{learner}}\;\le\;\underbrace{f_b\chi_b\,v^k}_{\text{estimate}}\;\le\;\underbrace{v^k}_{\text{comparator}},
\]
and the two gaps, of relative size at least $\epsilon_b$, pay for all errors.

\begin{algorithm}[t]
  \caption{\alg{} for tabular AMDPs with bandit feedback. $N^{\mathrm{blk}}(s)$ and $\Sigma^{\mathrm{blk}}(s,\cdot)$ are the counters $N_b$ and $\Sigma_b$ of the current block of $s$.}
  \label{alg:snpo}
  \begin{footnotesize}
  \begin{algorithmic}[1]
    \STATE \textbf{Input:} $\eta,\gamma_0,a_0,M,\Lmax$ as in \cref{tab:params_main}. If $\kappa:=a_0+A\gamma_0>1/4$, play $\Unif(\Aset)$ in all episodes.
    \STATE Initialize $\pi(\cdot\mid s)\leftarrow\Unif(\Aset)$ for all $s\in\Sset$.
    \FOR{period $j=1,2,\dots,\lceil K/\Lmax\rceil$}
      \FORALL{$s\in\Sset$}
        \STATE If $j=1$ or $N_{j-1}(s)<M$, mark $s$ inactive for period $j$. Otherwise mark $s$ active, set $\Lambda_j(s)$ by \cref{eq:blocklength}, and reset $N^{\mathrm{blk}}(s)\leftarrow 0$, $\Sigma^{\mathrm{blk}}(s,\cdot)\leftarrow 0$.
        \STATE Reset the period counter $N_j(s)\leftarrow 0$.
      \ENDFOR
      \FOR{episode $k=(j-1)\Lmax+1,\dots,\min\{j\Lmax,K\}$}
        \STATE Play $\pi^k=\pi$; observe $(s_h^k,a_h^k,c^k(s_h^k,a_h^k))_{h=1}^H$.
        \STATE Compute $Y_H^k,\dots,Y_1^k$ by \cref{eq:w,eq:sample_returns}.
        \STATE For each $h$, with $s=s_h^k$, $a=a_h^k$: $N_j(s)\leftarrow N_j(s)+1$ and, if $s$ is active, $N^{\mathrm{blk}}(s)\leftarrow N^{\mathrm{blk}}(s)+1$, $\Sigma^{\mathrm{blk}}(s,a)\leftarrow\Sigma^{\mathrm{blk}}(s,a)+Y_h^k$.
        \FORALL{active $s$ with $k \bmod \Lambda_j(s)=0$}
          \STATE Form $\widehat\ell_b$ by \cref{eq:estimator} from $N^{\mathrm{blk}}(s),\Sigma^{\mathrm{blk}}(s,\cdot)$; update $\pi(\cdot\mid s)$ by \cref{eq:omd_update}; reset $N^{\mathrm{blk}}(s),\Sigma^{\mathrm{blk}}(s,\cdot)\leftarrow 0$.
        \ENDFOR
      \ENDFOR
    \ENDFOR
  \end{algorithmic}
  \end{footnotesize}
\end{algorithm}

\subsection{Regret Guarantee}\label{sec:result}

The factor $D$ of \cref{tab:params_main} bounds the drift of the probability of a trajectory in units of the per-step drift of the policy: $H$ in the worst case, and $\sqrt{HJ}$ outside an event of negligible probability (\cref{sec:drift_sn}). The period length is $\Lmax=\Theta(\sqrt{SK/(AHD)})$, and a state can be active only if its visitation probability is about $M/\Lmax=\Theta(J\sqrt{AHD/(SK)})$ or more.

\begin{theorem}[Tabular AMDPs with bandit feedback]\label{thm:main}
Consider any layered AMDP with $S$ states, $A$ actions, horizon $H$, unknown transitions and bandit feedback, as in \cref{sec:setting}. For every $\delta\in(0,1)$, \textup{\alg{}} with the parameters of \cref{tab:params_main,tab:params} satisfies, with probability at least $1-\delta$,
\[
  \Reg = O\bigl(H^{3/2}\sqrt{D\,SAK}\,J\bigr)=\tO\bigl(H^{7/4}\sqrt{SAK}\bigr).
\]
Here $J=\log(64SAK^3/\delta)$ and $D=\min\{H,\sqrt{HJ}\}$. The same bound holds for an implementation with $O(SA)$ memory and $O(HA)$ amortized time per episode (\cref{app:extensions}).
\end{theorem}

The bound is optimal in $S$, $A$ and $K$ and within a factor $H^{3/4}$ of the lower bound, up to logarithmic factors. Compared with the $\tO(H^{5/2}\sqrt{SAK}+H^4S^2A)$ of \citet{kulkarni2026introducing}, the leading term improves by $H^{3/4}$, and there is no additive term: the bound has the form above for every $K$, while theirs is $\tO(H^{5/2}\sqrt{SAK})$ only for $K\ge H^3S^3A$. Compared with the $\tO(HS\sqrt{AK})$ bounds of \citet{jin2020learning} and \citet{li2026closing}, it is smaller whenever $S>H^{3/2}$.

%% file: sections/fig_dyadic.tex
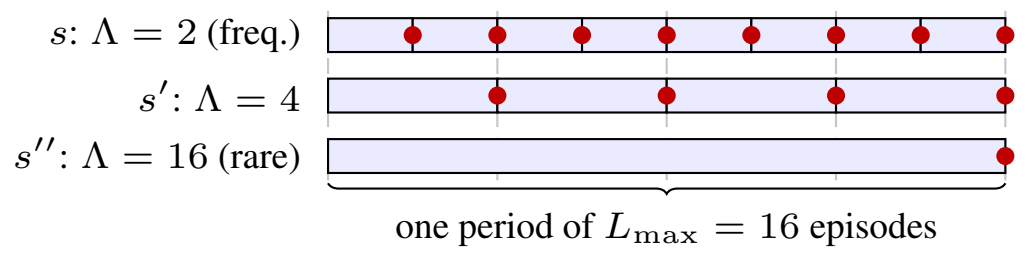
\begin{figure}[!t]
\centering
\begin{tikzpicture}[x=0.28cm,y=0.40cm]
  \foreach \g in {0,4,8,12,16}{
    \draw[densely dashed,gray!45] (\g,-0.40) -- (\g,2.40);
  }
  \foreach \row/\len/\lab in {
    2/2/{$s$: $\Lambda=2$ (freq.)},
    1/4/{$s'$: $\Lambda=4$},
    0/16/{$s''$: $\Lambda=16$ (rare)}}{
    \node[anchor=east,font=\scriptsize] at (-0.2,\row) {\lab};
    \pgfmathtruncatemacro{\nb}{16/\len}
    \foreach \i in {1,...,\nb}{
      \draw[fill=blue!8] ({(\i-1)*\len},\row-0.28) rectangle ({\i*\len},\row+0.28);
    }
    \foreach \i in {1,...,\nb}{
      \fill[red!75!black] ({\i*\len},\row) circle (1.7pt);
    }
  }
  \draw[decorate,decoration={brace,amplitude=2pt,mirror}] (0,-0.45) -- (16,-0.45)
     node[midway,below=2pt,font=\scriptsize] {one period of $\Lmax=16$ episodes};
\end{tikzpicture}
\caption{Aligned dyadic blocks in one period (illustrative values, $\Lmax=16$); dots mark updates. The blocks of any two states are nested or disjoint.}
\label{fig:dyadic}
\end{figure}

%% file: sections/4_analysis.tex
\section{Analysis}\label{sec:analysis}

We outline the proof of \cref{thm:main} in three steps. Complete proofs are in \crefrange{app:notation}{app:main_proof}, with a dependency graph in \cref{fig:roadmap}. Throughout, $C$ denotes a universal constant whose value may change from line to line, and we assume $\kappa\le 1/4$ and $\eta\ge 1/(HK)$; otherwise the bound of \cref{thm:main} exceeds $HK$.

\subsection{Regret Decomposition}\label{sec:identity}

The returns of \alg{} are unbiased samples of a discounted cost-to-go. Define, backward from layer $H$,
\begin{equation}\label{eq:vz}
\begin{gathered}
  v^k := c^k + p V^{z,k},\quad z^k := w^k v^k,\quad r^k := (1-w^k)\,v^k,\\
  V^{z,k}(s) := \ip{\pi^k(\cdot\mid s)}{z^k(s,\cdot)}.
\end{gathered}
\end{equation}
Then $v^k(s,a)=\E[Y_{h(s)}^k\mid\F_{k-1},\,s_{h(s)}^k=s,\,a_{h(s)}^k=a]$ whenever $(s,a)$ is reached with positive probability, so $v^k$ is the target of the estimator at $s$. The function $z^k$ discounts $v^k$ once more by the weight of the pair itself, and it is the $Q$-function of $\pi^k$ for the cost $c^k-r^k$, so $r^k\ge 0$ acts as an exploration bonus. For each state let
\begin{equation}\label{eq:Bresidual}
  \Phi^k(s) := \ip{\pi^k(\cdot\mid s)}{z^k(s,\cdot)} - \ip{\pi^\star(\cdot\mid s)}{v^k(s,\cdot)}.
\end{equation}

\begin{lemma}[Regret decomposition]\label{lem:identity}
For all $k$ and $(s,a)$, $0\le z^k(s,a)\le v^k(s,a)\le Q^k(s,a)\le H$, and
\begin{equation}\label{eq:decomp}
  \Reg = \sum_{s\in\Sset} q^\star(s)\sum_{k=1}^K \Phi^k(s)
       + \sum_{k=1}^K\sum_{s,a} q^k(s,a)\,r^k(s,a).
\end{equation}
If $s$ is inactive in episode $k$, then $\Phi^k(s)\le 0$.
\end{lemma}

The identity is the performance difference lemma for the cost $c^k-r^k$, as in the bonus lemma of \citet{luo2021policy}, without dilation. The first sum collects the \emph{local regret} of each state, weighted by the comparator's occupancy $q^\star$. The second sum is the \emph{cost of optimism}, weighted by the learner's own occupancy $q^k$. The distinction matters because $q^\star(s)$ is unknown and can be much larger than $q^k(s)$: an estimation error left in $\Phi^k(s)$ at a rarely visited state can be expensive, whereas a quantity charged through $r^k$ costs only in proportion to how often the learner visits $s$. The definition of $\Phi^k$ moves errors from the first sum to the second: the learner is charged $z^k$ and the comparator the larger $v^k$, and at an active state the difference is a multiplicative margin of order $\epsilon_b$, inside which all estimation errors at $s$ will fit.

\subsection{Self-Normalization under Drift}\label{sec:drift_sn}

Fix an active state $s\in\Sset_h$ and one of its blocks $b=\{t_0,\dots,t_0+\Lambda-1\}$, and let $X_k:=\ind\{s_h^k=s\}$. If all policies were frozen during $b$, the indicators $X_k$ would be i.i.d.\ with some mean $q$, and by symmetry $\E[X_k/N_b]=(1-(1-q)^\Lambda)/\Lambda$ for every $k\in b$. Self-normalization would then be exact up to the probability of an empty block, although the targets $v^k$ vary with $k$:
\begin{equation}\label{eq:frozen}
  \E\Bigl[\frac{\Lambda}{N_b}\frac{\Sigma_b(a)}{\pi_b(a)}\Bigr]=\bigl(1-(1-q)^\Lambda\bigr)\sum_{k\in b}v^k(s,a).
\end{equation}
In the algorithm, other states update during $b$, so $q^k(s)$ and $v^k(s,a)$ drift in a data-dependent way.

%\paragraph{Policies move little.}
We note a state $s'$ changes its policy inside $b$ only at the ends of its own blocks, which are disjoint and contained in $b$ because blocks are aligned, and an update after a block of length $\Lambda'$ moves its log-probabilities by at most $C\eta H\Lambda'$. Hence, on every sample path, every $\log\pi^k(a'\mid s')$ changes by at most $C\eta H\Lambda$ during $b$ (\cref{lem:dyadic}).

%\paragraph{Trajectory laws move little.}
The next bound concerns two fixed policies and is the origin of $D$.

\begin{lemma}[Path-law domination; \cref{lem:pathlaw}]\label{lem:pathlaw_informal}
Let $P$ and $P'$ be the laws of a trajectory under two policies whose log-probabilities differ by at most $\vartheta\le 1$ at every state-action pair. For every $y>0$ and every function $F$ of the trajectory with values in $[0,F_{\max}]$, $\E_P[F]\le e^{4H\vartheta^2+\vartheta\sqrt{2Hy}}\,\E_{P'}[F]+F_{\max}e^{-y}$, and also $\E_P[F]\le e^{H\vartheta}\,\E_{P'}[F]$.
\end{lemma}

The second bound is the worst case: the likelihood ratio of a trajectory is a product of $H$ factors in $e^{\pm\vartheta}$. The first uses that the logarithm of this ratio is a sum of $H$ bounded terms whose conditional means are of second order, so that it concentrates at the scale $\vartheta\sqrt H$. This is the argument of advanced composition in differential privacy \citep{dwork2010boosting}. With $y$ a multiple of $J$, the small per-step drift $\vartheta$ of the algorithm thus becomes a drift of order $\vartheta D$ of every trajectory probability, up to a negligible additive term.

\emph{Notation.} We write $x\in e^{\pm\varrho}y\pm c$ for $e^{-\varrho}y-c\le x\le e^{\varrho}y+c$, call a block \emph{complete} if it is not cut short by the last episode, and, for a block that starts at episode $t_0$, let $v^{\circ,k}$ be defined as $v^k$ in \eqref{eq:vz}, with the costs of episode $k$ but with the policies, and the weights computed from them, of episode $t_0$.

\begin{lemma}[Drift within a block; \cref{lem:dyadic,lem:drift}]\label{lem:drift_informal}
Let $b=\{t_0,\dots,t_0+\Lambda-1\}$ be a complete block of an active state $s$, let $k\in b$, $\varrho_b:=80\eta HD\Lambda$ and $\etail:=e^{-20J}$. On every sample path: (a) $v^k(s,a)\in e^{\pm\varrho_b}v^{\circ,k}(s,a)\pm\etail H$ for all $a\in\Aset$; (b) if $q^{t_0}(s)\Lambda\ge M/3$, then $q^k(s)\in e^{\pm\varrho_b}q^{t_0}(s)$.
\end{lemma}

For $v^k$ the discounting is represented as random stopping, so that \cref{lem:pathlaw_informal} applies to a trajectory that draws an action and a survival coin at every step. The additive error $\etail$ does not spoil the relative bound on $q^k(s)$, because $q^{t_0}(s)\ge M/(3K)$ is larger by many orders of magnitude. The next lemma is stated for a general block. In \alg{} it is applied with $R_k=\ind\{a_h^k=a\}Y_h^k/\pi_b(a)$, $g_k=v^k(s,a)$, $\bar q=q^{t_0}(s)$, $g_k^\circ=v^{\circ,k}(s,a)$ and $\varrho=\varrho_b\le 0.1$, on the event $\Omega_0$ that $b$ is a block of $s$ with $q^{t_0}(s)\Lambda\ge M/3$ (\cref{app:sn_applications}).

\begin{lemma}[Relative bias of self-normalization; simplified from \cref{lem:relative_sn}]\label{lem:sn_informal}
Fix $s\in\Sset_h$, a deterministic block $b=\{t_0,\dots,t_0+\Lambda-1\}$ and $\varrho\in[0,0.1]$. Let $R_k\ge 0$ be integrable and $\F_k$-measurable, let $\bar q\in(0,1]$ and $g_k^\circ\in[0,H]$ be $\F_{t_0-1}$-measurable, and let $\Omega_0\in\F_{t_0-1}$. Suppose that, almost surely on $\Omega_0$, $\bar q\Lambda\ge M/3$ and, for all $k\in b$,
\begin{enumerate}[label=(\roman*),leftmargin=*,nosep]
\item $\E[X_kR_k\mid\F_{k-1}]=q^k(s)\,g_k$ for some $\F_{k-1}$-measurable $g_k\in[0,H]$,
\item $q^k(s)\in e^{\pm\varrho}\bar q$ and $g_k\in e^{\pm\varrho}g_k^\circ\pm\etail H$.
\end{enumerate}
Then, on $\Omega_0$, with $N_b:=\sum_{k\in b}X_k$ and $0/0:=0$,
\begin{align*}
  &\Bigl|\E\Bigl[\frac{\Lambda}{N_b}\sum_{k\in b}X_kR_k-\sum_{k\in b}g_k\Bigm|\F_{t_0-1}\Bigr]\Bigr|\\
  &\le 5\varrho\,\E\Bigl[\sum_{k\in b}g_k\Bigm|\F_{t_0-1}\Bigr]+\Lambda H\bigl(3\etail+e^{-M/3}\bigr).
\end{align*}
\end{lemma}

Without drift the bound reduces to \eqref{eq:frozen}. The references $\bar q$ and $g_k^\circ$ must be fixed before the block starts, so that they are decoupled from the visit count $N_b$, which depends on the whole block. This is why \cref{lem:drift_informal} compares with the policies frozen at $t_0$. The proof couples the visits with independent uniform variables through $X_k=\ind\{U_k\le q^k(s)\}$. Given $X_k=1$, the count $N_b$ then lies between $1$ plus a binomial variable with success probability $e^{-\varrho}\bar q$ and $1$ plus one with success probability $e^{\varrho}\bar q$, both independent of episode $k$, with expected reciprocal $1/(\Lambda\bar q)$ up to a factor $e^{\pm\varrho}$ and an error $e^{-M/3}$, which cancels the factor $q^k(s)$ (\cref{app:selfnorm}). In \alg{}, \cref{lem:drift_informal} supplies hypothesis (ii), and each block sum is estimated in conditional expectation up to a factor $1\pm 5\varrho_b$ and a negligible term.

\subsection{Local Regret and Cost of Optimism}\label{sec:classification}

The hypothesis $q^{t_0}(s)\Lambda\ge M/3$ of \cref{lem:drift_informal} holds for all blocks with high probability, because the visit counts of a period predict the occupancies of the next: with high probability, in every episode $k$ of every period $j\ge 2$, every active state has $\frac13M\le q^k(s)\Lambda_j(s)\le 5M$ and every inactive state has $q^k(s)\le 3M/\Lmax$ (\cref{lem:classification}).

\paragraph{Local regret.}
Fix a state $s$, one of its complete blocks $b$ starting at $t_0$, and the smoothed comparator 
$$u:=(1-\frac1K)\pi^\star(\cdot\mid s)+\frac1K\Unif(\Aset).$$ Write $\E_b[\cdot]:=\E[\cdot\mid\F_{t_0-1}],$ $m_b:=\sum_{k\in b}\ip{\pi_b}{\chi_bv^k(s,\cdot)}$ and $m_b^{u}:=\sum_{k\in b}\ip{u}{v^k(s,\cdot)}$; both are random, since other states update inside $b$. By \eqref{eq:w}, the learner's part of $\sum_{k\in b}\Phi^k(s)$ is $f_b^2\,m_b$, and the comparator's part is $m_b^{u}$ up to the smoothing. Three steps compare them.

\emph{Bias.} \cref{lem:sn_informal} and, for the second line, $\chi_b\le 1$ give, up to additive terms that sum to at most $H$ over all blocks of $s$,
\begin{align*}
  \E_b\ip{\pi_b}{\widehat\ell_b}&\ge f_b(1-5\varrho_b)\,\E_b[m_b],\\
  \E_b\ip{u}{\widehat\ell_b}&\le f_b(1+5\varrho_b)\,\E_b[m_b^{u}].
\end{align*}
So, up to the bias, the learner's charge $f_b^2m_b$ is below its estimated loss by the margin $f_b\epsilon_bm_b$, and the estimated loss of $u$ is below its charge $m_b^{u}$ by the margin $\epsilon_bm_b^{u}$.

\emph{Mirror descent.} Summing over the blocks of $s$, log-barrier OMD compares the two estimated losses (\cref{lem:barrier}):
\[
  \sum_b(1-2\eta H\Lambda_b)\ip{\pi_b}{\widehat\ell_b}\le\sum_b\ip{u}{\widehat\ell_b}+\frac{A\log K}{\eta}.
\]

\emph{Concentration.} Freedman's inequality bounds the deviations from the conditional means by a variance term per block plus $J/\eta$, and implicit exploration bounds the upper deviations of $\ip{u}{\widehat\ell_b}$ by $HJ/(2\gamma_0)$ in total (\cref{lem:ix} in \cref{app:events}).

The bias, the stability term of OMD and the variance terms are all multiples of the block's own loss: $C\eta HD\Lambda_b$ for the bias and $C\eta H\Lambda_b$ for the other two. The margin $\epsilon_b=a_0\Lambda_b$ with $a_0=\Theta(\eta HD)$ therefore covers them in every block (\cref{tab:budget}); what remains occurs once per state.

\begin{lemma}[Local regret]\label{lem:local}
There is an event $\good$ of probability at least $1-\delta$ on which every $s\in\Sset$ satisfies
\[
  \sum_{k=1}^K \Phi^k(s) \le \frac{A\log K + 2J}{\eta} + \frac{HJ}{2\gamma_0} + H\Lmax + 2H.
\]
\end{lemma}

This bound does not depend on how often $s$ is visited, so summing it against $q^\star$ in \eqref{eq:decomp} costs only a factor $H$.

\paragraph{Cost of optimism.}
By \eqref{eq:w}, an active state has $\ip{\pi^k(\cdot\mid s)}{r^k(s,\cdot)}\le 2H\kappa\Lambda_j(s)$. The block length cancels against the occupancy, since $q^k(s)\Lambda_j(s)\le 5M$, so an active state costs $O(HM\kappa)$ per episode whatever its visitation probability, and an inactive state costs at most $Hq^k(s)=O(HM\kappa)$. Together with the first period, on $\good$,
\begin{equation}\label{eq:optcost}
  \sum_{k=1}^K\sum_{s,a} q^k(s,a)\,r^k(s,a) \le H\Lmax + C\,HSMK\kappa .
\end{equation}

\paragraph{Conclusion.}
Inserting \cref{lem:local} and \eqref{eq:optcost} into \eqref{eq:decomp} and using $H\Lmax=O(1/\eta)$ gives, on $\good$,
\[
  \Reg = O\Bigl(\tfrac{HAJ}{\eta}+\tfrac{H^2J}{\gamma_0}+HSMK\bigl(\eta HD+A\gamma_0\bigr)+H^2\Bigr),
\]
and the parameters of \cref{tab:params_main} balance the terms (\cref{app:main_proof}). With $D=1$ the bound would be $\tO(H^{3/2}\sqrt{SAK})$. The remaining factor $\sqrt D$ comes from the drift of trajectory probabilities within a block, which enters through $a_0$. \cref{lem:pathlaw_informal} is tight up to constants for some pairs of policies and rarely visited states (\cref{app:horizon}); whether the iterates of \alg{} produce such a drift is open.

%% file: sections/5_aggregate.tex
\section{Aggregate Feedback}\label{sec:aggregate}

Under aggregate feedback the learner observes $\tau^k$ and only the total cost $C^k\in[0,H]$. The schedule of \alg{} and its update carry over; what has to change is the return that a state records at a visit.

\paragraph{Technical Challenges.}\label{sec:agg_obstruction}
First, the discounted return \eqref{eq:sample_returns} cannot be computed from $C^k$, and in general no statistic of $(\tau^k,C^k)$ has the same conditional mean, even up to an action-independent offset (\cref{prop:agg_transfer}). The natural substitute is $C^k$ itself: given a visit to $(s,a)$ its mean is $\pre^k(s)+Q^k(s,a)$, where $\pre^k(s)$ is the expected cost collected before reaching $s$. This prefix cost cancels in the local regret of $s$ if $C^k$ enters with the same weight for the learner and the comparator \citep{lancewicki2025near}. Second, an undiscounted return offers no margin for the self-normalization bias and no weight zero at inactive states, so $\Phi^k(s)\le 0$ at inactive states (\cref{lem:identity}) fails.

\paragraph{Randomized surcharges.}\label{sec:agg_design}
\aggalg{} (\cref{alg:agg_diff}) keeps $C^k$ at unit weight and creates the margin with an additive term that the algorithm controls completely. Its survival weights are those of \eqref{eq:w} with the larger margin $\epsilon_b:=(a_0+4A\gamma_0)\Lambda_b$. After episode $k$ it draws, for every layer $h'\ge 2$, an independent coin that succeeds with probability $w^k(s_{h'}^k,a_{h'}^k)$. Let $\sigma_h^k\in\{h+1,\dots,H+1\}$ be the first layer after $h$ whose coin fails, with $\sigma_h^k=H+1$ if all succeed. A visit at layer $h$ to an active state, in a block $b$ of that state, records
\begin{equation}\label{eq:agg_loss}
  Y_h^k := C^k + f_b\cdot 2H\bigl(\sigma_h^k-h\bigr),
\end{equation}
that is, the total cost plus a \emph{surcharge} of $2H$ for layer $h$ and for every layer survived after it, multiplied by $f_b$. Returns lie in $[0,B]$ with $B:=2H^2+H$.

\begin{algorithm}[t]
  \caption{\aggalg{}, as changes to \cref{alg:snpo} (complete pseudocode: \cref{alg:agg} in \cref{app:aggregate})}
  \label{alg:agg_diff}
  \begin{footnotesize}
  \begin{algorithmic}[1]
    \STATE \textbf{Input:} parameters of \cref{sec:aggregate}; uniform policy if $\kagg>1/4$.
    \STATE \textbf{Weights \eqref{eq:w}:} with the margin $\epsilon_b:=(a_0+4A\gamma_0)\Lambda_b$.
    \STATE \textbf{Line 9:} observe only $\tau^k$ and $C^k=\sum_{h}c^k(s_h^k,a_h^k)$.
    \STATE \textbf{Line 10:} for each layer $h'\ge 2$ draw a coin with success probability $w^k(s_{h'}^k,a_{h'}^k)$; set $Y_h^k$ by \cref{eq:agg_loss}.
    \STATE \textbf{Line 13:} $\widehat\ell_b(a)\leftarrow\frac{\Lambda_b}{N_b}\frac{\Sigma_b(a)}{\pi_b(a)+\gamma_b}$, without the factor $f_b$.
  \end{algorithmic}
  \end{footnotesize}
\end{algorithm}

Stopping at random with success probabilities $w^k$ has the same expectation as multiplying by $w^k$, so the expected surcharge $\Delta^k(s,a):=2H\,\E[\sigma_h^k-h\mid s_h^k=s,a_h^k=a]$ is the discounted cost-to-go \eqref{eq:vz} of an MDP in which every step costs $2H$. All policies have the same value in that MDP, so adding \cref{lem:identity} for it to the performance difference lemma for $c^k$ gives the counterpart of \eqref{eq:decomp}.

\begin{lemma}[Surcharge decomposition; \cref{lem:agg_identity}]\label{lem:agg_identity_main}
Let 
\begin{itemize}
    \item $\widetilde v^k(s,a):=\pre^k(s)+Q^k(s,a)+\Delta^k(s,a),$
    \item $\widetilde r^k:=(1-w^k)\Delta^k,$
    \item $\widetilde z^k:=\widetilde v^k-\widetilde r^k,$
    \item $\widetilde\Phi^k(s):=\ip{\pi^k(\cdot\mid s)}{\widetilde z^k(s,\cdot)}-\ip{\pi^\star(\cdot\mid s)}{\widetilde v^k(s,\cdot)}$.
\end{itemize}
Then
\[
  \Reg=\sum_{s\in\Sset}q^\star(s)\sum_{k=1}^K\widetilde\Phi^k(s)+\sum_{k=1}^K\sum_{s,a}q^k(s,a)\,\widetilde r^k(s,a).
\]
\end{lemma}
In the above, $2H\le\Delta^k\le 2H^2$, and $\widetilde\Phi^k(s)\le -H$ if $s$ is inactive in episode $k$. If $s\in\Sset_h$ is active with block $b\ni k$, the mean of $Y_h^k$ given a visit to $(s,a)$ is $\pre^k(s)+Q^k(s,a)+f_b\,\Delta^k(s,a)$.

The surcharge thus inherits the bandit analysis, including the optimism at inactive states, where the coin always fails. Because it is at least $2H\ge 2C^k$, discounting it alone yields a multiplicative margin of order $\epsilon_b$ for the whole return (\cref{lem:agg_margins}), while $C^k$ keeps unit weight so that $\pre^k(s)$ cancels in $\widetilde\Phi^k$. The larger margin pays for the implicit-exploration shrinkage of the undiscounted $C^k$, and the coins can be replaced by their conditional expectation given the trajectory (\cref{rem:derandomize}). The price is the range $B$ of the returns, which costs a factor $H$ in the regret. We set $a_0=\Theta(\eta BD)$, $\eta=\Theta(\sqrt{A/(HBDSK)})$ and $\gamma_0=\Theta(\sqrt{B/(AHSK)})$, with $M$ and $D$ as in \cref{tab:params_main} and $\Lmax$ the largest power of two not exceeding $1/(4\kagg)$, where $\kagg:=a_0+5A\gamma_0$ (for constants see \cref{tab:agg_params}).

\begin{theorem}[Tabular AMDPs with aggregate feedback]\label{thm:agg_main}
In the setting of \cref{thm:main} with aggregate feedback, for every $\delta\in(0,1)$, \textup{\aggalg{}} with the parameters of \cref{tab:agg_params} satisfies $\Reg = O\bigl(H^{5/2}\sqrt{D\,SAK}\,J\bigr)=\tO(H^{11/4}\sqrt{SAK})$ with probability at least $1-\delta$, with the memory and running time of \cref{thm:main}.
\end{theorem}

The bound improves on the leading term $\tO(H^2S\sqrt{AK})$ of \citet{lancewicki2025near} whenever $S>H^{3/2}$ and has no additive term. The proof (\cref{app:aggregate}) repeats the three steps of \cref{sec:analysis} at scale $B$, starting from \cref{lem:agg_identity_main}. Within this argument the order $H^2$ of the surcharge cannot be reduced (\cref{prop:agg_barrier}).

%% file: sections/6_linear.tex
\section{Extension to Linear AMDPs}\label{sec:linear_block}

Block self-normalization extends to linear function approximation. In a \emph{linear AMDP} of dimension $d$ \citep{jin2020provably,luo2021policy} the state space may be infinite, and there is a known feature map $\phi:\Sset\times\Aset\to\R^d$ such that the transition probabilities and the costs are linear in $\phi(s,a)$ on every layer (\cref{def:linear}). The cost-to-go of any policy is then linear in the features, and the unknown feature covariance under $\pi^k$ takes the role of the visitation probability. The full algorithm \linblkalg{} (\cref{alg:linear_block}), its parameters and the proof are in \cref{app:linear_block}.

\paragraph{Algorithm.}\label{sec:lin_blk_design}
\linblkalg{} applies block self-normalization to features. Each period builds regions (\emph{shells}) of the feature space of each layer from the features of the previous period, by the region construction of \citet{kulkarni2026introducing}; the empirical covariance of a shell whitens its features. As in their algorithm, a bonus is subtracted along the sampled trajectory up to the first feature outside all shells, where we cut the return, and policies are exponential weights. The estimator and its analysis are our contribution. A shell plays the role of an active state: it receives blocks of length $\Lambda$ inversely proportional to its empirical mass, with $\tTheta(d)$ visits each. Let $X_k$ indicate that episode $k$ visits the shell at its layer $h$, with whitened feature $u_k$ and return $Y_h^k$. At the end of a block $b$ the algorithm forms the least-squares estimate
\[
  \what_b:=\acc_b\,\Lambda\,G_b^\dagger\sum_{k\in b}X_ku_kY_h^k,\quad G_b:=\sum_{k\in b}X_ku_ku_k^\top,
\]
where the Gram matrix $G_b$ takes the place of the visit count $N_b$ in \eqref{eq:estimator} and $\acc_b\in\{0,1\}$ discards a block with an ill-conditioned $G_b$ or unusually many visits. The loss of a pair is updated only when a block of its shell ends; the policy at a state, whose actions may lie in different shells, is not frozen. As in the tabular case, $\what_b$ is bounded on every sample path and unbiased for a \emph{sum} of targets that change from one episode to the next: if the visits $(X_k,u_k)_{k\in b}$ are exchangeable, then $\E[\acc_b\Lambda G_b^\dagger\sum_{k\in b}X_ku_k\ip{u_k}{\xi_k}]=\Prob(\acc_b=1)\sum_{k\in b}\xi_k$ for all fixed vectors $\xi_k$ in the range of the shell covariance (\cref{lem:exch_matrix_sn}). Under drifting policies we bound the bias through a coupling with exchangeable visits (\cref{lem:lin_blk_bias}). This bound is additive, not relative, which is why \linblkalg{} keeps additive bonuses.

\begin{theorem}[Linear AMDPs with bandit feedback]\label{thm:linear_block}\label{sec:lin_blk_results}
In any $d$-dimensional linear AMDP (\cref{def:linear}) with unknown transitions and bandit feedback, as in \cref{sec:setting}, for every $\delta\in(0,1)$, \textup{\linblkalg{}} with the parameters \eqref{eq:lin_blk_params} of \cref{app:linear_block} runs in time polynomial in $d,H,A,K$ and satisfies, with probability at least $1-\delta$,
\[
  \Reg = \tO\Bigl(\bigl(d^{3/2}H^{11/4} + d^{7/4}H^{9/4}\bigr)\sqrt{K}\Bigr).
\]
\end{theorem}

The proof balances the regret of exponential weights, $\tO(\sqrt d\,H^{5/2}\Lmax)$ for periods of length $\Lmax$, against the cost of optimism, $\tO((d^{5/2}H^3+d^3H^2)K/\Lmax)$, with $\Lmax=\tTheta\bigl(d\sqrt{(H^{1/2}+\sqrt{d/H})K}\bigr)$. The bound improves the $\tO(d^3H^4\sqrt K)$ stated by \citet{kulkarni2026introducing} by a factor $d^{3/2}H^{5/4}$ when $H\ge\sqrt d$ and by $d^{5/4}H^{7/4}$ otherwise, and it is smaller for all $d$ and $H$ than the $\tO((d^2H^4+d^3H^3)\sqrt K)$ that their proof states before simplifying. The improvement comes from the choice of the period length and from the block estimate: the pathwise bound of the estimate and \cref{lem:pathlaw_informal} allow a larger step size, disjoint blocks save a factor $d$ in the concentration terms, and the cost of optimism is smaller. \cref{app:lin_compare} gives details, also for other window lengths.

%% file: sections/7_related_discussion.tex
\section{Discussion}\label{sec:discussion}

Our method achieves optimal dependence on $S$, $A$, and $K$ across all settings—improving on the linear-$S$ scaling of prior aggregate-feedback bounds with unknown transitions \citep{lancewicki2025near,ito2025adapting} alongside the bandit-feedback guarantees surveyed in \cref{sec:intro}—while leaving sub-optimal factors in $H$ and $d$ of $H^{3/4}$ (bandit feedback), $H^{5/4}$ (aggregate feedback), and $\sqrt{d}\,H^{5/4}$ (linear AMDPs for $H \ge \sqrt{d}$). Methodologically, our self-normalized importance sampling adapts off-policy evaluation \citep{precup2000eligibility,swaminathan2015self} to remove an unknown normalizer while bounding relative bias, and averaging importance-weighted returns over frozen blocks parallels epoch-based log-barrier updates for mixing ergodic average-reward MDPs \citep{wei2020model} (see \cref{app:related} for a fuller review). Incorporating regularized $Q$-functions \citep{li2026closing} or variance-aware bonuses \citep{schlisselberg2026improved} offers a direct path to tighter horizon factors, as does extending block self-normalization from oblivious \citep{kulkarni2026introducing,lancewicki2025near,liu2026computationally} to adaptive adversaries \citep{jin2020learning,li2026closing} (\cref{rem:adaptive}). Minor caveats include unoptimized constants (\cref{rem:constants}), requiring $S$ or $d$ in advance while using doubling for unknown $K$ (\cref{rem:doubling}), and leaving delayed feedback and aggregate feedback in linear AMDPs for future study (\cref{prop:lin_agg_barrier})\label{sec:lin_agg_barrier}.

%% file: appendix/0_overview.tex
\section*{Appendix Overview}

The appendices contain complete proofs of all results. After the related work, they follow the order of the main text.
\begin{itemize}[leftmargin=*,nosep,topsep=4pt plus 2pt,itemsep=3pt plus 1pt]
\item \textbf{Additional related work.} \cref{app:related}, including a step-by-step comparison with \citet{kulkarni2026introducing} (\cref{app:compare}).
\item \textbf{Tabular AMDPs with bandit feedback (\alg{}, \cref{sec:algorithm,sec:analysis}).} \cref{fig:roadmap} shows the dependencies between the lemmas. \cref{app:notation} lists the parameters and elementary inequalities. \cref{app:identity} proves the regret decomposition (\cref{lem:identity}). \cref{app:omd} proves the stability and regret bounds of log-barrier OMD (\cref{lem:barrier}). \cref{app:drift} bounds the drift of policies, occupancies and targets within a block and across two periods (\cref{lem:dyadic,lem:pathlaw,lem:drift}). \cref{app:selfnorm} proves the relative bias bound for self-normalization (\cref{lem:relative_sn}). \cref{app:classification} proves the classification of active and inactive states (\cref{lem:classification}). \cref{app:events} sets up the stopping-time filtration, the concentration inequalities and the good event $\good$. \cref{app:local} bounds the local regret of each state (\cref{lem:local}), and \cref{app:main_proof} combines everything into \cref{thm:main} and discusses the constants (\cref{rem:constants}). \cref{app:extensions} covers regret on realized costs, random initial states, the running time, inexact OMD updates, an unknown number of episodes (\cref{rem:doubling}) and adaptive adversaries (\cref{rem:adaptive}), and \cref{app:horizon} discusses the horizon exponent.
\item \textbf{Aggregate feedback (\aggalg{}, \cref{sec:aggregate}).} \cref{app:aggregate} gives the algorithm in full (\cref{alg:agg}), proves the surcharge decomposition (\cref{lem:agg_identity}) and \cref{thm:agg_main}, and proves \cref{prop:agg_transfer,prop:agg_barrier}.
\item \textbf{Linear AMDPs (\linblkalg{}, \cref{sec:linear_block}).} \cref{app:linear_block} defines linear AMDPs, gives the algorithm in full (\cref{alg:linear_block,alg:regions}) and proves the decomposition for shifted costs (\cref{lem:lin_decomp}), the properties of the shells (\cref{lem:lin_peeling_det}), the pathwise bounds and drift (\cref{lem:lin_blk_stab}), uniform convergence for the slowly drifting samples of a period (\cref{lem:lin_coupling,lem:lin_uc_drift,lem:lin_blk_good}), the matrix self-normalization identity and its bias under drift (\cref{lem:exch_matrix_sn,lem:lin_blk_bias}), the block supermartingales (\cref{lem:lin_supermart}), \cref{thm:linear_block}, the comparison with the bound of \citet{kulkarni2026introducing} (\cref{app:lin_compare}), and \cref{prop:lin_agg_barrier}.
\end{itemize}

\input{appendix/fig_roadmap}

%% file: appendix/fig_roadmap.tex
\begin{figure}[ht]
\centering
\begin{tikzpicture}[
  box/.style={draw,rounded corners=2pt,align=center,font=\footnotesize,inner sep=3pt,minimum height=0.9cm,minimum width=3.4cm},
  arr/.style={-{Stealth[length=2.2mm]},thick}]
  \node[box] (omd)   at (0,0)      {Log-barrier OMD\\(\cref{lem:barrier})};
  \node[box] (dyad)  at (4.0,0)    {Dyadic budget\\(\cref{lem:dyadic})};
  \node[box] (path)  at (8.0,0)    {Path-law domination\\(\cref{lem:pathlaw})};
  \node[box] (drift) at (12.0,0)   {Block drift\\(\cref{lem:drift})};
  \node[box] (conc)  at (0,-1.7)   {Concentration \& IX\\(\cref{lem:freedman,lem:ix})};
  \node[box] (sn)    at (6.0,-1.7) {Relative self-normalization\\(\cref{lem:relative_sn})};
  \node[box] (cl)    at (12.0,-1.7){Classification\\(\cref{lem:classification})};
  \node[box] (id)    at (0,-3.4)   {Regret decomposition\\(\cref{lem:identity})};
  \node[box] (local) at (6.0,-3.4) {Local regret\\(\cref{lem:local})};
  \node[box] (opt)   at (12.0,-3.4){Cost of optimism\\(\cref{eq:optcost})};
  \node[box,fill=blue!6,minimum width=4.4cm] (thm) at (6.0,-5.1) {Master inequality (\cref{prop:master})\\$\Rightarrow$ \cref{thm:main}};
  \draw[arr] (omd) -- (dyad);
  \draw[arr] (dyad) to[out=20,in=160] (drift);
  \draw[arr] (path) -- (drift);
  \draw[arr] (drift) -- (cl);
  \draw[arr] (drift) -- (sn);
  \draw[arr] (cl.south west) -- node[midway,above,sloped,font=\scriptsize]{$\Omega_b$ holds on $\Ecl$} (local.north east);
  \draw[arr] (sn) -- (local);
  \draw[arr] (conc.south east) -- (local.165);
  \draw[arr] (omd.south east) to[out=-25,in=145] (local.north west);
  \draw[arr] (cl) -- (opt);
  \draw[arr] (id.east) -- (local.west);
  \draw[arr] (id.south east) -- (thm.north west);
  \draw[arr] (local) -- (thm);
  \draw[arr] (opt.south west) -- (thm.north east);
\end{tikzpicture}
\caption{Logical dependency graph for \cref{thm:main} (\alg{}). Arrows point from each lemma to the results that invoke it. \cref{app:aggregate} adapts the same proof structure to aggregate bandit feedback (\aggalg{}) at return scale $B=2H^2+H$.}
\label{fig:roadmap}
\end{figure}
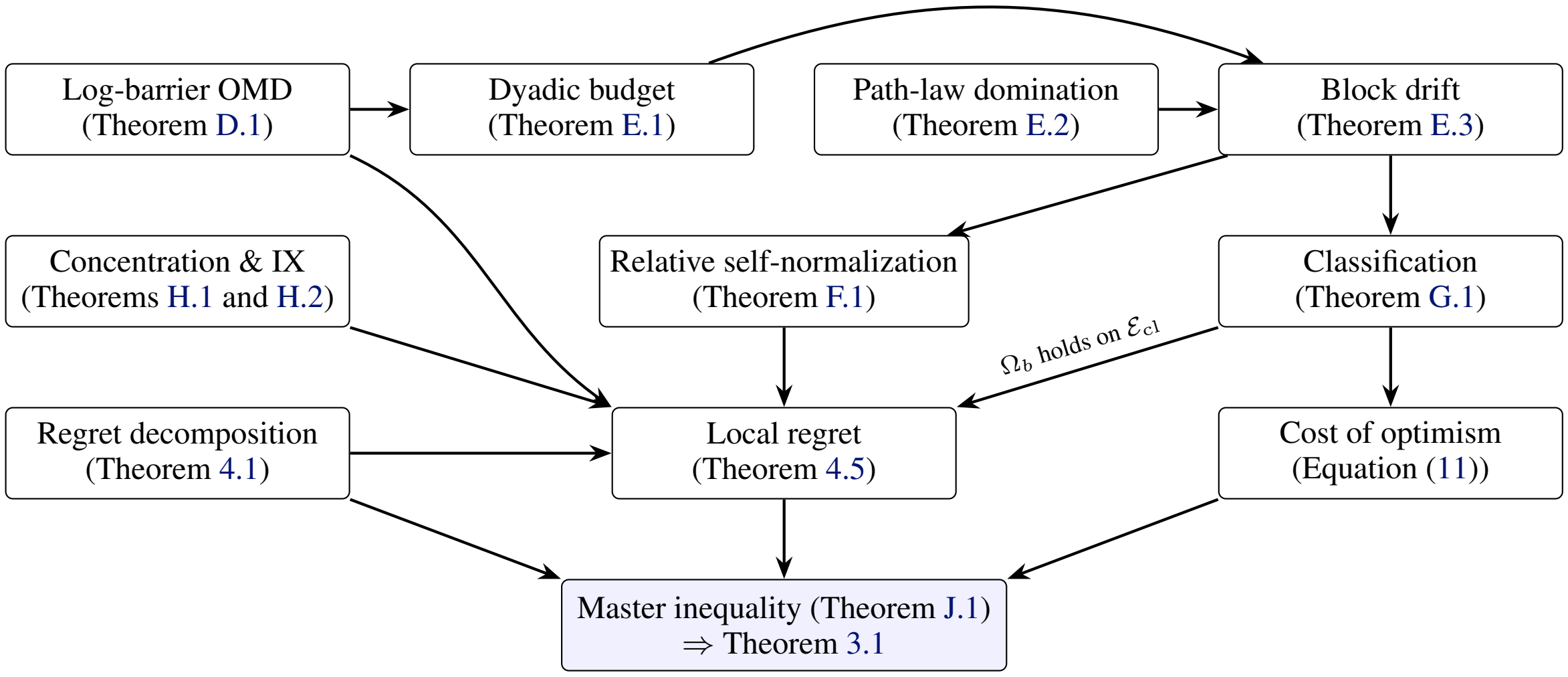

%% file: appendix/N_related.tex
\section{Additional Related Work}\label{app:related}

\paragraph{Tabular AMDPs.}
Online learning in MDPs with adversarial costs goes back to \citet{even2009online} and \citet{yu2009markov}; the layered episodic formulation used here is due to \citet{neu2010online,neu2012adversarial} and \citet{zimin2013online}. With known transitions, occupancy-measure methods attain $\tO(\sqrt{HSAK})$ regret under bandit feedback \citep{zimin2013online,jinluo2020simultaneously,lee2020bias}. With unknown transitions and bandit feedback, the rates improved from $K^{3/4}$ and $K^{2/3}$ \citep{rosenberg2019bandit,shani2020optimistic} to $\tO(HS\sqrt{AK})$ \citep{jin2020learning,jin2021best,dai2022follow}. Policy optimization reached the same dependence on $S$ with dilated bonuses \citep{luo2021policy,dann2023best} and with regularized $Q$-functions \citep{li2026closing}. In all of these works the occupancy of the current policy, or a bonus that depends on it, is computed from confidence sets of transition kernels, and the same holds for the works on delayed and cooperative feedback \citep{lancewicki2020learning,lancewicki2022cooperative,jin2022near,lancewicki2023delay,zierahn2025unified}. Under full-information feedback, \citet{rosenberg2019online} gave the first $\sqrt K$ bound with unknown transitions and the $\sqrt S$ dependence was obtained by \citet{tiapkin2025narrowing}; \citet{schlisselberg2026improved} recently obtained a leading term of $\tO(\sqrt{H^3SAK})$ with a variance-aware analysis of policy optimization, which also improves the horizon dependence of policy optimization under bandit feedback when the transitions are known. For bandit feedback and unknown transitions the $\sqrt S$ dependence was first obtained by \citet{kulkarni2026introducing}, whose relation to our method is discussed in \cref{sec:intro}. Their framework also covers AMDPs with delayed feedback, which we do not treat, and they state that they did not optimize the dependence of their bounds on $H$ and $d$.

\paragraph{Adversary models.}
\citet{jin2020learning} and \citet{li2026closing} allow the cost function of an episode to depend on the learner's algorithm and on the history of the previous episodes. \citet{kulkarni2026introducing}, \citet{lancewicki2025near} and the present paper assume that the cost functions are fixed in advance. \cref{rem:adaptive} shows where our analysis uses this; an extension to adaptive adversaries is left for future work.

\paragraph{Aggregate feedback.}
Learning from the total cost of an episode was first studied for stochastic costs \citep{efroni2021reinforcement,cassel2024near}. \citet{cohen2021online} initiated the study of AMDPs with this feedback and obtained $\poly(H,S,A)\sqrt K$ regret. With $S=H$ the problem contains bandit combinatorial optimization over paths \citep{cohen2017tight}. \citet{lancewicki2025near} obtained $\tO(H^{3/2}\sqrt{SAK})$ regret for known transitions, with a matching lower bound, and $\tO(H^2S\sqrt{AK}+HS^3A)$ for unknown transitions, in our convention for $S$. \citet{ito2025adapting} gave best-of-both-worlds bounds whose adversarial part is also linear in $S$ when the transitions are unknown.

\paragraph{Linear AMDPs.}
With a simulator or exploratory assumptions, linear AMDPs with bandit feedback were studied by \citet{neu2021online}; \citet{kong2024improved} obtained $K^{4/5}$ regret under an exploratory assumption. Without them, efficient algorithms improved from $K^{14/15}$ to $K^{3/4}$ regret \citep{luo2021policy,dai2023refined,sherman2023improved,liu2024towards}, while $\sqrt K$ regret was known only for an inefficient algorithm, with $\tO(d^{7/2}H^{7/2}\sqrt K)$ in expectation \citep{liu2024towards}, until the works of \citet{liu2026computationally} and \citet{kulkarni2026introducing}. The bound of \cref{thm:linear_block} is below $d^{7/2}H^{7/2}\sqrt K$ for all $d$ and $H$, up to logarithmic factors. Full-information feedback is studied by \citet{sherman2024rate} and \citet{cassel2024warm}, and linear mixture AMDPs by \citet{cai2020provably}, \citet{zhou2021nearly} and \citet{he2022near}. The lower bound $\Omega(dH^{3/2}\sqrt K)$ in \cref{tab:comparison} is that of \citet{zhou2021nearly}, who prove it for linear mixture MDPs and observe that their construction is also a linear MDP.

\paragraph{Frozen epochs and visit-averaged estimates.}
The closest antecedent of block self-normalization is the policy optimization method MDP-OOMD of \citet{wei2020model} for ergodic average-reward MDPs with a fixed, known reward function. It keeps the policy fixed during an epoch, estimates the value of each action at each state by averaging importance-weighted returns over the visits to the state within the epoch, and updates with log-barrier optimistic mirror descent, chosen for the stability of the policies. Its successor for linear function approximation \citep{wei2021learning} also executes a fixed policy in every epoch. Both rely on ergodicity: one epoch length, determined by the mixing and hitting times, serves all states, every state is visited many times in a typical epoch, and neither the policy nor the reward function changes inside an epoch. In the episodic adversarial setting none of this holds. The costs change in every episode, so the estimate must be unbiased for the sum of the targets of a block and not for a stationary value. States can be visited with arbitrarily small probability, so block lengths must adapt to the state, and rare states are left unestimated. And because states with shorter blocks keep updating inside the blocks of the others, the estimate is biased, which is what the nested blocks, the relative bias bound and the margins address. Decisions that are constant over batches of rounds also appear in adversarial bandits against adaptive adversaries, where the average loss of a mini-batch is fed to a base algorithm \citep{arora2012online}, and are studied under the name of batched bandits \citep{perchet2016batched}. There one batch length serves all arms; here every state has its own block length, set from data, and the blocks of different states interact.

\paragraph{Techniques.}
Self-normalized (weighted) importance sampling is classical in off-policy evaluation \citep{precup2000eligibility,mahmood2014weighted,swaminathan2015self}, where it trades a small bias for a lower variance, and ratio estimators of this kind are classical in survey sampling \citep{sarndal1992model}, where the self-normalized form of the Horvitz--Thompson estimator is known as the H\'ajek estimator. We use the term in this sense; it is unrelated to self-normalized martingale inequalities \citep{abbasi2011improved}. We use it for a different purpose, to remove an unknown normalizer, and we control its bias in relative terms. Implicit exploration \citep{kocak2014efficient,neu2015explore} and the log-barrier regularizer \citep{foster2016learning,agarwal2017corralling,wei2018more,lee2020bias} are standard tools for high-probability bounds and multiplicative stability in adversarial bandits. Updating a state only at the end of a stage also appears in model-free reinforcement learning with stochastic rewards \citep{jin2018q,bai2019provably,zhang2020almost,li2021breaking}, where the stage lengths grow with the visit count of a state-action pair; here the block lengths are set from the visit frequencies of the previous period and are aligned across states so that blocks are nested. Truncating the continuation value at insufficiently visited states follows the known/unknown state distinction of R-max \citep{brafman2002rmax}.

\subsection{Comparison with \texorpdfstring{\citet{kulkarni2026introducing}}{Kulkarni et al. (2026)}, Step by Step}\label{app:compare}

\cref{tab:compare} lists the ingredients of the two tabular algorithms and of their analyses side by side. The entries for \citet{kulkarni2026introducing} are taken from the overview and the tabular analysis in their Sections~3 and~4 and from the proofs in their Appendix~A.

\begin{table}[ht]
\caption{The tabular algorithm and analysis of \citet{kulkarni2026introducing} and \alg{}, ingredient by ingredient. Constants and logarithmic factors are omitted.}
\label{tab:compare}
\begin{center}
\begin{footnotesize}
\renewcommand{\arraystretch}{1.15}
\begin{tabular}{@{}p{0.19\textwidth}p{0.37\textwidth}p{0.37\textwidth}@{}}
\toprule
 & \citet{kulkarni2026introducing} & \alg{}\\
\midrule
Estimator & inverse visitation probability $1/q^{\pi_k}(s,a)$, by geometric resampling with rejection on the trajectories of later episodes & block sum of the cost-to-go, normalized by the visit count of the block\\
Policy during the evaluation & changes in every episode & action distribution of the evaluated state is frozen; all other states keep updating\\
Update & exponential weights in every episode, on pieces of the pending estimates & log-barrier OMD at the end of each block\\
Change of $\log\pi(a\mid s)$ & $\eta H^2$ per episode & $\eta H\Lambda$ per block of length $\Lambda$, that is, $\eta H$ per episode\\
Closeness of policies needed & likelihood ratio at most $2$ on every partial trajectory & trajectory laws within a factor $e^{\pm\varrho}$ outside an event of probability $e^{-20J}$\\
Trajectory factor & $H$ & $D=\min\{H,\sqrt{HJ}\}$\\
Estimated when & $q^{\pi_k}(s,a)\gtrsim\eta H^3$ & $q^k(s)\gtrsim M/\Lmax\approx\eta HD$\\
Optimism & bonus $Q$-function of range $H^2$, by dynamic programming under the most optimistic transition of a confidence set & discount inside the sampled returns, which stay in $[0,H]$; returns cut at inactive states\\
Transition model & confidence sets, for thresholds and bonuses & none\\
Cost of optimism & $\eta H^4SAK$ & $\eta H^2D\,SK$\\
Regret & $H^{5/2}\sqrt{SAK}+H^4S^2A$ & $H^{3/2}\sqrt{D\,SAK}$\\
Delayed feedback & covered & not covered\\
\bottomrule
\end{tabular}
\end{footnotesize}
\end{center}
\end{table}

\paragraph{Where the horizon factors differ.}
In both analyses the regret is, up to logarithmic factors, the sum of a penalty term of order $H/\eta$ (times $A$ for the log-barrier) and a cost of optimism of order $H$ times the estimation threshold times the number of state-action pairs or states times $K$. In both, the threshold is the change of the log-policy per episode times the trajectory factor. In their analysis this gives $H/\eta+\eta H^4SAK$, hence $H^{5/2}\sqrt{SAK}$; in ours it gives $HA/\eta+\eta H^2D\,SK$, together with the analogous pair of terms in $\gamma_0$, hence $H^{3/2}\sqrt{D\,SAK}$. The two differences are the change per episode ($\eta H^2$ against $\eta H$) and the trajectory factor ($H$ against $D$).

\paragraph{Which ingredient gives what.}
\emph{Optimism inside the returns.} The block estimate satisfies $\ip{\pi_b}{\widehat\ell_b}\le H\Lambda_b$ on every sample path, because the returns of the block are averaged over its visits, and with the log-barrier this bounds the change of the log-policy by $6\eta H$ per episode of the block. In their algorithm the loss received in an episode lies in $[-H^2,\tO(H)]$: the pieces of the pending estimates contribute $\tO(H)$, and the term of order $H^2$ is the bonus $Q$-function, which adds bonuses of size up to $H$ over the remaining layers (their Lemma~A.5). \alg{} has no bonus function, because its optimism sits inside the returns, as a discount at active states and a cut at inactive ones. This difference in how optimism is implemented, and not the estimator, accounts for the first factor $\sqrt H$. The estimator enters indirectly. Its bias is relative (\cref{lem:relative_sn}), so a multiplicative margin inside the returns suffices to pay for it. Up to its additive term, their bound is $\tO(H/\eta)+\tO(\eta H^3(\nu+H)SAK)$ with $\nu=\tO(1)$ (display~(43) in the proof of their Theorem~A.3). The factor $\nu+H$ is that of their threshold $\theta=\tTheta(\eta H^2(\nu+H))$, and its summand $H$ reflects the range $H^2$ of the bonus $Q$-function (their Lemma~A.5). An optimism device of range $O(H)$ in their algorithm would therefore plausibly give $\tO(H^2\sqrt{SAK})$, which is also the bound of \cref{thm:main} with $D=H$. We have not carried out this modification.

\emph{Trajectory factor.} \cref{lem:pathlaw} is, in substance, the advanced composition theorem of differential privacy \citep{dwork2010boosting} and is not new. \cref{lem:relative_sn} tolerates an additive error in its hypothesis \ref{H:target}, so it can be combined with \cref{lem:pathlaw}, whose domination holds only outside an event of small probability. Geometric resampling with rejection accepts a later trajectory with a probability proportional to a likelihood ratio. Their algorithm clips this probability at one, and their analysis chooses the parameters so that the clipping never occurs. Its guarantee (their Lemmas~3.1 and~A.4) assumes that the ratio is at most $2$ on every partial trajectory. Under the bound of \cref{lem:pathlaw} the clipping occurs with probability at most $e^{-y}$ and lowers the success probability of an attempt by at most $e^{-y}/2$. We have not checked whether the rest of their analysis, which uses that every attempt succeeds with the same probability whatever the history (their Lemmas~A.6 to~A.8), tolerates this perturbation. If it does, and if the bonus is modified as above, the closeness condition of their analysis (the fourth inequality of their Eq.~(38)) would allow a threshold of order $\eta HD$, as in ours. Their threshold is, however, also bounded below by a multiple of $\nu\eta H^2$ through the first and third inequalities of their Eq.~(38), which keep the bonus of an estimated pair below $H$ and control the concentration of the numbers of attempts (their Lemmas~A.5 and~A.9), and the corresponding part of their bonus (their Eqs.~(39) and~(40)) contributes a term of order $\eta H^3SAK$ to their bound, up to logarithmic factors. Both come from the random stops and from second-order terms of their estimates. With the two modifications alone, their analysis would therefore still give $\tO(H^2\sqrt{SAK})$. We have not examined whether these terms can be reduced, so we do not know whether their estimator can reach the exponent $7/4$.

\emph{The estimator.} Independently of the exponent of $H$, block self-normalization needs no transition model, no auxiliary randomization and no bookkeeping of pending estimates, and its bound has no additive term. Its estimate is bounded on every sample path and has a relative bias, which is what makes optimism inside the returns applicable. For the trajectory bound a second property matters: apart from the bias caused by drift, every error of a block is of order $\eta H\Lambda_b$ times the block's own loss (\cref{tab:budget}), so the factor $D$ of the drift is the only one that enters the margin and hence the threshold, whereas their threshold is tied to $\eta H^2$ by other terms, as described above. Their linear algorithm already replaces confidence sets by empirical frequencies, and the same may be possible in their tabular algorithm.

\emph{Aggregate feedback.} We are not aware of an analysis of geometric resampling under aggregate feedback. Applying it to the episode total in the manner of \citet{lancewicki2025near} is a natural candidate for another $\sqrt S$ bound in that setting, and we do not know which power of $H$ it would give.

%% file: appendix/A_notation.tex
\section{Notation and Parameters for \alg}\label{app:notation}

\begin{table}[ht]
\caption{Parameters of \alg{} ($S=|\Sset|=\sum_{h=1}^H|\Sset_h|\ge H$, $A=|\Aset|\ge 2$, $H\ge 1$, $K\ge 2$, $\delta\in(0,1)$; all logarithms are natural).}
\label{tab:params}
\begin{center}
\begin{footnotesize}
\setlength{\tabcolsep}{4pt}
\begin{tabular}{lll}
\toprule
Symbol & Value & Role\\
\midrule
$J$ & $\log(64SAK^3/\delta)$ & log-confidence factor ($J>\log 1024 > 6.93$)\\
$M$ & $\lceil 200J\rceil$ & target visit count per active block\\
$D$ & $\min\{H,\sqrt{HJ}\}\ge 1$ & horizon drift factor (\cref{lem:drift})\\
$\etail$ & $e^{-20J}$ & additive path-law tail (analysis only)\\
$C_0,\ C_1$ & $1000,\ 24$ & margin and cost-of-optimism constants\\
$\aeta,\ \beeta$ & $2HAJ,\ \ C_1C_0MH^2DSK$ & coefficients of $1/\eta$ and $\eta$ (\cref{prop:master})\\
$\agam,\ \bgam$ & $H^2J/2,\ \ C_1AMHSK$ & coefficients of $1/\gamma_0$ and $\gamma_0$ (\cref{prop:master})\\
$\eta$ & $\sqrt{\aeta/\beeta}=\sqrt{2AJ/(C_1C_0MHDSK)}$ & exact tuned OMD step size\\
$\gamma_0$ & $\sqrt{\agam/\bgam}=\sqrt{HJ/(2C_1AMSK)}$ & exact tuned implicit-exploration rate\\
$a_0$ & $C_0\eta H D$ & multiplicative margin rate\\
$\kappa$ & $a_0+A\gamma_0$ & total discount rate (play $\Unif(\Aset)$ if $\kappa>1/4$)\\
$\Lmax$ & $2^{\lfloor\log_2(1/(4\kappa))\rfloor}\in(\frac1{8\kappa},\frac1{4\kappa}]$ ($\kappa\le 1/4$) & dyadic period length\\
$\Lambda_j(s)$ & \cref{eq:blocklength}, $M\le\Lambda_j(s)\le\Lmax$ & block length of active $s\in\Sset$ in period $j$\\
$\epsilon_j,f_j,\gamma_j$ & $a_0\Lambda_j,\ 1-\epsilon_j,\ \gamma_0\Lambda_j$ & block margin, survival, and IX parameter\\
\bottomrule
\end{tabular}
\end{footnotesize}
\end{center}
\end{table}

\paragraph{Summary of analytical notation.}
Fix a state $s\in\Sset_h$ ($h=h(s)\in[H]$) and write $X_k=\ind\{s_h^k=s\}$ for its visit indicator in episode $k$, so that $\E[X_k\mid\F_{k-1}]=q^k(s)$. For an active block $b$ of $s$, we write $\Lambda_b$ for its length, $\pi_b$ for its frozen local policy $\pi(\cdot\mid s)$, $\epsilon_b=a_0\Lambda_b$ for its margin, $f_b=1-\epsilon_b$, $\gamma_b=\gamma_0\Lambda_b$, $\chi_b(a)=\pi_b(a)/(\pi_b(a)+\gamma_b)$, and $\tau_b=\eta H\Lambda_b$ (the subscripted $\tau_b$ is a stability ratio and is unrelated to the trajectory $\tau^k$). This gives $w^k(s,a)=f_b^2\chi_b(a)$ for all $k\in b$. Depending on context, the same block quantities are indexed by period $j$ and state $s$ as $\Lambda_j(s)$, by a specific block $b$ as $\Lambda_b$, or by the $i$th complete active block of a fixed state $s$ as $\Lambda_i$ (\cref{app:events}). The analytical functions $v^k,z^k,V^{z,k},r^k,\Phi^k$ are defined in \cref{eq:vz,eq:Bresidual}. We also write $B^k=Q^k-z^k$ and $V^{B,k}(s)=\ip{\pi^k(\cdot\mid s)}{B^k(s,\cdot)}$. The drift radii are $\rho_b^0=76\eta H D\Lambda_b$, $\varrho_b=80\eta H D\Lambda_b$, and $\rho_P=76\eta H D\Lmax$. The block bias parameters are $\beta_b=5\varrho_b$ and $e_b=\Lambda_b H(3\etail+e^{-M/3})$. At $s$, the smoothed comparator distribution is $u=(1-\frac1K)\pi^\star(\cdot\mid s)+\frac1K\Unif(\Aset)$.

\paragraph{Elementary parameter inequalities.}
Assume $\kappa\le 1/4$. Because $S\ge H\ge 1$, $A\ge 2$, $K\ge 2$, and $\delta\in(0,1)$, we have $64SAK^3/\delta > 1024$ and hence $J>\log 1024 > 6.93$. Also $64AK^3 \le \frac14 A^6 K^6$ since $A^5 K^3 \ge 256$, so $J < 6\log(SAK/\delta)$. The following consequences hold:
\begin{enumerate}[label=(P\arabic*),leftmargin=*]
\item\label{P:margin} For every active $s\in\Sset$ in period $j\ge 2$, $\epsilon_j(s)=a_0\Lambda_j(s)\le a_0\Lmax\le\kappa\Lmax\le 1/4$, so $f_j(s)\in[3/4,1)$. Likewise, $A\gamma_j(s)\le\kappa\Lmax\le 1/4$.
\item\label{P:stab} Since $D\ge 1$, $\eta H\Lmax\le\eta H/(4\kappa)\le\eta H/(4C_0\eta H D)=1/(4C_0 D)\le 1/(4C_0)=2.5\cdot 10^{-4}$. Thus $\tau_b=\eta H\Lambda_b\le 2.5\cdot 10^{-4}$ for every block length $\Lambda_b\le\Lmax$.
\item\label{P:radius} $\varrho_b=80\eta H D\Lambda_b = 80\epsilon_b/C_0 = 2\epsilon_b/25 \le 0.02$, $\rho_b^0 = 76\eta H D\Lambda_b \le \varrho_b$, and $\beta_b=5\varrho_b=2\epsilon_b/5\le 0.1$. Since $(e^{4t}-1)/t$ increases on $(0,\infty)$ and $(e^{0.08}-1)/0.02 < 4.17 < 5$, we also have $\beta_b\ge e^{4\varrho_b}-1$.
\item\label{P:slack} $\tau_b=\eta H\Lambda_b = \epsilon_b/(C_0 D) \le \epsilon_b/1000$, since $\epsilon_b = C_0\eta H D\Lambda_b$ and $D\ge 1$.
\item\label{P:tiny} $\etail=e^{-20J}$ satisfies $12K^2\etail \le 12 K^2 (64SAK^3)^{-20} < 10^{-55}$, and $e^{-M/3}\le\etail$ since $M\ge 200J$.
\end{enumerate}

%% file: appendix/B_identity.tex
\section{Proof of the Regret Decomposition}\label{app:identity}

We prove the conditional-mean identity stated after \cref{eq:vz} and \cref{lem:identity}. Fix an episode $k\in[K]$, omit superscript $k$ when clear from context, and recall from \cref{sec:setting} that $(p g)(s,a)=0$ for $s\in\Sset_H$. By the episodic performance difference lemma \citep{kakade2002approximately}, for any Markov policy $\pi$ and cost function $c$,
\begin{equation}\label{eq:pdl}
  V^\pi(s_1;c) - V^{\pi^\star}(s_1;c) = \sum_{s\in\Sset} q^\star(s)\ip{\pi(\cdot\mid s) - \pi^\star(\cdot\mid s)}{Q^\pi(s,\cdot;c)}.
\end{equation}

\paragraph{Conditional mean of sampled returns.}
Call $(s,a)\in\Sset_h\times\Aset$ \emph{reachable} in episode $k$ if $\Prob(s_h^k=s,a_h^k=a\mid\F_{k-1})>0$. Every local policy $\pi^k(\cdot\mid s)$ has full support, since it starts uniform and \cref{lem:barrier} preserves full support. Thus $(s,a)$ is reachable if and only if $q^k(s)>0$, and for $h<H$ any $s'\in\Sset_{h+1}$ with $p(s'\mid s,a)>0$ also satisfies $q^k(s')>0$. We verify $\E[Y_h^k\mid\F_{k-1},s_h^k=s,a_h^k=a] = v(s,a)$ for all reachable $(s,a)\in\Sset_h\times\Aset$ by backward induction on $h$. At $h=H$, given $s_H^k=s,a_H^k=a$, we have $Y_H^k = c(s,a) = v(s,a)$. For $h<H$, \cref{eq:sample_returns} and the Markov property under the $\F_{k-1}$-measurable policy $\pi$ give
\begin{align*}
  \E[Y_h^k\mid\F_{k-1},s_h^k=s,a_h^k=a]
  &= c(s,a) + \sum_{s'\in\Sset_{h+1}:\,p(s'\mid s,a)>0} p(s'\mid s,a)\sum_{a'\in\Aset} \pi(a'\mid s')\,w(s',a')\\
  &\qquad\qquad\qquad\qquad\qquad\qquad\cdot \E[Y_{h+1}^k\mid\F_{k-1},s_{h+1}^k=s',a_{h+1}^k=a'],
\end{align*}
where $w(s',a')$ is $\F_{k-1}$-measurable and each $(s',a')$ in the sum is reachable. Applying the induction hypothesis turns the right-hand side into $c(s,a) + (p V^z)(s,a) = v(s,a)$.

\paragraph{Ordering and range bounds.}
We show $0\le z(s,a)\le v(s,a)\le Q(s,a)\le H-h+1\le H$ for $s\in\Sset_h$ by backward induction on $h$. At $h=H$, $0\le v = c = Q \le 1$ since $c\in[0,1]$, and $0\le z = w v \le v$ since $w\in[0,1]$. Given $0\le V^z \le V$ on $\Sset_{h+1}$, we have $0\le v = c + p V^z \le c + p V = Q \le H-h+1$ on $\Sset_h$, $0\le z = w v \le v$, and averaging under $\pi(\cdot\mid s)$ gives $0\le V^z \le V$ on $\Sset_h$. Thus $r = (1-w)v \in [0,H]$ and $B := Q - z \in [0,H]$.

\paragraph{Residual recursion.}
Define $V^B(s) = V(s) - V^z(s) = \ip{\pi(\cdot\mid s)}{B(s,\cdot)}$. Then
\[
  B = Q - z = (Q - v) + (v - z) = p(V - V^z) + r = r + p V^B.
\]
Unrolling this linear recursion forward from $s_1\in\Sset_1$ under $\pi$ yields
\begin{equation}\label{eq:VB_unroll}
  V^B(s_1) = \sum_{s\in\Sset} q^\pi(s)\ip{\pi(\cdot\mid s)}{r(s,\cdot)}.
\end{equation}

\paragraph{Assembling the decomposition.}
Substituting $Q = z + B$ into \cref{eq:pdl} gives
\begin{align*}
  V^\pi(s_1;c) - V^{\pi^\star}(s_1;c)
  &= \sum_{s\in\Sset} q^\star(s)\bigl(\ip{\pi(\cdot\mid s)}{z(s,\cdot)} - \ip{\pi^\star(\cdot\mid s)}{z(s,\cdot)}\bigr)\\
  &\quad + \sum_{s\in\Sset} q^\star(s)\bigl(V^B(s) - \ip{\pi^\star(\cdot\mid s)}{B(s,\cdot)}\bigr).
\end{align*}
Using $B = r + p V^B$ and the occupancy flow identity $\sum_{s\in\Sset_h} q^\star(s)\sum_{a\in\Aset}\pi^\star(a\mid s)p(s'\mid s,a) = q^\star(s')$ for $s'\in\Sset_{h+1}$ ($h\in[H-1]$), the second sum telescopes across layers:
\begin{align*}
  &\sum_{h=1}^H\sum_{s\in\Sset_h} q^\star(s)V^B(s) - \sum_{h=1}^{H-1}\sum_{s'\in\Sset_{h+1}} q^\star(s')V^B(s')
  - \sum_{s\in\Sset} q^\star(s)\ip{\pi^\star(\cdot\mid s)}{r(s,\cdot)}\\
  &\qquad = V^B(s_1) - \sum_{s\in\Sset} q^\star(s)\ip{\pi^\star(\cdot\mid s)}{r(s,\cdot)},
\end{align*}
since $\Sset_1=\{s_1\}$ and $q^\star(s_1)=1$. Combining $-\ip{\pi^\star}{z} - \ip{\pi^\star}{r} = -\ip{\pi^\star}{v}$ produces $\Phi(s) = \ip{\pi(\cdot\mid s)}{z(s,\cdot)} - \ip{\pi^\star(\cdot\mid s)}{v(s,\cdot)}$. Summing over $k=1,\dots,K$ and inserting \cref{eq:VB_unroll} proves \cref{eq:decomp}. When $s$ is inactive, $w(s,\cdot)\equiv 0$, so $z(s,\cdot)\equiv 0$ and $\Phi(s) = -\ip{\pi^\star(\cdot\mid s)}{v(s,\cdot)} \le 0$. \qed

%% file: appendix/C_omd.tex
\section{Log-Barrier Mirror Descent}\label{app:omd}

\begin{lemma}[Multiplicative stability and regret of log-barrier OMD]\label{lem:barrier}
Let $\pi\in\Delta(\Aset)$ have full support, let $\eta>0$, let $\ell\in\R^{\Aset}$, and set $\bar\ell=\sum_{a\in\Aset}\pi(a)|\ell(a)|$. With $\psi(p)=-\sum_{a\in\Aset}\log p(a)$, the minimizer $\pi'$ of $p\mapsto\eta\ip{p}{\ell}+D_\psi(p,\pi)$ over $\Delta(\Aset)$ exists, is unique, and has full support. If $\eta\bar\ell\le 1/32$, then
\begin{equation}\label{eq:stability}
  \Bigl|\log\frac{\pi'(a)}{\pi(a)}\Bigr| \le 6\eta\bar\ell \qquad \text{for every } a\in\Aset.
\end{equation}
For any $n\ge 1$, any sequence $\pi_1=\Unif(\Aset),\pi_2,\dots,\pi_{n+1}$ generated by such updates on losses $\ell_1,\dots,\ell_n$ satisfying $\eta\bar\ell_i\le 1/32$, and any comparator $u\in\Delta(\Aset)$ with $\min_a u(a)\ge 1/(KA)$,
\begin{equation}\label{eq:omd_bound}
  \sum_{i=1}^n \ip{\pi_i - u}{\ell_i} \le \frac{A\log K}{\eta} + 2\eta\sum_{i=1}^n\sum_{a\in\Aset} \pi_i(a)^2 \ell_i(a)^2.
\end{equation}
Both bounds hold deterministically for arbitrary (possibly adaptive) loss sequences.
\end{lemma}

\begin{proof}
\emph{Existence and optimality condition.} Because $D_\psi(p,\pi)=\sum_a\bigl(p(a)/\pi(a)-1-\log(p(a)/\pi(a))\bigr)$ is strictly convex on the relative interior of $\Delta(\Aset)$ and diverges to $+\infty$ at the boundary, a unique minimizer $\pi'$ exists and has full support. Writing the Lagrange multiplier for $\sum_a p(a)=1$ as $\eta\lambda$, the KKT condition gives $\eta(\ell(a)-\lambda)-1/\pi'(a)+1/\pi(a)=0$, or equivalently
\[
  \pi'(a) = \frac{\pi(a)}{1+\zeta_a},\qquad \zeta_a = \eta\pi(a)(\ell(a)-\lambda),\qquad \Psi(\lambda) := \sum_{a\in\Aset}\frac{\pi(a)}{1+\zeta_a} = 1.
\]
On the open half-line $\lambda < \min_a(\ell(a)+1/(\eta\pi(a)))$ where every $1+\zeta_a>0$, $\Psi(\lambda)$ is continuous and strictly increasing, so its root $\lambda$ is unique.

\emph{Bracketing the root.} If $\bar\ell=0$, then $\ell=0$ and $\pi'=\pi$. Otherwise set $\Sigma_2=\sum_a\pi(a)^2$, $\pi_{\max}=\max_a\pi(a)$, $\lambda_0=\Sigma_2^{-1}\sum_a\pi(a)^2\ell(a)$, and $r_I=32\eta\pi_{\max}\bar\ell^2/\Sigma_2$. Observe that $\pi_{\max}^2\le\Sigma_2$, $\pi(a)\pi_{\max}\le\Sigma_2$, $\pi(a)|\ell(a)|\le\bar\ell$, and $|\lambda_0|\le\pi_{\max}\bar\ell/\Sigma_2$. For any $\lambda\in I=[\lambda_0-r_I,\lambda_0+r_I]$, we have $|\lambda|\le(\pi_{\max}\bar\ell/\Sigma_2)(1+32\eta\bar\ell)\le 2\pi_{\max}\bar\ell/\Sigma_2$, hence $\eta\pi(a)|\lambda|\le 2\eta\bar\ell$ and $|\zeta_a|\le 3\eta\bar\ell\le 3/32$, so $I$ lies inside the domain of $\Psi$. Also, $\sum_a\pi(a)\zeta_a=\eta\Sigma_2(\lambda_0-\lambda)$. Using $(1+t)^{-1}\ge 1-t$ at $\lambda=\lambda_0+r_I$ gives
\[
  \Psi(\lambda_0+r_I) \ge 1 - \sum_a\pi(a)\zeta_a = 1 + \eta\Sigma_2 r_I > 1.
\]

For the lower endpoint, using $(1+t)^{-1}\le 1-t+2t^2$ (valid for $t\ge -1/2$) at $\lambda=\lambda_0-r_I$ gives
\[
  \Psi(\lambda_0-r_I) \le 1 - \eta\Sigma_2 r_I + 2\sum_a\pi(a)\zeta_a^2.
\]
Note that $\sum_a\pi(a)^3\ell(a)^2\le\bar\ell\sum_a\pi(a)^2|\ell(a)|\le\pi_{\max}\bar\ell^2$, $\lambda_0^2\sum_a\pi(a)^3\le(\pi_{\max}\bar\ell/\Sigma_2)^2\pi_{\max}\Sigma_2\le\pi_{\max}\bar\ell^2$, and $\sum_a\pi(a)^3\le\pi_{\max}\Sigma_2$. Expanding $(\ell(a)-\lambda_0+r_I)^2\le 4\ell(a)^2+4\lambda_0^2+2r_I^2$ gives
\[
  2\sum_a\pi(a)\zeta_a^2 \le 16\eta^2\pi_{\max}\bar\ell^2 + 4\eta^2 r_I^2\pi_{\max}\Sigma_2 = \tfrac12\eta\Sigma_2 r_I + 128\eta^3 r_I\pi_{\max}^2\bar\ell^2 \le \tfrac58\eta\Sigma_2 r_I.
\]
The final step uses $(32\eta\bar\ell)^2\le 1$ and $\pi_{\max}^2\le\Sigma_2$, which imply $128\eta^2\pi_{\max}^2\bar\ell^2\le\Sigma_2/8$. Thus $\Psi(\lambda_0-r_I)<1<\Psi(\lambda_0+r_I)$, so the root lies in $I$. Since $|\log(1+t)|\le 2|t|$ for $|t|\le 1/2$, we obtain $|\log(\pi'(a)/\pi(a))|=|\log(1+\zeta_a)|\le 2|\zeta_a|\le 6\eta\bar\ell$.

\emph{Regret bound.} First-order optimality of $\pi_{i+1}$ and the three-point Bregman identity give, for any interior $u\in\Delta(\Aset)$,
\[
  \eta\ip{\pi_i-u}{\ell_i} \le D_\psi(u,\pi_i) - D_\psi(u,\pi_{i+1}) + \eta\ip{\pi_i-\pi_{i+1}}{\ell_i} - D_\psi(\pi_{i+1},\pi_i).
\]
By \cref{eq:stability}, $\pi_{i+1}(a)/\pi_i(a)\in[e^{-6/32},e^{6/32}]\subset[1/2,2]$. The function $t\mapsto t-1-\log t$ vanishes with its derivative at $t=1$ and has second derivative $1/t^2\ge 1/4$ on $[1/2,2]$, so Taylor's theorem yields $D_\psi(\pi_{i+1},\pi_i)=\sum_a\bigl(\tfrac{\pi_{i+1}(a)}{\pi_i(a)}-1-\log\tfrac{\pi_{i+1}(a)}{\pi_i(a)}\bigr)\ge\frac18\sum_a(\pi_i(a)-\pi_{i+1}(a))^2/\pi_i(a)^2$. Applying $\eta(\pi_i(a)-\pi_{i+1}(a))\ell_i(a)\le (\pi_i(a)-\pi_{i+1}(a))^2/(8\pi_i(a)^2)+2\eta^2\pi_i(a)^2\ell_i(a)^2$ gives
\[
  \eta\ip{\pi_i-u}{\ell_i} \le D_\psi(u,\pi_i) - D_\psi(u,\pi_{i+1}) + 2\eta^2\sum_a\pi_i(a)^2\ell_i(a)^2.
\]
Summing over $i\in[n]$ and dropping $-D_\psi(u,\pi_{n+1})\le 0$ telescopes the divergences. Since $D_\psi(u,\pi_1)=-\sum_a\log(Au(a))\le A\log K$, \cref{eq:omd_bound} follows.
\end{proof}

\paragraph{Pathwise bounds for self-normalized block estimates.}
Let $b$ be a complete active block of $s\in\Sset_h$, set $N_b(a)=\sum_{k\in b}\ind\{s_h^k=s,a_h^k=a\}$, and let $\widehat L_b=\ip{\pi_b}{\widehat\ell_b}$. Because $Y_h^k\in[0,H]$, we have $\Sigma_b(a)\le H N_b(a)$, and since $f_b\pi_b(a)/(\pi_b(a)+\gamma_b)\le 1$, every sample path satisfies
\begin{equation}\label{eq:selfnorm_pathwise}
  \pi_b(a)\widehat\ell_b(a) \le \Lambda_b H\,\frac{N_b(a)}{N_b},\qquad
  \widehat L_b \le \Lambda_b H,\qquad
  \sum_{a\in\Aset}\pi_b(a)^2\widehat\ell_b(a)^2 \le \Lambda_b H\,\widehat L_b
\end{equation}
(reading $0/0$ as $0$ when $N_b=0$). Since $\widehat\ell_b\ge 0$, $\bar\ell=\widehat L_b$, and by \ref{P:stab} we have $\eta\widehat L_b\le\eta H\Lmax\le 1/(4C_0 D)<1/32$. Thus \cref{lem:barrier} applies on every sample path to every update of \alg{} and shifts each $\log\pi(a\mid s)$ by at most $6\eta\widehat L_b\le 6\eta H\Lambda_b$.

%% file: appendix/D_drift.tex
\section{Policy Drift}\label{app:drift}

We bound how much the occupancy $q^k(s)$ and expected return targets $v^k(s,a)$ of an active state $s\in\Sset_h$ can drift during a block of length $\Lambda$. Nested dyadic blocks and log-barrier stability keep per-decision log-probability changes within $6\eta H\Lambda$ across the entire MDP (\cref{lem:dyadic}). A martingale log-likelihood-ratio bound then converts per-decision log-ratios $\vartheta$ into path-law domination of radius $O(H\vartheta^2+\vartheta\sqrt{H\log(1/\etail)})$ up to an additive tail $\etail$ (\cref{lem:pathlaw}). Representing $w^k$ through auxiliary survival coins extends this domination to discounted targets and gives our blockwise and two-period drift bounds (\cref{lem:drift}).

\subsection{Aligned Dyadic Drift Budget}

\begin{lemma}[Dyadic budget]\label{lem:dyadic}
Assume $\kappa\le 1/4$ and $a_0=C_0\eta HD$, so that $\eta H\Lmax\le 1/32$ by \ref{P:stab}. During any complete active block $b$ of length $\Lambda$, every local action probability $\pi^k(a'\mid s')$, at every state $s'\in\Sset$ and action $a'\in\Aset$, changes in logarithm by at most $6\eta H\Lambda$. Across any two consecutive periods, the change in logarithm is at most $12\eta H\Lmax$. Both bounds hold deterministically on every sample path.
\end{lemma}

\begin{proof}
Fix a complete active block $b=\{t_0,\dots,t_0+\Lambda-1\}$ of $s$ in period $j$, any state $s'\in\Sset$, and episodes $k<k'$ in $b$. The policy $\pi(\cdot\mid s')$ updates only at the ends of complete active blocks of $s'$ and takes effect at the start of the next episode. Thus $\pi^k(\cdot\mid s')$ and $\pi^{k'}(\cdot\mid s')$ differ only through updates at the ends of blocks $b'$ of $s'$ whose final episode lies in $\{k,\dots,k'-1\}\subseteq\{t_0,\dots,t_0+\Lambda-2\}$. Because blocks never cross period boundaries, every such $b'$ belongs to period $j$. Any two aligned dyadic intervals within period $j$ are nested or disjoint. Since $b'$ intersects $b$ and ends strictly before $b$ ends, we have $b'\subsetneq b$. As the blocks of $s'$ are pairwise disjoint, their lengths sum to at most $\Lambda$. By \cref{lem:barrier} and \cref{eq:selfnorm_pathwise}, the update after block $b'$ shifts $\log\pi(a'\mid s')$ by at most $6\eta H\Lambda_{b'}$. Summing over $b'\subsetneq b$ gives $|\log\pi^{k'}(a'\mid s')-\log\pi^k(a'\mid s')|\le 6\eta H\Lambda$.

Similarly, for episodes $k<k'$ spanning periods $j-1$ and $j$, any intervening update at $s'$ completes a block contained in period $j-1$ or period $j$. These blocks are disjoint of total length at most $2\Lmax$, which gives the bound $12\eta H\Lmax$.
\end{proof}

\subsection{Approximate Domination of Sequential Path Laws}

Consider a sequential experiment generating random variables $O_1,\dots,O_n$ under two probability laws $P$ and $Q$. Each index $i$ has a fixed deterministic type: either a \emph{common transition} (taking values in an arbitrary measurable space) where the regular conditional distributions satisfy $P_i(\cdot\mid O_{<i})=Q_i(\cdot\mid O_{<i})$, or a finite-valued \emph{decision} where $P_i(\cdot\mid O_{<i})$ and $Q_i(\cdot\mid O_{<i})$ share the same finite support and satisfy $|\log(P_i(o\mid O_{<i})/Q_i(o\mid O_{<i}))|\le \vartheta$ on that support. The following lemma is, in substance, the advanced composition theorem of differential privacy \citep{dwork2010boosting}: the two laws play the role of the outputs of $H$ adaptively composed mechanisms on neighboring inputs, $\vartheta$ that of the privacy parameter of each mechanism, and $e^{-y}$ that of the additive slack. We include the short proof because its constants enter \cref{lem:drift}.
\begin{lemma}[Path-law domination]\label{lem:pathlaw}
Let $P,Q$ be sequential laws as above with at most $H$ decision indices (padding with trivial point-mass decisions gives exactly $H$) and with per-decision log-likelihood ratios in $[-\vartheta,\vartheta]$, where $0\le \vartheta\le 1$. For any $y>0$, set $\rho = 4H\vartheta^2 + \vartheta\sqrt{2Hy}$. Then for every measurable $F$ satisfying $0\le F\le F_{\max}$,
\[
  \E_P[F] \le e^\rho \E_Q[F] + F_{\max}e^{-y}
  \qquad\text{and}\qquad
  \E_Q[F] \le e^\rho \E_P[F] + F_{\max}e^{-y}.
\]
Both inequalities also hold with $\rho$ replaced by $H\vartheta$ and zero additive error.
\end{lemma}

\begin{proof}
By common support, $P\ll Q$ with log-likelihood ratio $\mathcal L := \log\frac{dP}{dQ}(O) = \sum_{i\text{ decision}}\lambda_i$, where $\lambda_i = \log(P_i(O_i\mid O_{<i})/Q_i(O_i\mid O_{<i}))\in[-\vartheta,\vartheta]$. Under $P$, $\E_P[\lambda_i\mid O_{<i}] = \KL(P_i\,\|\,Q_i) \le \chi^2(P_i\,\|\,Q_i) = \sum_o Q_i(o)(P_i(o)/Q_i(o)-1)^2 \le (e^\vartheta-1)^2 \le 4\vartheta^2$ since $\vartheta\le 1$. Let $i_1<\dots<i_H$ be the deterministic decision indices, and set $i_{H+1}:=n+1$. The centered increments $\lambda_{i_m}-\E_P[\lambda_{i_m}\mid O_{<i_m}]$ ($m=1,\dots,H$) form a martingale difference sequence with respect to $(\sigma(O_{<i_{m+1}}))_{m=0}^H$, each taking values in an $O_{<i_m}$-measurable interval of length $2\vartheta$. By the Hoeffding--Azuma inequality, $P(\sum_{m=1}^H(\lambda_{i_m}-\E_P[\lambda_{i_m}\mid O_{<i_m}]) > \vartheta\sqrt{2Hy}) \le \exp(-4Hy\vartheta^2/(4H\vartheta^2)) = e^{-y}$. Adding $\sum_m\E_P[\lambda_{i_m}\mid O_{<i_m}]\le 4H\vartheta^2$ gives $P(\mathcal L > \rho)\le e^{-y}$. Hence
\[
  \E_P[F] = \E_Q[F e^{\mathcal L}\ind\{\mathcal L\le\rho\}] + \E_P[F\ind\{\mathcal L>\rho\}] \le e^\rho \E_Q[F] + F_{\max}e^{-y}.
\]
Swapping $P$ and $Q$ yields the reverse inequality, and $|\mathcal L|\le H\vartheta$ pathwise gives $\E_P[F]\le e^{H\vartheta}\E_Q[F]$ without additive error.
\end{proof}

\subsection{Augmented Survival-Coin Representation of Discounted Targets}

Fix an episode $k$, a state $s\in\Sset_h$, and an action $a\in\Aset$. Consider the \emph{augmented rollout} starting from $(s_h,a_h)=(s,a)$: pay immediate cost $c^k(s,a)$, and for each layer $h'=h+1,\dots,H$, first draw $s_{h'}\sim p(\cdot\mid s_{h'-1},a_{h'-1})$ (a common transition), then draw an action-survival pair $(a_{h'},\varsigma_{h'})\in\Aset\times\{0,1\}$ (a decision) with probabilities
\[
  \Prob(a_{h'}=a',\varsigma_{h'}=1) = \pi^k(a'\mid s_{h'})\,w^k(s_{h'},a'),\quad
  \Prob(a_{h'}=a',\varsigma_{h'}=0) = \pi^k(a'\mid s_{h'})\,(1-w^k(s_{h'},a')).
\]
If $\varsigma_{h'}=1$, pay $c^k(s_{h'},a_{h'})$ and continue; if $\varsigma_{h'}=0$, terminate the rollout so all remaining layers are trivial. Letting $F\in[0,H]$ be the total cost paid, backward induction on \cref{eq:vz} shows $\E[F] = v^k(s,a)$.

\paragraph{Log-sensitivity of augmented decisions.}
At an active state $s'$ with period parameters $f\in[3/4,1)$ and $\gamma>0$, the survival and termination probabilities of action $a'$ as functions of $\pi=\pi(a'\mid s')$ are
\[
  \varphi_{\mathrm{surv}}(\pi) = f^2\frac{\pi^2}{\pi+\gamma},\qquad
  \varphi_{\mathrm{kill}}(\pi) = \pi\,\frac{(1-f^2)\pi+\gamma}{\pi+\gamma}.
\]
Differentiating with respect to $\log\pi$ gives
\[
  \frac{\partial\log\varphi_{\mathrm{surv}}}{\partial\log\pi} = 2 - \frac{\pi}{\pi+\gamma}\in[1,2],\qquad
  \frac{\partial\log\varphi_{\mathrm{kill}}}{\partial\log\pi} = 1 + \frac{(1-f^2)\pi}{(1-f^2)\pi+\gamma} - \frac{\pi}{\pi+\gamma}\in[0,1].
\]
By the mean value theorem, if two policies share the same period parameters $(f,\gamma)$ and satisfy $|\log\pi'(a'\mid s')-\log\pi(a'\mid s')|\le t$, their augmented decision log-probabilities differ by at most $2t$. At inactive states, $\varphi_{\mathrm{surv}}\equiv 0$ and $\varphi_{\mathrm{kill}}=\pi$, so the log-ratio is at most $t$.

\subsection{Block and Two-Period Drift Bounds}

\begin{lemma}[Block drift]\label{lem:drift}
Assume $\kappa\le 1/4$ and recall $\etail=e^{-20J}$. The following hold deterministically on every sample path:
\begin{enumerate}[label=(\alph*),leftmargin=*]
\item\label{drift:add} \emph{Additive block drift.} Let $b$ be a complete active block of $s\in\Sset_h$ of length $\Lambda$ starting at episode $t_0$, let $v^{\circ,k}(s,a)$ denote the targets in \cref{eq:vz} evaluated with all policies frozen at $\pi^{t_0}$ (using episode $k$'s costs and the current period's statuses and margins), and set $\rho_b^0 = 76\eta H D\Lambda$. For all $k\in b$ and $a\in\Aset$,
\begin{gather*}
  e^{-\rho_b^0}q^{t_0}(s)-\etail \le q^k(s) \le e^{\rho_b^0}q^{t_0}(s)+\etail,\\
  e^{-\rho_b^0}v^{\circ,k}(s,a)-\etail H \le v^k(s,a) \le e^{\rho_b^0}v^{\circ,k}(s,a)+\etail H,
\end{gather*}
and the target bounds extend to any nonnegative action combination $\sum_a \alpha_a v^k(s,a)$ with $\sum_a\alpha_a\le 1$.
\item\label{drift:rel} \emph{Relative block occupancy.} If, in addition, $\eta\ge 1/(HK)$ and $q^{t_0}(s)\Lambda\ge M/3$, then for all $k\in b$,
\[
  e^{-\varrho_b}q^{t_0}(s) \le q^k(s) \le e^{\varrho_b}q^{t_0}(s),\qquad \varrho_b = 80\eta H D\Lambda \le 0.02.
\]
\item\label{drift:two} \emph{Two-period occupancy drift.} For any state $s\in\Sset$, any period $j\ge 2$ where $t$ is the first episode of period $j-1$, and any episode $k$ in period $j-1$ or period $j$,
\begin{equation}\label{eq:twoperiod}
  e^{-\rho_P}q^t(s)-\etail \le q^k(s) \le e^{\rho_P}q^t(s)+\etail,\qquad \rho_P = 76\eta H D\Lmax \le 0.02.
\end{equation}
\end{enumerate}
The reference quantities $q^{t_0}(s)$ and $v^{\circ,k}(s,a)$ are $\F_{t_0-1}$-measurable, and $q^t(s)$ is $\F_{t-1}$-measurable.
\end{lemma}

\begin{proof}
Fix $k\in b$ and write $\tau=\eta H\Lambda$. By \cref{lem:dyadic}, $|\log\pi^k(a'\mid s')-\log\pi^{t_0}(a'\mid s')|\le 6\tau$ everywhere, and period statuses and margins are identical at $t_0$ and $k$. Set $\vartheta=12\tau$, $y=20J$ (so $e^{-y}=\etail$), and $\bar\tau=\eta H D\Lambda$. By \ref{P:stab}, $\vartheta\le 12/(4C_0 D)<1$ and $\bar\tau\le 1/(4C_0)$. In \cref{lem:pathlaw}, if $D=\sqrt{HJ}\le H$, the radius is
\[
  \rho = 4H\vartheta^2 + \vartheta\sqrt{40HJ} = \frac{576\,\bar\tau^2}{J} + 12\sqrt{40}\,\bar\tau \le \Bigl(\frac{576}{4C_0 J} + 75.90\Bigr)\bar\tau \le 76\bar\tau = \rho_b^0,
\]
since $J\ge\log 1024 > 6.93$. Otherwise $D=H<\sqrt{HJ}$, and the zero-tail pathwise radius in \cref{lem:pathlaw} is $H\vartheta = 12\bar\tau \le \rho_b^0$.

Applying \cref{lem:pathlaw} to the ordinary trajectory indicator $F=\ind\{s_h=s\}\in[0,1]$ (whose per-step log-ratios are at most $6\tau\le \vartheta$) gives $e^{-\rho_b^0}q^{t_0}(s)-\etail\le q^k(s)\le e^{\rho_b^0}q^{t_0}(s)+\etail$. Applying \cref{lem:pathlaw} to the continuation cost in $[0,H]$ of the augmented survival-coin rollout (whose per-step log-ratios are at most $2\cdot 6\tau = \vartheta$) and adding the common initial cost $c^k(s,a)$ gives $e^{-\rho_b^0}v^{\circ,k}(s,a)-\etail H\le v^k(s,a)\le e^{\rho_b^0}v^{\circ,k}(s,a)+\etail H$, and linearity extends this to weights $\alpha_a\ge 0$ with $\sum_a\alpha_a\le 1$. Because $c^k$ is deterministic (\cref{sec:setting}) and $\pi^{t_0}$ plus period-$j$ parameters are $\F_{t_0-1}$-measurable, $v^{\circ,k}(s,a)\in[0,H]$ is $\F_{t_0-1}$-measurable.

For \ref{drift:rel}, assume $\eta\ge 1/(HK)$ and $\bar q:=q^{t_0}(s)\ge M/(3\Lambda)\ge M/(3K)$, and set $\varpi=\etail/\bar q\le 3K\etail/M$. Part~\ref{drift:add} gives $q^k(s)\le\bar q(e^{\rho_b^0}+\varpi)\le\bar q e^{\rho_b^0+\varpi}$. Since $e^{\rho_b^0}\le 2$ by \ref{P:radius} and $\varpi e^{\rho_b^0}\le 6K\etail/M \le 1/2$ by \ref{P:tiny}, applying $1-t\ge e^{-2t}$ on $[0,1/2]$ to part~\ref{drift:add} gives $q^k(s)\ge\bar q e^{-\rho_b^0}(1-\varpi e^{\rho_b^0})\ge\bar q e^{-\rho_b^0-4\varpi}$. Because $\eta\ge 1/(HK)$, $D\ge 1$, and $\Lambda\ge M\ge 1$, we have $\eta H D\Lambda\ge 1/K$, whereas $4\varpi\le 12K\etail/M < 10^{-55}/K \le 4\eta H D\Lambda$ by \ref{P:tiny}. Hence $\rho_b^0+4\varpi\le 80\eta H D\Lambda = \varrho_b$, and $\varrho_b\le 0.02$ by \ref{P:radius}.

Finally, for \ref{drift:two}, \cref{lem:dyadic} bounds ordinary trajectory log-ratios across periods $j-1$ and $j$ by $\vartheta_P = 12\eta H\Lmax$. Because $\vartheta_P$ has the same coefficient $12$ as $\vartheta=12\tau$ above, repeating the calculation with $\Lmax$ in place of $\Lambda$ gives \cref{eq:twoperiod} with $\rho_P = 76\eta H D\Lmax \le 76/(4C_0) < 0.02$.
\end{proof}

%% file: appendix/E_selfnorm.tex
\section{Relative Self-Normalization}\label{app:selfnorm}

Throughout this section we fix a state $s\in\Sset_h$ ($h=h(s)$) and write $X_k=\ind\{s_h^k=s\}$.

\subsection{Per-State Uniform Coupling}\label{app:coupling}

Let $U_1,\dots,U_K$ and $\zeta_1,\dots,\zeta_K$ be independent $\Unif(0,1]$ random variables. Given the first $k-1$ episodes (their trajectories and, for an algorithm with internal randomness, the random bits it drew in them), which determine $\pi^k$ and $q^k(s)$, we recursively set $X_k=\ind\{U_k\le q^k(s)\}$. Using $\zeta_k$, we then generate episode $k$'s trajectory $\tau^k$ (and any internal episode-$k$ randomness) from the conditional law under $\pi^k$ given $\{s_h=s\}$ when $X_k=1$ and given $\{s_h\ne s\}$ when $X_k=0$. Because $U_k\in(0,1]$, the selected conditioning event always has positive probability, and the coupled process has the exact joint distribution of the original interaction with $X_k=\ind\{s_h^k=s\}$. Writing $\widetilde\F_k=\sigma(U_{\le k},\zeta_{\le k})\supseteq\F_k$, the following properties hold:
\begin{enumerate}[label=(C\arabic*),leftmargin=*]
\item\label{C:future} $(U_{k'})_{k'>k}$ is independent of $\widetilde\F_k$.
\item\label{C:episode} $(U_k,\zeta_k)$ is independent of $\mathcal M_k := \widetilde\F_{k-1}\vee\sigma(U_{k'}:k'>k)$, so the conditional law of episode $k$'s trajectory given $\mathcal M_k$ coincides with its conditional law given $\F_{k-1}$.
\item\label{C:obs} All $\F_k$-measurable events and conditional expectations given $\F_k$ agree with those in the original space.
\end{enumerate}

\subsection{Statement and Proof of the Relative Bias Bound}

\begin{lemma}[Relative self-normalization bias]\label{lem:relative_sn}
Fix a deterministic block $b=\{t_0,\dots,t_0+\Lambda-1\}\subseteq\{1,\dots,K\}$. For each $k\in b$ let $R_k\ge 0$ be an integrable $\F_k$-measurable random variable and let $g_k\in[0,H]$ be $\F_{k-1}$-measurable. Let $\varrho\in[0,0.1]$ and $\etail\in[0,1]$ be constants, and let $\bar q$ and $g_k^\circ\in[0,H]$ ($k\in b$) be $\F_{t_0-1}$-measurable. Let $\Omega_0\in\F_{t_0-1}$ be an event such that, for almost every $\omega\in\Omega_0$, the following hold simultaneously for all $k\in b$ (that is, $\Omega_0$ guarantees them along every continuation of the block):
\begin{enumerate}[label=(H\arabic*),leftmargin=*]
\item\label{H:size} $\bar q\in(0,1]$ and $\bar q\Lambda\ge M/3$;
\item\label{H:occ} $e^{-\varrho}\bar q\le q^k(s)\le e^\varrho\bar q$;
\item\label{H:target} $\E[X_k R_k\mid\F_{k-1}] = q^k(s)g_k$ and $e^{-\varrho}g_k^\circ - \etail H \le g_k \le e^\varrho g_k^\circ + \etail H$.
\end{enumerate}
Let $N=\sum_{k\in b}X_k$, $\widehat G=\frac{\Lambda}{N}\sum_{k\in b}X_k R_k$ ($\widehat G=0$ if $N=0$), and $G=\sum_{k\in b}g_k$. Then on $\Omega_0$,
\[
  \bigl|\E[\widehat G - G \mid \F_{t_0-1}]\bigr| \le (e^{4\varrho}-1)\,\E[G\mid\F_{t_0-1}] + \Lambda H\bigl(3\etail + e^{-M/3}\bigr).
\]
\end{lemma}

\begin{proof}
Work in the coupled space of \cref{app:coupling}, writing $\E_0[\cdot]=\E[\cdot\mid\widetilde\F_{t_0-1}]$, $G^\circ=\sum_{k\in b}g_k^\circ\le\Lambda H$, and $\theta_{\mathrm{sn}}=e^{-M/3}$.

\emph{Step 1 (Uniform sandwich).} Set $p_\circ=e^{-\varrho}\bar q$, $p_+=\min(1,e^\varrho\bar q)$, $X_k^\circ=\ind\{U_k\le p_\circ\}$, and $X_k^+=\ind\{U_k\le p_+\}$. By \ref{H:occ}, $X_k^\circ\le X_k\le X_k^+$ for all $k\in b$ on $\Omega_0$. Defining $N_{-k}^\circ=\sum_{k'\in b\setminus\{k\}}X_{k'}^\circ$ and $N_{-k}^+=\sum_{k'\in b\setminus\{k\}}X_{k'}^+$, we have on $\Omega_0\cap\{X_k=1\}$
\begin{equation}\label{eq:sn_sandwich}
  1 + N_{-k}^\circ \le N \le 1 + N_{-k}^+.
\end{equation}
By \ref{C:future}, conditionally on $\widetilde\F_{t_0-1}$, $(U_k)_{k\in b}$ are i.i.d., so $N_{-k}^\circ\sim\mathrm{Bin}(\Lambda-1,p_\circ)$ and $N_{-k}^+\sim\mathrm{Bin}(\Lambda-1,p_+)$.

\emph{Step 2 (Conditioning outside episode $k$).} Both $N_{-k}^\circ$ and $N_{-k}^+$ depend only on $\bar q$ and $(U_{k'})_{k'\ne k}$, so both are $\mathcal M_k$-measurable. Meanwhile, $X_k R_k$ depends only on $\widetilde\F_{k-1}$ and $(U_k,\zeta_k)$, so \ref{C:episode} and \ref{H:target} give $\E[X_k R_k\mid\mathcal M_k]=\E[X_k R_k\mid\F_{k-1}]=q^k(s)g_k$. Because $R_k\ge 0$, \cref{eq:sn_sandwich} yields on $\Omega_0$ (with $0/0:=0$ when $N=0$)
\[
  \frac{X_k R_k}{1+N_{-k}^+} \le \frac{X_k R_k}{N} \le \frac{X_k R_k}{1+N_{-k}^\circ}
  \quad\Longrightarrow\quad
  \frac{q^k(s)g_k}{1+N_{-k}^+} \le \E\Bigl[\frac{X_k R_k}{N}\Bigm|\mathcal M_k\Bigr] \le \frac{q^k(s)g_k}{1+N_{-k}^\circ}.
\]

\emph{Step 3 (Evaluating binomial reciprocals).} On $\Omega_0$, \ref{H:occ} and \ref{H:target} give $e^{-\varrho}\bar q(e^{-\varrho}g_k^\circ-\etail H)_+ \le\allowbreak q^k(s)g_k \le\allowbreak e^\varrho\bar q(e^\varrho g_k^\circ+\etail H)$, where both outer envelopes are $\widetilde\F_{t_0-1}$-measurable. Since $\E[(1+\mathrm{Bin}(n,p))^{-1}] = \frac{1-(1-p)^{n+1}}{(n+1)p}$ for all $p\in(0,1]$ (including $p_+=1$), we obtain
\[
  \E_0\Bigl[\frac{1}{1+N_{-k}^\circ}\Bigr] \le \frac{1}{\Lambda p_\circ} = \frac{e^\varrho}{\Lambda\bar q},\qquad
  \E_0\Bigl[\frac{1}{1+N_{-k}^+}\Bigr] = \frac{1-(1-p_+)^\Lambda}{\Lambda p_+} \ge \frac{1-\theta_{\mathrm{sn}}}{\Lambda e^\varrho\bar q},
\]
since $\bar q\le p_+\le e^\varrho\bar q$ (from $\bar q\le 1$ and $\varrho\ge 0$) gives $(1-p_+)^\Lambda \le e^{-\Lambda\bar q}\le\theta_{\mathrm{sn}}$ by \ref{H:size}. Taking $\E_0$ in Step~2, multiplying by $\Lambda$, and canceling $\bar q$ gives $\Lambda\,\E_0[q^k(s)g_k/(1+N_{-k}^+)] \ge (1-\theta_{\mathrm{sn}})e^{-2\varrho}(e^{-\varrho}g_k^\circ-\etail H)_+ \ge e^{-3\varrho}g_k^\circ - H(\etail+\theta_{\mathrm{sn}})$ since $g_k^\circ\le H$, and the analogous computation with $N_{-k}^\circ$ gives the upper bound. Summing over $k\in b$ gives, on $\Omega_0$,
\begin{equation}\label{eq:sn_tilde}
  e^{-3\varrho}G^\circ - \Lambda H(\etail+\theta_{\mathrm{sn}}) \le \E_0[\widehat G] \le e^{3\varrho}G^\circ + e^{2\varrho}\Lambda\etail H.
\end{equation}

\emph{Step 4 (Eliminating the frozen reference $G^\circ$).} Summing \ref{H:target} over $k\in b$ gives $e^{-\varrho}(\E_0[G]-\Lambda\etail H)\le G^\circ \le e^\varrho(\E_0[G]+\Lambda\etail H)$. Substituting into \cref{eq:sn_tilde} and using $1-e^{-4\varrho}\le e^{4\varrho}-1$ and $e^{4\varrho}+e^{2\varrho}\le 3$ for $\varrho\le 0.1$ yields $|\E_0[\widehat G-G]|\le (e^{4\varrho}-1)\E_0[G] + \Lambda H(3\etail+\theta_{\mathrm{sn}})$ on $\Omega_0$. Conditioning down to $\F_{t_0-1}\subseteq\widetilde\F_{t_0-1}$ via Jensen's inequality and invoking \ref{C:obs} completes the proof.
\end{proof}

\begin{remark}[Other ranges]\label{rem:sn_range}
The bound of \cref{lem:relative_sn} is homogeneous in the scale of the observations. If $g_k,g_k^\circ\in[0,B]$ for some $B>0$ and \ref{H:target} holds with $\etail B$ in place of $\etail H$, then applying the lemma to $R_kH/B$, $g_kH/B$ and $g_k^\circ H/B$ gives the same conclusion with $\Lambda B(3\etail+e^{-M/3})$ in place of $\Lambda H(3\etail+e^{-M/3})$. \cref{app:aggregate} uses the lemma in this form.
\end{remark}

\subsection{Applications to the Block Estimator}\label{app:sn_applications}

Assume $\kappa\le 1/4$ and $\eta\ge 1/(HK)$. For any deterministic interval $b=\{t_0,\dots,t_0+\Lambda_b-1\}$, define $\pi_b := \pi^{t_0}(\cdot\mid s)$ (which has full support everywhere) and $\chi_b(a) := \pi_b(a)/(\pi_b(a)+\gamma_0\Lambda_b)$, and let $\Omega_b\in\F_{t_0-1}$ be the event that $b$ is a complete active block of $s$ satisfying $q^{t_0}(s)\Lambda_b\ge M/3$. On $\Omega_b$, state $s$ is not updated inside $b$, so $\pi^k(\cdot\mid s)=\pi_b$ for all $k\in b$, and the conditional-mean identity of \cref{app:identity} gives, for every $\F_{t_0-1}$-measurable $\varphi:\Aset\to[0,\infty)$ and every $k\in b$,
\[
  \E\bigl[X_k\,\varphi(a_h^k)\,Y_h^k\bigm|\F_{k-1}\bigr]=q^k(s)\sum_{a\in\Aset}\pi_b(a)\varphi(a)\,v^k(s,a).
\]
Hence $R_k:=\varphi(a_h^k)Y_h^k$ satisfies the identity in \ref{H:target} with $g_k:=\sum_a\alpha_a v^k(s,a)$ and $\alpha_a:=\pi_b(a)\varphi(a)$. If $\sum_a\alpha_a\le 1$, then \cref{lem:drift}\ref{drift:add}--\ref{drift:rel} verify \ref{H:size}, \ref{H:occ} and the two-sided bound in \ref{H:target} with $\bar q=q^{t_0}(s)$, $\varrho=\varrho_b\le 0.02$, $\etail=e^{-20J}$ and $g_k^\circ=\sum_a\alpha_a v^{\circ,k}(s,a)$. (Outside $\Omega_b$ the policy at $s$ may change inside the interval, which is why the bounds below are claimed on $\Omega_b$ only.) With $\beta_b=5\varrho_b\ge e^{4\varrho_b}-1$ and $e_b=\Lambda_b H(3\etail+e^{-M/3})$, \cref{lem:relative_sn} yields two bounds on $\Omega_b$:
\begin{enumerate}[label=(A\arabic*),leftmargin=*]
\item\label{A:vector} \emph{Actionwise estimate.} Taking $\varphi=\ind\{\cdot=a\}/\pi_b(a)$, that is, $R_k = \ind\{a_h^k=a\}Y_h^k/\pi_b(a)$, gives $\widehat G = \widehat G_b^{\SN}(a) := \frac{\Lambda_b}{N_b}\frac{\Sigma_b(a)}{\pi_b(a)}$ and $G = G_b(a) := \sum_{k\in b}v^k(s,a)$, so
\[
  \bigl|\E[\widehat G_b^{\SN}(a) - G_b(a)\mid\F_{t_0-1}]\bigr| \le \beta_b\E[G_b(a)\mid\F_{t_0-1}] + e_b.
\]
\item\label{A:scalar} \emph{On-policy estimate.} Taking $\varphi=\chi_b$, that is, $R_k = \chi_b(a_h^k)Y_h^k$, gives $\widehat G = \widehat L_b/f_b$ where $\widehat L_b=\ip{\pi_b}{\widehat\ell_b}$, and $G = m_b := \sum_{k\in b}\sum_a\pi_b(a)\chi_b(a)v^k(s,a)$, so
\[
  \bigl|\E[\widehat L_b\mid\F_{t_0-1}] - f_b\E[m_b\mid\F_{t_0-1}]\bigr| \le f_b\bigl(\beta_b\E[m_b\mid\F_{t_0-1}] + e_b\bigr).
\]
\end{enumerate}

%% file: appendix/F_classification.tex
\section{Active and Inactive Classification}\label{app:classification}

\begin{lemma}[Active and inactive states]\label{lem:classification}
Assume $\kappa\le 1/4$. With probability at least $1-2SKe^{-J}$, for every period $j\ge 2$, every episode $k$ in period $j$ and every state $s$: if $s$ is active then $\frac13 M\le q^k(s)\Lambda_j(s)\le 5M$, and if $s$ is inactive then $q^k(s)\le 3M/\Lmax$.
\end{lemma}

For the proof we use two standard multiplicative Chernoff bounds for $Z\sim\mathrm{Bin}(n,p)$ with mean $\mu=np$:
\begin{equation}\label{eq:binom_tails}
  \Prob(Z \ge 2\mu + 3J) \le e^{-J},\qquad
  \Prob(Z \le \mu/2) \le e^{-\mu/8}.
\end{equation}
Markov's inequality on $2^Z$ gives $\Prob(Z\ge 2\mu+3J)\le (1+p)^n 2^{-2\mu-3J}\le \exp(\mu(1-2\log 2)-3J\log 2)\le e^{-J}$. Similarly, Markov's inequality on $2^{-Z}$ gives $\Prob(Z\le\mu/2)\le 2^{\mu/2}(1-p/2)^n\le\exp(\mu(\tfrac12\log 2-\tfrac12))\le e^{-\mu/8}$.

\begin{proof}[Proof of \cref{lem:classification}]
Fix a state $s\in\Sset_h$ and a period $j\ge 2$, and abbreviate $L=\Lmax$. Because period $j$ exists, period $j-1$ is complete and $L<K$. Let $t$ be the first episode of period $j-1$ and set $\bar q=q^t(s)$, which is $\F_{t-1}$-measurable. By \cref{eq:twoperiod}, for every episode $k$ in period $j-1$ or period $j$,
\[
  p_- := (e^{-\rho_P}\bar q - \etail)_+ \le q^k(s) \le p_+ := \min\{1,\, e^{\rho_P}\bar q + \etail\}.
\]
Applying the coupling of \cref{app:coupling} to $s$ and setting $N_\pm = \sum_{k\in\text{period }j-1}\ind\{U_k\le p_\pm\}$, we have $N_-\le N_{j-1}(s)\le N_+$ pathwise, and conditionally on $\widetilde\F_{t-1}$ we have $N_\pm\sim\mathrm{Bin}(L,p_\pm)$ with means $\mu_\pm = L p_\pm$. By \cref{eq:binom_tails}, outside an event of conditional probability at most $2e^{-J}$,
\begin{equation}\label{eq:cl_good}
  N_+ \le 2\mu_+ + 3J \qquad\text{and}\qquad \bigl(\mu_-\ge 8J \implies N_-\ge\mu_-/2\bigr).
\end{equation}
We verify that \cref{eq:cl_good} implies both conclusions for $s$ across all episodes $k$ of period $j$, using $\rho_P\le 0.02$, $M\ge 200J$, and $\etail L\le\etail K < 10^{-55}\le 10^{-6}$ by \ref{P:tiny}.

\emph{Active case ($N_{j-1}(s)\ge M$).} From $M\le N_+\le 2L(e^{\rho_P}\bar q+\etail)+3J$, we get $\bar q L\ge e^{-\rho_P}(M-3J-2\etail L)/2\ge 0.48M$. Hence $\mu_-\ge e^{-\rho_P}\bar q L-\etail L\ge 0.47M\ge 8J$, so $N_{j-1}(s)\ge N_-\ge\mu_-/2\ge 0.49\bar q L$ (since $e^{-0.02}/2\ge 0.49009$ and $\etail L\le 10^{-5}\bar q L$). Likewise $N_{j-1}(s)\le N_+\le 2.041\bar q L + 2\etail L + 3J \le 2.1\bar q L$, since $3J\le 0.015M\le 0.032\bar q L$. For every episode $k$ in period $j$, \cref{eq:twoperiod} gives $|q^k(s)/\bar q - 1|\le 0.021$, and \cref{eq:blocklength} gives $ML/N_{j-1}(s)\le\Lambda_j(s)<2ML/N_{j-1}(s)$. Therefore
\[
  q^k(s)\Lambda_j(s) \ge 0.979\bar q\,\frac{ML}{2.1\bar q L} \ge \frac{M}{3},\qquad
  q^k(s)\Lambda_j(s) \le 1.021\bar q\,\frac{2ML}{0.49\bar q L} \le 5M.
\]

\emph{Inactive case ($N_{j-1}(s)<M$).} Whether $\mu_-\ge 8J$ (so $\mu_-/2\le N_-\le N_{j-1}(s)<M$) or $\mu_-<8J$, we have $L(e^{-\rho_P}\bar q-\etail) \le \mu_- < 2M$. Thus $\bar q L < e^{\rho_P}(2M+\etail L)$, and for every episode $k$ in period $j$, $q^k(s)L \le e^{\rho_P}\bar q L + \etail L \le e^{2\rho_P}(2M+\etail L)+\etail L \le 3M$.

\emph{Union bound.} By \ref{C:obs}, for each $s\in\Sset$ and period $j\ge 2$, the trajectory-measurable failure event has probability at most $2e^{-J}$. Taking a union bound over the $S$ states and at most $K$ periods yields failure probability at most $2SKe^{-J}$.
\end{proof}

Let $\Ecl$ denote the event on which the conclusions of \cref{lem:classification} hold. On $\Ecl$, at the initial episode $t_0$ of every complete active block $b$ of $s$, we have $q^{t_0}(s)\Lambda_b\ge M/3$, which certifies $\Omega_b$ in \cref{app:sn_applications}.

%% file: appendix/G_events.tex
\section{Concentration and the Good Event}\label{app:events}

\subsection{Stopping-Time Indexing of Active Blocks}

Fix a state $s\in\Sset_h$, and let $n_s<K$ denote the realized number of complete active blocks of $s$ in periods $j\ge 2$. Let $T_i$ be the last episode preceding the $i$th complete active block of $s$, setting $T_i=K$ for $i>n_s$, and define $\G_i=\F_{T_i}$. Because period statuses and block lengths are fixed at period boundaries from previous-period visit counts, $\{T_i=t\}\in\F_t$ for all $t\le K$, so each $T_i$ is a stopping time with respect to $(\F_t)$. For $i\le n_s$, $T_i+\Lambda_i\le T_{i+1}$ is also a stopping time, so all statistics of block $b_i=\{T_i+1,\dots,T_i+\Lambda_i\}$ are measurable with respect to $\F_{T_i+\Lambda_i}\subseteq\G_{i+1}$. We index the $\G_i$-measurable parameters of \cref{app:notation} on block $b_i$ by subscript $i$ ($\Lambda_i,\pi_i,\epsilon_i,f_i,\gamma_i,\chi_i,\tau_i,\beta_i,e_i$). For $i>n_s$ (where $\{T_i=K\}\in\G_i$), we set $\Lambda_i=0$ (so $\epsilon_i=\gamma_i=\tau_i=\beta_i=e_i=0$ and $f_i=1$) and set all block sums and estimators to zero, so their $\G_i$-conditional expectations also vanish.

\paragraph{Localization to deterministic intervals.}
For any deterministic $t<K$ and dyadic length $\Lambda\in[M,\Lmax]$ with $t+\Lambda\le K$, on $\{T_i=t,\Lambda_i=\Lambda\}\in\F_t$ the block $b_i$ equals $b=\{t+1,\dots,t+\Lambda\}$ and $\E[\cdot\mid\G_i]=\E[\cdot\mid\F_t]$ almost surely. Also $\{T_i=t,\Lambda_i=\Lambda\}\cap\Ecl\subseteq\Omega_b$ (\cref{app:classification}). Whenever $\eta\ge 1/(HK)$, the almost-sure $\F_t$-measurable bounds \ref{A:vector} and \ref{A:scalar} hold on $\Omega_b$. On the complementary event $\{T_i=K\}\in\G_i$, all block sums and estimators (along with $e_i$) vanish, so \ref{A:vector} and \ref{A:scalar} hold trivially as $0\le 0$. Patching across $\{T_i=K\}$ and the finitely many pairs $(t,\Lambda)$ with $t<K$ shows that on $\Ecl$, both bounds hold at $\G_i$ for every $i\ge 1$ almost surely. Note that $\Ecl$, which is not $\F_t$-measurable, is used only to certify that the sample point lies in $\Omega_b$: \ref{A:vector} and \ref{A:scalar} are inequalities between $\F_t$-measurable random variables that hold almost surely on all of $\Omega_b\in\F_t$, so no conditional expectation is restricted to, or conditioned on, $\Ecl$. The patching involves finitely many almost-sure statements; their exceptional null set has probability zero and is discarded throughout.

\subsection{Exponential Supermartingale Bounds}

\begin{lemma}[One-sided Freedman inequality]\label{lem:freedman}
Let $(\G_i)_{i\ge 1}$ be a filtration and $(\xi_i)_{i\ge 1}$ be square-integrable $\G_{i+1}$-measurable increments satisfying $\E[\xi_i\mid\G_i]=0$ and $\lambda\xi_i\le 1$ almost surely for a constant $\lambda>0$. Then
\[
  \Prob\Bigl(\exists n\ge 1:\ \sum_{i=1}^n \xi_i > \lambda\sum_{i=1}^n \E[\xi_i^2\mid\G_i] + \frac{J}{\lambda}\Bigr) \le e^{-J}.
\]
\end{lemma}
\begin{proof}
Since $e^t\le 1+t+t^2$ for $t\le 1$, $\E[e^{\lambda\xi_i}\mid\G_i]\le 1+\lambda^2\E[\xi_i^2\mid\G_i]\le\exp(\lambda^2\E[\xi_i^2\mid\G_i])$. Thus $\mathcal M_n=\exp(\lambda\sum_{i=1}^n\xi_i - \lambda^2\sum_{i=1}^n\E[\xi_i^2\mid\G_i])$ is a nonnegative supermartingale with $\mathcal M_0=1$, and Ville's maximal inequality gives $\Prob(\exists n:\mathcal M_n\ge e^J)\le e^{-J}$.
\end{proof}

\begin{lemma}[Block-scaled implicit exploration]\label{lem:ix}
Fix a state $s\in\Sset_h$ and a deterministic $u\in\Delta(\Aset)$, and let $\widehat G_i^{\SN}(a)=\frac{\Lambda_i}{N_i}\frac{\Sigma_i(a)}{\pi_i(a)}$ ($0$ if $N_i=0$ or $i>n_s$) for $i\ge 1$. Then
\[
  \Prob\Bigl(\exists n\ge 1:\ \sum_{i=1}^n\bigl(\ip{u}{\widehat\ell_i} - f_i\,\E[\ip{u}{\widehat G_i^{\SN}}\mid\G_i]\bigr) > \frac{HJ}{2\gamma_0}\Bigr) \le e^{-J}.
\]
\end{lemma}
\begin{proof}
On block $i$ with $\gamma=\gamma_i=\gamma_0\Lambda_i$, $f=f_i$, $\pi=\pi_i$, and visit episode set $\mathcal V_i$ ($|\mathcal V_i|=N_i$), for any $y\in[0,1]$ and $p>0$, setting $t=2\gamma y/p\ge 0$ gives
\[
  \frac{2\gamma y}{p+\gamma} \le \frac{2\gamma y}{p+\gamma y} = \frac{2t}{2+t} \le \log(1+t),
\]
since $\frac{d}{dt}(\log(1+t)-\frac{2t}{2+t}) = \frac{t^2}{(1+t)(2+t)^2}\ge 0$. Multiplying by $f u(a)\in[0,1]$ and using $(1+t)^c\le 1+ct$ for $c\in[0,1]$ gives, with $y_k = Y_h^k/H\in[0,1]$,
\[
  \exp\Bigl(\frac{2\gamma f u(a_h^k)y_k}{\pi(a_h^k)+\gamma}\Bigr) \le 1 + \frac{2\gamma f u(a_h^k)y_k}{\pi(a_h^k)}.
\]
Because $\frac{2\gamma_0}{H}\ip{u}{\widehat\ell_i} = \frac{1}{N_i}\sum_{k\in\mathcal V_i}\frac{2\gamma f u(a_h^k)y_k}{\pi(a_h^k)+\gamma}$ is a uniform average over $\mathcal V_i$, Jensen's inequality for $\exp$ gives
\[
  \exp\Bigl(\frac{2\gamma_0}{H}\ip{u}{\widehat\ell_i}\Bigr) \le \frac{1}{N_i}\sum_{k\in\mathcal V_i}\Bigl(1 + \frac{2\gamma f u(a_h^k)y_k}{\pi(a_h^k)}\Bigr) = 1 + \frac{2\gamma_0}{H}f\ip{u}{\widehat G_i^{\SN}}
\]
(and $1\le 1$ when $N_i=0$ or $i>n_s$). All quantities here are bounded: $\frac{2\gamma_0}{H}\ip{u}{\widehat\ell_i}\le 2$, and $\widehat G_i^{\SN}(a)\le\Lambda_iH/\pi_i(a)$ with $\pi_i(a)\ge A^{-1}e^{-6\eta HK}$, since every update at $s$ moves $\log\pi(a\mid s)$ by at most $6\eta H\Lambda_{b'}$ and the blocks of $s$ are disjoint. Taking $\E[\cdot\mid\G_i]$ and using $1+t\le e^t$ makes $\exp(\frac{2\gamma_0}{H}\sum_{i=1}^n(\ip{u}{\widehat\ell_i}-f_i\E[\ip{u}{\widehat G_i^{\SN}}\mid\G_i]))$ a nonnegative supermartingale, and Ville's inequality concludes.
\end{proof}

\subsection{Definition and Probability of the Good Event}

For $i\le n_s$ on block $b_i$ of $s\in\Sset_h$ (and setting all block quantities below to $0$ for $i>n_s$), define
\begin{equation}\label{eq:blockquantities}
\begin{gathered}
  u = (1-\tfrac1K)\pi^\star(\cdot\mid s) + \tfrac1K\Unif(\Aset),\qquad
  G_i(a) = \sum_{k\in b_i}v^k(s,a),\qquad
  m_i = \sum_{a\in\Aset}\pi_i(a)\chi_i(a)G_i(a),\\
  m_i^u = \ip{u}{G_i},\qquad
  \widehat L_i = \ip{\pi_i}{\widehat\ell_i},\qquad
  \tau_i = \eta H\Lambda_i,\qquad
  W_i = f_i^2 m_i - (1-2\tau_i)\widehat L_i,\\
  \xi_i^\lrn = W_i - \E[W_i\mid\G_i],\qquad
  \xi_i^\cmp = \E[m_i^u\mid\G_i] - m_i^u.
\end{gathered}
\end{equation}
By \cref{eq:selfnorm_pathwise} and \cref{lem:identity}, $f_i^2 m_i$, $(1-2\tau_i)\widehat L_i$, and $m_i^u$ all lie in $[0,\Lambda_i H]$, so $|\xi_i^\lrn|\le 2\Lambda_i H$, $|\xi_i^\cmp|\le\Lambda_i H$, and $2\eta\Lambda_i H\le 1$ by \ref{P:stab}.

\begin{definition}[Good event]\label{def:good}
For each state $s\in\Sset$, let $\good_s^\lrn$ and $\good_s^\cmp$ be the complements of the failure events in \cref{lem:freedman} with $\lambda=\eta$ applied to $(\xi_i^\lrn)$ and $(\xi_i^\cmp)$, respectively, and let $\good_s^{\mathrm{IX}}$ be the complement of the failure event in \cref{lem:ix} for $u$ in \cref{eq:blockquantities}. Set $\good = \Ecl\cap\bigcap_{s\in\Sset}\bigl(\good_s^\lrn\cap\good_s^\cmp\cap\good_s^{\mathrm{IX}}\bigr)$.
\end{definition}

\begin{lemma}[Good-event probability]\label{lem:good}
Assuming $\kappa\le 1/4$, $\Prob(\good)\ge 1-\delta$.
\end{lemma}
\begin{proof}
By \cref{lem:classification} and \cref{lem:freedman,lem:ix}, a union bound over all $S$ states gives $\Prob(\good^c)\le 2SKe^{-J} + 3Se^{-J} \le 5SKe^{-J} = \frac{5\delta}{64AK^2} \le \delta$.
\end{proof}

%% file: appendix/H_local.tex
\section{Local Regret of Each State}\label{app:local}

Assume $\kappa\le 1/4$ and work on the good event $\good$ (\cref{def:good}); we prove \cref{lem:local}. Fix a state $s\in\Sset_h$. If $\eta < 1/(HK)$, then since $\Phi^k(s)\le H$ for every episode $k$ (\cref{lem:identity}) and $2J > 2\log 1024 > 1$, we have deterministically $\sum_{k=1}^K \Phi^k(s) \le HK < (A\log K + 2J)/\eta$, so \cref{lem:local} holds trivially. Assume henceforth that $\eta \ge 1/(HK)$, so that \ref{A:vector} and \ref{A:scalar} apply at every block start $\G_i$ on $\Ecl$ (\cref{app:events}).

By \ref{P:margin}, \ref{P:radius}, and \ref{P:slack}, the block parameters of \cref{eq:blockquantities} satisfy
\begin{equation}\label{eq:slack}
  \beta_i = \tfrac25\epsilon_i,\qquad \tau_i = \eta H\Lambda_i \le \frac{\epsilon_i}{C_0 D} \le \frac{\epsilon_i}{1000},\qquad f_i \ge \tfrac34.
\end{equation}
\cref{tab:budget} summarizes how every block-level error term is absorbed by the learner and comparator margins or bounded once per state.

\begin{table}[ht]
\caption{Error budget for the local regret of state $s$ on block $i$ ($\E_i[\cdot]=\E[\cdot\mid\G_i]$). The learner margin $f_i\epsilon_i\E_i[m_i]$ and comparator margin $\epsilon_i\E_i[m_i^u]$ absorb all per-block multiplicative bias and variance terms.}
\label{tab:budget}
\begin{center}
\begin{footnotesize}
\setlength{\tabcolsep}{3.5pt}
\begin{tabular}{llll}
\toprule
Error term & Source & Bound & How it is paid\\
\midrule
self-normalization bias (learner) & \ref{A:scalar} & $f_i\beta_i\E_i[m_i]$ ($+e_i$) & learner margin ($+e_i$: Step 5)\\
log-barrier OMD stability & \cref{lem:barrier} & $2\tau_i f_i\E_i[m_i]$ (\cref{eq:Umean}) & learner margin\\
martingale variance (learner) & \cref{lem:freedman} & $3\tau_i\E_i[m_i]$ ($+e_i$) & learner margin ($+e_i$: Step 5)\\
self-normalization bias (comparator) & \ref{A:vector} & $\beta_i\E_i[m_i^u]$ ($+e_i$) & comparator margin ($+e_i$: Step 5)\\
martingale variance (comparator) & \cref{lem:freedman} & $\tau_i\E_i[m_i^u]$ & comparator margin\\
Freedman tails ($J/\eta$ each) & \cref{lem:freedman} & $2J/\eta$ total & additive, once per state\\
implicit-exploration tail & \cref{lem:ix} & $HJ/(2\gamma_0)$ total & additive, once per state\\
log-barrier OMD penalty & \cref{lem:barrier} & $A\log K/\eta$ total & additive, once per state\\
comparator smoothing ($u$ vs.\ $\pi^\star$) & Step 2 & $\Lambda_i H/K$ & additive, $\le H$ total\\
exponentially small remainders $e_i$ & Step 5 & $\Lambda_i H(3\etail+e^{-M/3})$ & additive, $\le H$ total\\
final incomplete active block & Step 1 & $H\Lmax$ & additive, once per state\\
\bottomrule
\end{tabular}
\end{footnotesize}
\end{center}
\end{table}

\paragraph{Step 1 (Episodes outside complete active blocks).}
Whenever $s$ is inactive in episode $k$, $\Phi^k(s)\le 0$ by \cref{lem:identity}. Because all periods except possibly the last have length $\Lmax$ and every active block length $\Lambda_j(s)$ divides $\Lmax$, state $s$ has at most one incomplete active block (truncated at episode $K$), whose episodes contribute at most $H\Lmax$ since $\Phi^k(s)\le H$.

\paragraph{Step 2 (OMD ledger across complete active blocks).}
Recall from \cref{app:events} that $s$ has $n_s<K$ complete active blocks $b_1,\dots,b_{n_s}$, and all block quantities in \cref{eq:blockquantities} vanish for $i>n_s$. On each $b_i$ ($1\le i\le n_s$), $\pi^k(\cdot\mid s)=\pi_i$ and $w^k(s,\cdot)=f_i^2\chi_i$, so $\sum_{k\in b_i}\ip{\pi^k(\cdot\mid s)}{z^k(s,\cdot)} = f_i^2 m_i$. Since $G_i\ge 0$ and $\ip{\Unif(\Aset)}{G_i}\le\Lambda_i H$, comparator smoothing satisfies $m_i^u\le\ip{\pi^\star(\cdot\mid s)}{G_i} + \Lambda_i H/K$ and sums to at most $H\sum_{i=1}^K\Lambda_i/K\le H$ across all blocks. Updates at $s$ occur exclusively at the ends of $b_1,\dots,b_{n_s}$, so \cref{lem:barrier} and \cref{eq:selfnorm_pathwise} telescope over $1\le i\le n_s$ (and trivially when $n_s=0$) while terms with $i>n_s$ vanish:
\[
  \sum_{i=1}^K (1-2\tau_i)\widehat L_i \le \sum_{i=1}^K \ip{u}{\widehat\ell_i} + \frac{A\log K}{\eta}.
\]
Combining these bounds yields
\begin{equation}\label{eq:localledger}
  \sum_{i=1}^{n_s}\sum_{k\in b_i}\Phi^k(s) \le \sum_{i=1}^K\bigl(f_i^2 m_i - m_i^u\bigr) + H
  \le \frac{A\log K}{\eta} + \sum_{i=1}^K W_i + \sum_{i=1}^K\bigl(\ip{u}{\widehat\ell_i} - m_i^u\bigr) + H.
\end{equation}

\paragraph{Step 3 (Learner side).}
On $\Ecl$, \ref{A:scalar} gives at every $\G_i$
\begin{equation}\label{eq:Umean}
  f_i\bigl[(1-\beta_i)\E[m_i\mid\G_i] - e_i\bigr] \le \E[\widehat L_i\mid\G_i] \le f_i\bigl[(1+\beta_i)\E[m_i\mid\G_i] + e_i\bigr].
\end{equation}
Since $1-2\tau_i\ge 0$, the lower bound in \cref{eq:Umean} yields
\[
  \E[W_i\mid\G_i] \le \bigl(f_i^2 - (1-2\tau_i)f_i(1-\beta_i)\bigr)\E[m_i\mid\G_i] + e_i
  \le f_i(-\epsilon_i + \beta_i + 2\tau_i)\E[m_i\mid\G_i] + e_i.
\]
Because $f_i^2 m_i$ and $(1-2\tau_i)\widehat L_i$ both lie in $[0,\Lambda_i H]$, we have $W_i^2\le\Lambda_i H(f_i^2 m_i + (1-2\tau_i)\widehat L_i)$, so the upper bound in \cref{eq:Umean} and $\beta_i\le 0.1$ give
\[
  \Var(W_i\mid\G_i) \le \E[W_i^2\mid\G_i] \le \Lambda_i H\bigl(\E[m_i\mid\G_i] + \E[\widehat L_i\mid\G_i]\bigr) \le \Lambda_i H\bigl(3\E[m_i\mid\G_i] + e_i\bigr).
\]
On $\good_s^\lrn$, $\sum_{i=1}^K\xi_i^\lrn\le\eta\sum_{i=1}^K\Var(W_i\mid\G_i) + J/\eta$. Since $\eta\Lambda_i H = \tau_i\le 1$, we obtain
\[
  \sum_{i=1}^K W_i \le \frac{J}{\eta} + \sum_{i=1}^K\bigl(f_i(-\epsilon_i + \beta_i + 2\tau_i) + 3\tau_i\bigr)\E[m_i\mid\G_i] + 2\sum_{i=1}^K e_i
  \le \frac{J}{\eta} + 2\sum_{i=1}^K e_i,
\]
where the bracketed coefficient is at most $f_i(-\epsilon_i + \frac25\epsilon_i + \frac{2}{1000}\epsilon_i) + \frac{3}{1000}\epsilon_i \le -\frac34\cdot 0.598\epsilon_i + 0.003\epsilon_i \le 0$ by \cref{eq:slack}.

\paragraph{Step 4 (Comparator side).}
Write
\[
  \ip{u}{\widehat\ell_i} - m_i^u = \bigl(\ip{u}{\widehat\ell_i} - f_i\E[\ip{u}{\widehat G_i^{\SN}}\mid\G_i]\bigr)
  + \bigl(f_i\E[\ip{u}{\widehat G_i^{\SN}}\mid\G_i] - \E[m_i^u\mid\G_i]\bigr) + \xi_i^\cmp.
\]
On $\good_s^{\mathrm{IX}}$, the first sum over $i=1,\dots,K$ is at most $HJ/(2\gamma_0)$. By \ref{A:vector} averaged under $u\in\Delta(\Aset)$, the middle term is at most $f_i(1+\beta_i)\E[m_i^u\mid\G_i] + f_i e_i - \E[m_i^u\mid\G_i] \le (-\epsilon_i+\beta_i)\E[m_i^u\mid\G_i] + e_i$. Since $m_i^u\in[0,\Lambda_i H]$, $\Var(m_i^u\mid\G_i)\le\Lambda_i H\E[m_i^u\mid\G_i]$, so on $\good_s^\cmp$ we have $\sum_{i=1}^K\xi_i^\cmp\le\sum_{i=1}^K\tau_i\E[m_i^u\mid\G_i]+J/\eta$. Since $-\epsilon_i+\beta_i+\tau_i\le (-1+\frac25+\frac{1}{1000})\epsilon_i \le 0$, summing over $i=1,\dots,K$ gives
\[
  \sum_{i=1}^K\bigl(\ip{u}{\widehat\ell_i} - m_i^u\bigr) \le \frac{HJ}{2\gamma_0} + \frac{J}{\eta} + \sum_{i=1}^K e_i.
\]

\paragraph{Step 5 (Summing remainders).}
Since the complete blocks of $s$ are disjoint, $\sum_{i=1}^K\Lambda_i\le K$, and \ref{P:tiny} gives
\[
  3\sum_{i=1}^K e_i \le 3 K H(3\etail + e^{-M/3}) \le 12 K\etail H \le H.
\]
Combining Steps 1--5 in \cref{eq:localledger} proves $\sum_{k=1}^K \Phi^k(s) \le \frac{A\log K+2J}{\eta} + \frac{HJ}{2\gamma_0} + H\Lmax + 2H$. \qed

%% file: appendix/I_main_proof.tex
\section{Master Inequality and Main Regret Proof}\label{app:main_proof}

\subsection{Untuned Master Inequality}

Recall from \cref{tab:params} (\cref{app:notation}) the constants $C_0=1000$, $C_1=24$, $\aeta=2HAJ$, $\beeta=C_1 C_0 M H^2 D S K$, $\agam=H^2 J/2$, and $\bgam=C_1 A M H S K$.

\begin{proposition}[Master inequality]\label{prop:master}
Let $\eta,\gamma_0>0$ satisfy $\kappa = C_0\eta H D + A\gamma_0 \le 1/4$, and run \textup{\alg{}} with $a_0=C_0\eta H D$, $M=\lceil 200J\rceil$, and $\Lmax = 2^{\lfloor\log_2(1/(4\kappa))\rfloor}$. With probability at least $1-\delta$,
\begin{equation}\label{eq:master}
  \Reg \le \frac{\aeta}{\eta} + \beeta\,\eta + \frac{\agam}{\gamma_0} + \bgam\,\gamma_0 + 2H^2.
\end{equation}
\end{proposition}

\begin{proof}
Work on the event $\good$, which satisfies $\Prob(\good)\ge 1-\delta$ by \cref{lem:good}.

\emph{Local regrets.} Summing \cref{lem:local} against $\sum_{s\in\Sset}q^\star(s)=H$ in \cref{lem:identity} gives
\begin{equation}\label{eq:aggregate1}
  \Reg \le H\Bigl(\frac{A\log K + 2J}{\eta} + \frac{HJ}{2\gamma_0} + H\Lmax + 2H\Bigr) + \sum_{k=1}^K\sum_{s\in\Sset} q^k(s)\ip{\pi^k(\cdot\mid s)}{r^k(s,\cdot)}.
\end{equation}

\emph{Cost of optimism (\cref{eq:optcost}).} For an active state $s\in\Sset$ in episode $k$ of period $j\ge 2$ with $\Lambda=\Lambda_j(s)$, $f=f_j(s)$, $\gamma=\gamma_j(s)$, and $\pi=\pi^k(\cdot\mid s)$, we have $r^k=(1-w^k)v^k$ and $v^k\le H$, so
\[
  \ip{\pi}{r^k(s,\cdot)} \le H\sum_{a\in\Aset}\pi(a)\Bigl[(1-f^2) + f^2\frac{\gamma}{\pi(a)+\gamma}\Bigr]
  \le H\bigl(2\epsilon_j(s) + A\gamma\bigr) = H\Lambda(2a_0 + A\gamma_0) \le 2H\kappa\Lambda.
\]
On $\Ecl$, $q^k(s)\Lambda\le 5M$, so each active state contributes at most $10HM\kappa$ per episode. Each inactive state $s$ in period $j\ge 2$ contributes at most $H q^k(s)\le 3HM/\Lmax < 24HM\kappa$ since $\Lmax > 1/(8\kappa)$. Summing over all $S$ states in periods $j\ge 2$ gives at most $24H S M\kappa = C_1 H S M\kappa$ per episode. In period $1$, $\sum_{s\in\Sset}q^k(s)\ip{\pi^k(\cdot\mid s)}{r^k(s,\cdot)} = V^{B,k}(s_1)\le V^k(s_1)\le H$ by \cref{eq:VB_unroll}, totaling at most $H\min(K,\Lmax)\le H\Lmax$. This establishes \cref{eq:optcost}.

\emph{Combining terms.} By \ref{P:stab}, $(H+1)H\Lmax \le (H+1)/(4C_0 D\eta) \le (H+1)/\eta$, and since $A\ge 2$ and $J > 4.8 + 3\log K$, we have $H(A\log K + 2J + 1) + 1 \le 2HAJ = \aeta$. Substituting these bounds and \cref{eq:optcost} into \cref{eq:aggregate1} and expanding $\kappa = C_0\eta H D + A\gamma_0$ gives \cref{eq:master}.
\end{proof}

\subsection{Completing the Main Tabular Regret Proof}

\begin{proof}[Proof of \cref{thm:main}]
Choose the exact tuned rates $\eta = \sqrt{\aeta/\beeta}$ and $\gamma_0 = \sqrt{\agam/\bgam}$ from \cref{tab:params}; \cref{tab:params_main} states them up to universal constants. Define
\[
  \mathcal B := 2\sqrt{\aeta\beeta} + 2\sqrt{\agam\bgam} + 2H^2.
\]
If $\kappa = C_0\eta H D + A\gamma_0 \le 1/4$, \cref{prop:master} implies $\Reg\le\mathcal B$ with probability at least $1-\delta$. If instead $\kappa > 1/4$, the learner plays the uniform policy throughout, and since $S\ge H\ge 1$,
\[
  \mathcal B \ge \beeta\eta + \bgam\gamma_0 = C_1 H S M K(C_0\eta H D + A\gamma_0) = C_1 H S M K\kappa > 6 H S M K \ge HK \ge \Reg
\]
deterministically. Finally, since $\aeta = 2HAJ$, $M = O(J)$, and $S\ge H$, we have
\[
  \sqrt{\aeta\beeta} = O\bigl(\sqrt{HAJ\cdot J H^2 D S K}\bigr) = O\bigl(H^{3/2}\sqrt{D}\,J\sqrt{SAK}\bigr),\qquad
  \sqrt{\agam\bgam} = O\bigl(H^{3/2} J\sqrt{SAK}\bigr),
\]
and $2H^2 \le 2H^{3/2}\sqrt{S} \le H^{3/2}\sqrt{SAK}$ since $A,K\ge 2$. Since $1\le D=\min\{H,\sqrt{HJ}\}$, this yields the bound of \cref{thm:main} in both of its forms; moreover $J< 6\log(SAK/\delta)$. Memory and runtime bounds are proved in \cref{app:extensions}.
\end{proof}

\begin{remark}[Constants]\label{rem:constants}
The absolute constants of \cref{tab:params} were chosen for a transparent proof and were not optimized. With them, $\mathcal B<HK$ only for $K$ above about $K_1:=4\cdot 10^7J^2HDSA$, and for $K$ below about $3\cdot 10^5J^2AHD/S$ no state can become active, so that \alg{} is the uniform policy. In terms of the parameters, the bound is thus nontrivial from $K=\tOmega(HDSA)$ on, whereas the additive term $H^4S^2A$ of the earlier bound is dominated from $K=H^3S^3A$ on; with the constants above the first threshold is the smaller one when $HS\gtrsim 6\cdot 10^3J\sqrt D$. Since $J>85$ whenever $K\ge K_1$, the regime $D<H$ of \cref{thm:main} then requires $H>85$. \cref{thm:main} is a statement about the dependence of the regret on $S$, $A$, $H$ and $K$. The constants of \linblkalg{} are far larger (end of \cref{sec:lin_alg_full}).
\end{remark}

%% file: appendix/J_extensions.tex
\section{Extensions and Computational Complexity}\label{app:extensions}

\paragraph{Regret on realized trajectory costs.}
Let $\widehat{\mathrm{Reg}}_K := \sum_{k=1}^K\bigl(\sum_{h=1}^H c^k(s_h^k,a_h^k) - V^{\pi^\star}(s_1;c^k)\bigr)$. The difference $\widehat{\mathrm{Reg}}_K - \Reg = \sum_{k=1}^K\bigl(\sum_{h=1}^H c^k(s_h^k,a_h^k) - V^k(s_1)\bigr)$ is a sum of martingale differences with increments in $[-H,H]$, so the one-sided Hoeffding--Azuma inequality gives $\widehat{\mathrm{Reg}}_K \le \Reg + H\sqrt{2K\log(2/\delta)}$ with probability at least $1-\delta/2$. Running \alg{}, \aggalg{}, or \linblkalg{} at failure level $\delta/2$ and taking a union bound, the bounds of \cref{thm:main,thm:agg_main,thm:linear_block} hold for $\widehat{\mathrm{Reg}}_K$ with probability at least $1-\delta$, up to constants.

\paragraph{Random initial state.}
An unknown initial-state distribution $q_1\in\Delta(\Sset_1)$ reduces to a singleton initial layer by prepending a dummy layer-$0$ state $s_0$ with zero cost whose transitions under every action follow $q_1$ (in a linear AMDP, augmenting features in $\R^{d+1}$ with $(0,1)$ at $s_0$ and $(\phi(s,a),0)$ on layers $1,\dots,H$ preserves \cref{def:linear} with dimension $d+1$ since $\sum_{s'\in\Sset_1}q_1(s')=1$). Replacing $(H,S,d)$ by $(H+1,S+1,d+1)$ leaves all asymptotic bounds unchanged, and none of our algorithms needs prior knowledge of $q_1$ because all actions at $s_0$ are identical.

\paragraph{Computing the log-barrier update and $O(HA)$ amortized runtime.}
Both \alg{} (\cref{alg:snpo}) and \aggalg{} (\cref{alg:agg}) sample $\tau^k$ in $O(HA)$ time and run their backward sweep in $O(H)$ operations per episode. To avoid $O(S)$ scans at period boundaries when $S\gg K$, initialize $\pi(\cdot\mid s)\equiv\Unif(\Aset)$ lazily on first visit (with a hash table over visited states, in expected constant time per access, if $S$ is too large to allocate arrays), and store for each state $s$ a period count $N(s)$ and stamp $\mathrm{active\_period}[s]$ (initially $0$). During period $j$, append $s$ to the visited list $\mathcal V_j$ when $N(s)$ transitions $0\to 1$. At the start of period $j\ge 2$, scan only $s\in\mathcal V_{j-1}$: if $N(s)\ge M$, set $\mathrm{active\_period}[s]\leftarrow j$, compute $s$'s period-$j$ parameters, and zero $N_b$ and $\Sigma_b(\cdot)$ (forming $\mathcal P_j^{\mathrm{act}}$), then reset $N(s)\leftarrow 0$ across $\mathcal V_{j-1}$. Any previously active state $s\notin\mathcal V_{j-1}$ had $N_{j-1}(s)=0<M$ and fails the check $\mathrm{active\_period}[s]=j$ in $O(1)$ time without a scan. Across all period transitions, $\sum_{j=2}^{\lceil K/\Lmax\rceil}|\mathcal V_{j-1}| \le (\lceil K/\Lmax\rceil-1)H\Lmax < HK$ and $\sum_{j\ge 2}|\mathcal P_j^{\mathrm{act}}| < HK/M$. Each $s\in\mathcal P_j^{\mathrm{act}}$ has $\Lambda_j(s)\ge M\Lmax/N_{j-1}(s)$ by \cref{eq:blocklength} and completes at most $\lfloor\Lmax/\Lambda_j(s)\rfloor \le N_{j-1}(s)/M$ blocks in period $j$, for fewer than $HK/M$ complete active blocks across all $K$ episodes. Complete blocks with $N_b=0$ have $\widehat\ell_b\equiv 0$ and leave $\pi(\cdot\mid s)$ unchanged, so they need not be scheduled. Pushing $s$ into a min-heap (of size at most $S$) keyed by its block-end episode $k_{\mathrm{end}}\le K$ when $N_b$ transitions $0\to 1$ and popping at episode $k_{\mathrm{end}}$ incurs at most $2HK/M$ heap operations of cost $O(\log(1+S))=O(M)$, or $O(HK)$ total scheduling work. Zeroing $\Sigma_b(\cdot)$ across $\mathcal P_j^{\mathrm{act}}$ and completed visited blocks takes $O((HK/M)A)$ total time.

For each scheduled update with $\widehat L_b = \ip{\pi_b}{\widehat\ell_b} > 0$, we apply bisection on the bracket $I=[\lambda_0-r_I,\lambda_0+r_I]$ from the proof of \cref{lem:barrier} (\cref{app:omd}). On $I$, every $p_\lambda(a)=\pi_b(a)/(1+\zeta_a(\lambda))$ satisfies $|\zeta_a(\lambda)|\le 3/32$ and $0 < \partial_\lambda p_\lambda(a) \le 4\eta\pi_b(a)^2$. Normalizing $\tilde\pi_\lambda := p_\lambda/\|p_\lambda\|_1$ at the midpoint $\lambda$ of the bracket obtained after $T_{\mathrm{bis}}$ bisection steps gives $\|\tilde\pi_\lambda-\pi'\|_1 \le 2\|p_\lambda-\pi'\|_1 \le 8\eta\Sigma_2 r_I 2^{-T_{\mathrm{bis}}} = 256(\eta\widehat L_b)^2\pi_{\max} 2^{-T_{\mathrm{bis}}} \le \frac14 2^{-T_{\mathrm{bis}}}$, since $\eta\widehat L_b\le 2.5\cdot 10^{-4}$ in both \alg{} (\ref{P:stab}) and \aggalg{} (\ref{aggQ:stab}). Achieving $\ell_1$ accuracy $\xi\in(0,1/2)$ therefore takes $O(A\log(1/\xi))$ operations per update, or $O((HA/M)\log(1/\xi))$ amortized time per episode. Any polynomial accuracy $\xi=(SAK/\delta)^{-\Theta(1)}$ satisfies $\log(1/\xi)=O(M)$, which gives $O(HA)$ amortized time per episode and $O(SA)$ space. The regret analysis of \cref{app:omd,app:local} is written for the exact minimizer of \cref{eq:omd_update}. The following remark shows that it also covers the output of the bisection, with no change in the constants of the drift analysis.

\begin{remark}[Inexact root-finding]\label{rem:inexact}
Consider an update with $\widehat L_b>0$, and recall from the proof of \cref{lem:barrier} the function $\Psi(\lambda)=\sum_a p_\lambda(a)$ with $p_\lambda(a)=\pi_b(a)/(1+\zeta_a(\lambda))$ and the bracket $I$, at whose endpoints $\Psi(\lambda_0-r_I)<1<\Psi(\lambda_0+r_I)$. Run $T:=\lceil\log_2(18AK)\rceil$ bisection steps on $I$, maintaining $\Psi(\lambda_{\mathrm{lo}})\le 1\le\Psi(\lambda_{\mathrm{hi}})$, and output $\tilde\pi:=p_{\hat\lambda}/Z$ with $\hat\lambda:=\lambda_{\mathrm{lo}}$ and $Z:=\Psi(\hat\lambda)\le 1$. Then:
\begin{enumerate}[label=(\roman*),leftmargin=*]
\item $\tilde\pi$ is the \emph{exact} minimizer of \cref{eq:omd_update} for the perturbed loss $\tilde\ell(a):=\widehat\ell_b(a)-(1-Z)/(\eta\,p_{\hat\lambda}(a))\le\widehat\ell_b(a)$. Indeed, $1/p_{\hat\lambda}(a)=1/\pi_b(a)+\eta(\widehat\ell_b(a)-\hat\lambda)$ gives $1/\tilde\pi(a)-1/\pi_b(a)=\eta(\tilde\ell(a)-\hat\lambda)$, which is the optimality condition of \cref{eq:omd_update} for $\tilde\ell$ with multiplier $\hat\lambda$.
\item On $I$ we have $|\zeta_a|\le 3/32$ and hence $0<\Psi'\le(32/29)^2\eta\Sigma_2\le 4\eta\Sigma_2$, so $0\le 1-Z\le 4\eta\Sigma_2\cdot 2r_I2^{-T}=256(\eta\widehat L_b)^2\pi_{\max}2^{-T}\le 8\eta\widehat L_b\,2^{-T}$.
\item $|\log(1+\zeta_a(\hat\lambda))|\le|\zeta_a|/(1-|\zeta_a|)\le\frac{96}{29}\eta\widehat L_b$ and $|\log Z|\le 2(1-Z)$, so $|\log(\tilde\pi(a)/\pi_b(a))|\le(\tfrac{96}{29}+16\cdot 2^{-T})\eta\widehat L_b\le 6\eta\widehat L_b$. The per-update bound used in \cref{lem:dyadic,lem:drift} therefore holds verbatim.
\item Let $\delta_b:=\ip{\pi_b}{\widehat\ell_b-\tilde\ell}=\frac{1-Z}{\eta}\sum_a(1+\zeta_a(\hat\lambda))\le 9A\,\widehat L_b\,2^{-T}$. The policies actually played at $s$ are exact OMD iterates for the losses $(\tilde\ell_i)_i$, and $\eta\sum_a\pi_i(a)|\tilde\ell_i(a)|\le\eta(\widehat L_i+\delta_i)<1/32$, so \cref{eq:omd_bound} applies to them. Since $0\le\widehat\ell_i-\tilde\ell_i$ and $\pi_i(a)|\tilde\ell_i(a)|\le\max\{\pi_i(a)\widehat\ell_i(a),\pi_i(a)(\widehat\ell_i-\tilde\ell_i)(a)\}$, this gives
\[
  \sum_i\ip{\pi_i-u}{\widehat\ell_i}\le\frac{A\log K}{\eta}+2\eta\sum_i\sum_a\pi_i(a)^2\widehat\ell_i(a)^2+\sum_i\bigl(\delta_i+2\eta\delta_i^2\bigr)\le\frac{A\log K}{\eta}+2\sum_i\tau_i\widehat L_i+2\sum_i\delta_i ,
\]
and $2\sum_i\delta_i\le 18A\,2^{-T}H\sum_i\Lambda_i\le H$.
\end{enumerate}
All other steps of the analysis refer to $\widehat\ell_i$, to the policies actually played, and to the drift budget only. Consequently \cref{lem:local} holds for the bisection output with $2H$ replaced by $3H$, and \cref{eq:master} with $2H^2$ replaced by $3H^2$, at $O(A\log(AK))=O(AM)$ operations per update. The same remark applies to \aggalg{} with $H$ replaced by $B$.
\end{remark}

\begin{remark}[Unknown number of episodes]\label{rem:doubling}
If $K$ is not known in advance, run \alg{} in phases $i=1,2,\dots$, where phase $i$ consists of the next $2^i$ episodes and runs a fresh copy of \alg{} with the parameters of \cref{tab:params} for $2^i$ episodes and confidence level $\delta_i:=\delta/(2i^2)$. A copy whose phase is cut short by the end of the interaction behaves as it would on a sequence of $2^i$ cost functions that ends with zero costs. Appending zero costs changes neither the regret of the copy nor the set of best policies of the phase, so \cref{thm:main} applies to every phase, and the regret of a phase against $\pi^\star$ is at most its regret against the best policy of the phase. Since $\sum_i\delta_i<\delta$ and at most $\lceil\log_2(K+1)\rceil$ phases start, a union bound gives, with probability at least $1-\delta$,
\[
  \Reg\le\sum_{i=1}^{\lceil\log_2(K+1)\rceil}O\Bigl(H^{3/2}\sqrt{D\,SA\,2^i}\,\log\frac{SA\,2^i}{\delta_i}\Bigr)=O\Bigl(H^{3/2}\sqrt{D\,SAK}\,\log\frac{SAK}{\delta}\Bigr),
\]
with $D$ evaluated for the longest phase, which changes it by at most a constant factor. The same argument applies to \aggalg{} and \linblkalg{}.
\end{remark}

\begin{remark}[Adaptive adversaries]\label{rem:adaptive}
The proofs use that the costs are fixed in advance in three places: the comparator $\pi^\star$ and its occupancy $q^\star$ are deterministic, the smoothed comparator $u$ in the applications of \cref{lem:freedman,lem:ix} is deterministic, and the frozen targets $v^{\circ,k}$ of \cref{lem:drift} are $\F_{t_0-1}$-measurable. The third use is tied to the estimator, as the following heuristic calculation shows. Consider a single state with frozen policies, so that its visits $X_1,\dots,X_\Lambda$ in a block are i.i.d.\ Bernoulli with mean $q$, write $N_k:=\sum_{i\le k}X_i$, and let the target of episode $k$ be $g_k:=\ind\{N_{k-1}\le q(k-1)\}$: the cost is one when the block is behind its expected visit count and zero when it is ahead, which an adversary that observes the previous episodes can arrange. Given the first $k-1$ episodes, $\E[\Lambda X_k/N_\Lambda]$ is approximately $q\Lambda/(N_{k-1}+q(\Lambda-k+1))$, which exceeds one when the block is behind. To first order in the deviation $S_{k-1}:=N_{k-1}-q(k-1)$, whose negative part has mean about $\sqrt{q(1-q)(k-1)/(2\pi)}$,
\[
  \E\Bigl[\frac{\Lambda}{N_\Lambda}\sum_{k=1}^\Lambda X_kg_k-\sum_{k=1}^\Lambda g_k\Bigr]\approx\frac{1}{q\Lambda}\sum_{k=1}^{\Lambda}\E\bigl[\max\{-S_{k-1},0\}\bigr]\approx\frac23\sqrt{\frac{(1-q)\Lambda}{2\pi q}},
\]
while $\E[\sum_kg_k]\approx\Lambda/2$. The relative bias is therefore about $0.53\sqrt{(1-q)/(q\Lambda)}$, of order $M^{-1/2}$ for a block with $q\Lambda=\Theta(M)$ expected visits, and the opposite rule (cost one when the block is ahead) gives the opposite sign. An adversary that makes one action costly when the block is behind and another when it is ahead can thus shift the estimated difference between two actions by a relative amount of order $M^{-1/2}$, whereas the margin $\epsilon_b=a_0\Lambda_b$ of a frequently visited state is of order $M\sqrt{AHD/(SK)}$ and vanishes as $K$ grows. This calculation concerns the estimator and is not a lower bound on the regret of \alg{}.
\end{remark}

%% file: appendix/K_horizon.tex
\section{Where the Horizon Dependence Comes From}\label{app:horizon}

This appendix explains where each factor of $H^{7/4}$ enters \emph{our analysis} of \alg{}, and why three natural modifications of the algorithm do not improve the exponent within this analysis. We do not prove a lower bound for the algorithm, and the statements below about alternatives are heuristic except where a computation is given.

\paragraph{The two places where the drift factor enters.}
Up to logarithmic factors, \cref{prop:master} balances the $q^\star$-weighted local regrets, of order $HAJ/\eta$, against the $q^k$-weighted cost of optimism, of order $HSMK\kappa$ with $\kappa\ge C_0\eta HD$, which gives the bound $O(H^{3/2}\sqrt{D\,SAJKM})$. The drift factor $D=\min\{H,\sqrt{HJ}\}$ enters $\kappa$ in two ways.
\begin{enumerate}[label=(\roman*),leftmargin=*]
\item \emph{Margin for the self-normalization bias.} By \ref{A:vector}, \ref{A:scalar} and \cref{tab:budget}, the proof needs the margin $\epsilon_b=a_0\Lambda_b$ to dominate the relative bias coefficient $\beta_b=5\varrho_b=400\eta HD\Lambda_b$ of \cref{lem:drift}, which we enforce with $a_0=C_0\eta HD$.
\item \emph{Stability across two periods and truncation at inactive states.} The classification of states (\cref{app:classification}) uses that occupancies vary by at most a constant factor over two consecutive periods, $\rho_P=76\eta HD\Lmax\le 0.02$, so $\Lmax=O(1/(\eta HD))$. An inactive state has $w^k=0$ and is charged up to $Hq^k(s)\le 3HM/\Lmax$ per episode in \cref{eq:optcost}, which is of order $\eta H^2DM$.
\end{enumerate}

\paragraph{The path-law bound is tight for the occupancies of rarely visited states.}
For two full-support Markov policies $\pi,\pi'$ let $\vartheta:=\max_{s',a'}|\log(\pi'(a'\mid s')/\pi(a'\mid s'))|\le 1$. Consider a layered chain whose state at layer $h$ records the number $m\in\{0,1,\dots,h-1\}$ of ``up'' actions taken before layer $h$, and shift every local policy from $\pi(\mathrm{up}\mid s)=1/2$ to $\pi'(\mathrm{up}\mid s)=(1+\upsilon)/2$, which changes each per-step log-probability by $\upsilon+O(\upsilon^2)$. At the state with count $m=(h-1)/2+t$ in layer $h$,
\[
  \log\frac{q^{\pi'}(s)}{q^\pi(s)} = m\log(1+\upsilon) + (h-1-m)\log(1-\upsilon) = 2\upsilon t - \frac{h-1}{2}\upsilon^2 + O(t\upsilon^3+h\upsilon^4),
\]
whereas $q^\pi(s)=\binom{h-1}{m}2^{-(h-1)}$ satisfies $\log(1/q^\pi(s)) = 2t^2/(h-1) + \frac12\log h + O(1)$ for $t=o(h^{3/4})$. A state with occupancy $q^\pi(s)=e^{-y}$, where $\log h\ll y\ll h^{1/2}$, has $t\approx\sqrt{hy/2}$, so for $0<\upsilon\ll\sqrt{y/h}$ its occupancy changes by a factor $e^{\Theta(\upsilon\sqrt{hy})}$ under per-step log-ratios of order $\upsilon$. This matches the radius of \cref{lem:pathlaw} up to constants. The computation shows that the bound of \cref{lem:pathlaw} cannot be improved for the occupancies of states visited with probability $e^{-y}$. It does not show that the iterates of \alg{} realize such a coherent shift, and it does not imply a lower bound on the regret of \alg{}.

\paragraph{Three modifications that do not help within this analysis.}
\begin{enumerate}[label=(\alph*),leftmargin=*]
\item \emph{The range of the returns.} Without a transition model with which to propagate one-step backups $c^k+pV^k$ across actions, the local learners apply implicit exploration to sampled suffix returns $Y_h^k\in[0,H]$ (\cref{lem:ix}). Balancing the resulting term $H^2J/\gamma_0$ over the $H$ layers against the cost $\gamma_0HSAMK$ of optimism already yields $\tO(H^{3/2}\sqrt{SAK})$, even with $D=1$. When the transitions are known, a variance-aware bonus removes a factor $\sqrt H$ from policy optimization bounds \citep{schlisselberg2026improved}; we do not know whether it can be combined with block self-normalization.
\item \emph{Centering the losses.} Adding a multiple of the all-ones vector to $\widehat\ell_b$ leaves the log-barrier update unchanged, and \cref{lem:barrier} applied to the centered loss bounds $|\log(\pi'(a)/\pi_b(a))|$ by $6\eta\sum_{a'}\pi_b(a')|\widehat\ell_b(a')-\widehat L_b|$. This does not reduce the pathwise bound. Even when all returns equal the same value $Y\approx H$ (so that the true action values coincide), inverse-propensity weighting at $\pi_b=\Unif(\Aset)$ over $N_b\in[M,2M)$ visits gives, in expectation,
\[
  \sum_{a'\in\Aset}\pi_b(a')\bigl|\widehat\ell_b(a')-\widehat L_b\bigr|
  \asymp \Lambda_b H\sum_{a'\in\Aset}\Bigl|\frac{N_b(a')}{N_b}-\frac{1}{A}\Bigr|
  = \Theta\bigl(\Lambda_b H\min\{1,\sqrt{A/M}\}\bigr),
\]
which is of order $\Lambda_bH$ whenever $A\ge M$, and on some sample paths the sum is of order $\Lambda_bH$ for every $M$. For $A\ll M$, increasing $M$ by a factor $H^\alpha$ shrinks the typical fluctuation by $H^{-\alpha/2}$ but multiplies the cost $HSMK\kappa$ of optimism by $H^\alpha$, which is a net loss.
\item \emph{Baselines and exponential weights.} Subtracting an empirical baseline from the returns leaves fluctuations of the same order $\Lambda_bH\min\{1,\sqrt{A/M}\}$ when the transitions lead with probability $1/2$ to a zero-cost and to a unit-cost continuation, and it destroys the nonnegativity $Y_h^k\ge 0$ that \cref{lem:relative_sn,lem:ix} use. Replacing log-barrier OMD by exponential weights makes the pathwise log-ratio of an update as large as $\eta\max_a\widehat\ell_b(a)$, which can be of order $\eta H/\gamma_0$.
\end{enumerate}

%% file: appendix/L_aggregate.tex
\section{Analysis for Aggregate Feedback}\label{app:aggregate}

This appendix specifies \aggalg{} in full (\cref{alg:agg}), states and proves the surcharge decomposition (\cref{lem:agg_identity}), proves \cref{thm:agg_main}, and proves two propositions on the limits of the construction (\cref{prop:agg_transfer,prop:agg_barrier}).

\subsection{The Algorithm}\label{sec:agg_alg_full}

\aggalg{} keeps the period schedule, the threshold $M$, the block lengths $\Lambda_j(s)$ of \cref{eq:blocklength}, and the OMD update \eqref{eq:omd_update} of \alg{}. Define the surcharge schedule $\Rb_h := 2H(H-h+1)$ for $h\in[H]$ and $\Rb_{H+1}:=0$, so that $\Rb_h-\Rb_{h'}=2H(h'-h)$, and the return bound $B:=2H^2+H$. Let $\bar a := a_0+4A\gamma_0$ and $\kagg := a_0+5A\gamma_0$, and let the period length satisfy $\kagg\Lmax\le 1/4$. In period $j\ge 2$, every active state $s$ sets
\[
  \epsilon_j(s):=\bar a\,\Lambda_j(s)\le\tfrac14,\qquad f_j(s):=1-\epsilon_j(s),\qquad \gamma_j(s):=\gamma_0\Lambda_j(s),\qquad \chi^k(s,a):=\frac{\pi^k(a\mid s)}{\pi^k(a\mid s)+\gamma_j(s)},
\]
and has death probability $\omega^k(s,a):=1-f_j(s)^2\chi^k(s,a)$; inactive states have $\omega^k(s,\cdot):=1$. The extra term $4A\gamma_0$ in $\bar a$ lets the margin absorb the implicit-exploration bias of order $HA\gamma_j(s)$ on the undiscounted total $C^k\le H$. After observing $(\tau^k,C^k)$, the algorithm draws a coin $\varsigma_h^k\in\{0,1\}$ for each layer $h\in\{2,\dots,H\}$; given the history and $\tau^k$, the coins are independent and $\varsigma_h^k=0$ with probability $\omega^k(s_h^k,a_h^k)$. It sets
\[
  \sigma_h^k:=\min\bigl\{h'\in\{h+1,\dots,H\}:\varsigma_{h'}^k=0\bigr\}\qquad(\min\emptyset:=H+1).
\]
A visit to an active state at layer $h$ in period $j\ge 2$ records the return \eqref{eq:agg_loss}, which in this notation reads $Y_h^k=C^k+f_j(s_h^k)(\Rb_h-\Rb_{\sigma_h^k})\in[0,B]$, and at the end of a block $b$ the state forms
\begin{equation}\label{eq:agg_estimator}
  \widehat\ell_b(a):=\frac{\Lambda_b}{N_b}\,\frac{\Sigma_b(a)}{\pi_b(a)+\gamma_b}\qquad(\widehat\ell_b:=0\text{ if }N_b=0).
\end{equation}
The unit weight on $C^k$ makes the prefix cost cancel between $\pi^k$ and $\pi^\star$, keeps $\ip{\pi_b}{\widehat\ell_b}\le\Lambda_bB$ on every sample path, and induces a bonus of at least $2H\ge Q^k$ at inactive states.

\begin{algorithm}[ht]
\caption{\aggalg{} for tabular AMDPs with aggregate feedback}
\label{alg:agg}
\begin{small}
\begin{algorithmic}[1]
\STATE \textbf{Input:} $\eta,\gamma_0,a_0>0$, threshold $M$, surcharges $\{\Rb_h\}_{h=1}^{H+1}$, period length $\Lmax=2^m$ with $\kagg\Lmax\le 1/4$, where $\bar a = a_0+4A\gamma_0$ and $\kagg = a_0+5A\gamma_0$. If $\kagg>1/4$, play $\Unif(\Aset)$ in all episodes.
\STATE Initialize $\pi(\cdot\mid s)\leftarrow\Unif(\Aset)$ for all $s\in\Sset$.
\FOR{period $j=1,2,\dots,\lceil K/\Lmax\rceil$}
  \STATE Classify the states and set $\Lambda_j(s)$ as in lines~4--7 of \cref{alg:snpo}; an active $s$ uses $\epsilon_j(s)=\bar a\Lambda_j(s)$, $f_j(s)=1-\epsilon_j(s)$ and $\gamma_j(s)=\gamma_0\Lambda_j(s)$.
  \FOR{episode $k=(j-1)\Lmax+1,\dots,\min\{j\Lmax,K\}$}
    \STATE Play $\pi^k=\pi$; observe $(\tau^k,C^k)$; set $N_j(s_h^k)\leftarrow N_j(s_h^k)+1$ for all $h\in[H]$.
    \IF{$j\ge 2$}
      \STATE $\sigma_{\mathrm{next}}\leftarrow H+1$.
      \FOR{$h=H,H-1,\dots,1$}
        \STATE $\sigma_h^k \leftarrow \sigma_{\mathrm{next}}$.
        \IF{$s_h^k$ is inactive}
          \STATE $\sigma_{\mathrm{next}}\leftarrow h$.
        \ELSE
          \STATE $Y_h^k \leftarrow C^k + f_j(s_h^k)(\Rb_h-\Rb_{\sigma_h^k})$; $N_b\leftarrow N_b+1$; $\Sigma_b(a_h^k)\leftarrow\Sigma_b(a_h^k)+Y_h^k$ (for the current block $b$ of $s_h^k$).
          \STATE If $h\ge 2$: draw $\varsigma_h^k\sim\mathrm{Bern}(1-\omega^k(s_h^k,a_h^k))$ and set $\sigma_{\mathrm{next}}\leftarrow h$ if $\varsigma_h^k=0$.
        \ENDIF
      \ENDFOR
      \FORALL{active $s$ with $k\bmod\Lambda_j(s)=0$}
        \STATE Form $\widehat\ell_b$ by \cref{eq:agg_estimator}; update $\pi(\cdot\mid s)$ by \cref{eq:omd_update}; reset $N_b,\Sigma_b\leftarrow 0$.
      \ENDFOR
    \ENDIF
  \ENDFOR
\ENDFOR
\end{algorithmic}
\end{small}
\end{algorithm}

\paragraph{Augmented probability space and parameters.}
In each episode $k$ of a period $j\ge 2$, after observing $(\tau^k,C^k)$, \aggalg{} draws for every layer $h\in\{2,\dots,H\}$ at which $s_h^k$ is active a coin $\varsigma_h^k\in\{0,1\}$ that equals $0$ with conditional probability $\omega^k(s_h^k,a_h^k)$ given the history before episode $k$, the trajectory $\tau^k$ and the other coins of the episode. For notational convenience we set $\varsigma_h^k:=0$ for $h\ge 2$ whenever $s_h^k$ is inactive (in particular throughout period~$1$), consistently with $\omega^k\equiv 1$ there, and $\varsigma_1^k:=1$ in every episode; the latter is never used, because $\sigma_h^k$ depends only on $(\varsigma_{h'}^k)_{h'>h}$. With these conventions $\varsigma^k = (\varsigma_1^k,\dots,\varsigma_H^k)\in\{0,1\}^H$ is defined in every episode. We work on the finite canonical sample space $\Omagg := \bigl((\Sset\times\Aset)^H\times\{0,1\}^H\bigr)^K$ with filtration $\Fagg_k := \sigma\bigl((\tau^1,\varsigma^1),\dots,(\tau^k,\varsigma^k)\bigr)$ for $k\in[K]$ and $\Fagg_0 := \{\emptyset,\Omagg\}$, so $\pi^k$ is $\Fagg_{k-1}$-measurable. Because $c^1,\dots,c^K$ are fixed prior to episode~$1$ (\cref{sec:setting}), $C^k = \sum_{h=1}^H c^k(s_h^k,a_h^k)$ is a deterministic function of $\tau^k$. Throughout this appendix, $\eta,\gamma_0,a_0,\Lmax,\epsilon_j,f_j,\gamma_j,Y_h^k,\widehat\ell_b$ denote their \aggalg{} definitions collected in \cref{tab:agg_params}.

\begin{table}[ht]
\caption{Parameters of \aggalg{} ($S=\sum_{h=1}^H|\Sset_h|\ge H$, $A=|\Aset|\ge 2$, $H\ge 1$, $K\ge 2$, $\delta\in(0,1)$; all logarithms are natural).}
\label{tab:agg_params}
\begin{center}
\begin{footnotesize}
\setlength{\tabcolsep}{4pt}
\begin{tabular}{lll}
\toprule
Symbol & Value & Role\\
\midrule
$J$ & $\log(64SAK^3/\delta)$ & log-confidence factor ($J>\log 1024 > 6.93$)\\
$M$ & $\lceil 200J\rceil$ & visit threshold per active period ($M\ge 1387$)\\
$D$ & $\min\{H,\sqrt{HJ}\}\ge 1$ & horizon drift factor (\cref{lem:agg_drift})\\
$\Rb_h,\ B$ & $2H(H-h+1)\ (\Rb_{H+1}=0),\ B=2H^2+H$ & layer-$h$ surcharge schedule and return bound\\
$\etail$ & $e^{-20J}$ & additive path-law tail (analysis only)\\
$\tilde C_0,\ \tilde C_1$ & $4000,\ 48$ & margin and cost-of-optimism constants\\
$\taeta,\ \tbeeta$ & $2HAJ,\ \ \tilde C_1\tilde C_0 M S H^2 B D K$ & coefficients of $1/\eta$ and $\eta$ (\cref{prop:agg_master})\\
$\tagam,\ \tbgam$ & $H B J/2,\ \ 5\tilde C_1 A M S H^2 K$ & coefficients of $1/\gamma_0$ and $\gamma_0$ (\cref{prop:agg_master})\\
$\eta,\ \gamma_0$ & $\sqrt{\taeta/\tbeeta},\ \ \sqrt{\tagam/\tbgam}$ & exact tuned OMD step size and IX rate\\
$a_0,\ \bar a$ & $\tilde C_0\eta B D,\ \ a_0 + 4A\gamma_0$ & drift margin rate and active-margin rate\\
$\kagg$ & $\bar a + A\gamma_0 = a_0 + 5A\gamma_0$ & total discount rate (play $\Unif(\Aset)$ if $\kagg>1/4$)\\
$\Lmax$ & $2^{\lfloor\log_2(1/(4\kagg))\rfloor}\in(\frac{1}{8\kagg},\frac{1}{4\kagg}]$ ($\kagg\le 1/4$) & dyadic period length\\
$\Lambda_j(s)$ & \cref{eq:blocklength}, $M\le\Lambda_j(s)\le\Lmax$ & dyadic block length of active $s$ ($j\ge 2$)\\
$\epsilon_j,f_j,\gamma_j$ & $\bar a\Lambda_j,\ 1-\epsilon_j,\ \gamma_0\Lambda_j$ & block margin, survival, and IX parameter\\
\bottomrule
\end{tabular}
\end{footnotesize}
\end{center}
\end{table}

\paragraph{Elementary parameter inequalities.}
Because $S\ge H\ge 1$, $A\ge 2$, $K\ge 2$, and $\delta\in(0,1)$, we have $64SAK^3/\delta \ge 1024$, so $J > 6.93$ and $J < 6\log(SAK/\delta)$. Whenever $\kagg\le 1/4$, for every $\Lambda_b\in[1,\Lmax]$ (with $\epsilon_b=\bar a\Lambda_b$, $f_b=1-\epsilon_b$, $\gamma_b=\gamma_0\Lambda_b$) the following bounds hold.
\begin{enumerate}[label=(Q\arabic*),leftmargin=*]
\item\label{aggQ:margin} $\epsilon_b = \bar a\Lambda_b \le \kagg\Lmax \le \frac14$, so $f_b\in[\frac34,1)$. Likewise $5A\gamma_b \le \kagg\Lmax \le \frac14$ (so $A\gamma_b\le\frac14$) and $\epsilon_b \ge 4A\gamma_b$.
\item\label{aggQ:stab} $\tau_b := \eta B\Lambda_b \le \eta B\Lmax \le \frac{\eta B}{4\kagg} \le \frac{1}{4\tilde C_0 D} \le 6.25\times 10^{-5} < \frac{1}{32}$.
\item\label{aggQ:radius} With $\tilde\varrho_b := 161\eta B D\Lambda_b \le \frac{161}{4\tilde C_0} < 0.011$ and $\beta_b := 5\tilde\varrho_b = 805\eta B D\Lambda_b < 0.055$, monotonicity of $t\mapsto (e^{4t}-1)/t$ on $(0,\infty)$ gives $e^{4\tilde\varrho_b}-1 < 4.10\,\tilde\varrho_b < \beta_b$.
\item\label{aggQ:slack} Since $D\ge 1$ and $\tilde C_0=4000$, $\beta_b + 5\tau_b \le 810\eta B D\Lambda_b = \frac{810}{4000}a_0\Lambda_b < \frac14\epsilon_b < \frac23\epsilon_b$.
\item\label{aggQ:tiny} $\etail = e^{-20J}$ satisfies $e^{-2J} < \frac{1}{4096 K^6}$, $12 K^2\etail < 10^{-55}$, and $e^{-M/3}\le \etail$ since $M\ge 200J$.
\end{enumerate}

\begin{remark}[Derandomized surcharges]\label{rem:derandomize}
Given the history and the trajectory $\tau^k$, the conditional expectation of $\sigma_h^k-h$ is
\[
  T_h^k:=1+\sum_{h'=h+1}^{H}\ \prod_{m=h+1}^{h'}\bigl(1-\omega^k(s_m^k,a_m^k)\bigr),
\]
which the backward sweep $T_h^k=1+(1-\omega^k(s_{h+1}^k,a_{h+1}^k))\,T_{h+1}^k$ with $T_H^k:=1$ computes without drawing coins. Recording $\bar Y_h^k:=C^k+2H f_j(s_h^k)\,T_h^k$ in place of \eqref{eq:agg_loss} gives a deterministic function of $\tau^k$ and of the weights of episode $k$. The algorithm that records $\bar Y_h^k$ draws no coins, so we argue within it directly. Since $1\le T_h^k\le H-h+1$, the recorded return lies in $[0,B]$, as $Y_h^k$ does. By backward induction on the sweep, $\E[2H\,T_h^k\mid\F_{k-1},s_h^k=s,a_h^k=a]=\Delta^k(s,a)$, because the sweep is the recursion that defines $\Delta^k$; so \cref{lem:agg_identity}\ref{aggid:returns} holds for $\bar Y_h^k$. The proofs of this appendix use the recorded returns only through these two properties: the range, including nonnegativity, in \cref{eq:agg_pathwise} and \cref{lem:agg_ix} and in the Freedman steps, and the conditional mean in \cref{lem:agg_local} through \cref{lem:relative_sn}. Both versions are functions of $\F_{k-1}$-measurable quantities, of $\tau^k$ and, for the randomized one, of the coins of episode $k$, as the coupling of \cref{app:coupling} requires; for the derandomized one the filtration generated by the trajectories suffices. The targets $\Gamma^k$, $\Delta^k$ and $g^k$ and the drift bounds of \cref{lem:agg_drift} are defined through $p$, $\pi^k$ and $\omega^k$ and do not depend on which version is recorded; the coins then serve only as a device in the proof of \cref{lem:agg_drift}, as they do in \cref{app:drift}. Hence \cref{thm:agg_main} holds for the derandomized variant with the same parameters. We state the randomized version because the identity between the surcharge and a randomly stopped rollout is what the analysis uses.
\end{remark}

\subsection{Surcharge Decomposition}\label{sec:agg_det_prelim}

Recall that $\pre^k(s):=\E_{p,\pi^k}\bigl[\sum_{m<h(s)}c^k(s_m,a_m)\bigm| s_{h(s)}=s\bigr]$ is the expected cost collected before reaching $s$ (and $\pre^k(s):=0$ if $q^k(s)=0$), so that $\E[C^k\mid\Fagg_{k-1},s_h^k=s,a_h^k=a]=\pre^k(s)+Q^k(s,a)$ whenever $q^k(s,a)>0$. Define $\Gamma^k$ backward from $\Gamma^k\equiv 0$ on $\Sset_H\times\Aset$ by
\begin{equation}\label{eq:agg_Gamma}
  \Gamma^k(s,a) := \sum_{s'\in\Sset_{h+1}} p(s'\mid s,a)\sum_{a'\in\Aset}\pi^k(a'\mid s')\bigl[\omega^k(s',a')\Rb_{h+1} + (1-\omega^k(s',a'))\Gamma^k(s',a')\bigr]\qquad (s\in\Sset_h,\ h<H),
\end{equation}
which is the expectation of $\Rb_{\sigma_h^k}$ given a visit to $(s,a)$, and set, for $s\in\Sset_h$,
\begin{equation}\label{eq:agg_targets}
\begin{aligned}
  \bar C^k(s,a) &:= \pre^k(s) + Q^k(s,a),\qquad
  \Delta^k(s,a) := \Rb_{h} - \Gamma^k(s,a),\qquad
  \widetilde r^k := \omega^k\Delta^k,\\
  \widetilde v^k &:= \bar C^k + \Delta^k,\qquad
  \widetilde z^k := \widetilde v^k - \widetilde r^k,\qquad
  \widetilde\Phi^k(s) := \ip{\pi^k(\cdot\mid s)}{\widetilde z^k(s,\cdot)} - \ip{\pi^\star(\cdot\mid s)}{\widetilde v^k(s,\cdot)},
\end{aligned}
\end{equation}
together with $g^k(s,a) := \bar C^k(s,a) + f_j(s)\Delta^k(s,a)$ on active states. Here $\Delta^k$ is the expected surcharge, $\widetilde v^k$ and $\widetilde z^k$ play the roles of $v^k$ and $z^k$ in \cref{eq:vz}, $\widetilde r^k$ is the induced bonus, and $g^k$ is the conditional mean of the recorded return.

\begin{lemma}[Surcharge decomposition]\label{lem:agg_identity}
Assume $\kagg\le 1/4$ and that every $\pi^k$ has full support. Then:
\begin{enumerate}[label=(\alph*),leftmargin=*]
\item\label{aggid:decomp} For every episode, and hence in total,
  \begin{equation}\label{eq:agg_decomp}
    \Reg = \sum_{s\in\Sset} q^\star(s)\sum_{k=1}^K \widetilde\Phi^k(s) + \sum_{k=1}^K\sum_{s,a} q^k(s,a)\,\widetilde r^k(s,a).
  \end{equation}
\item\label{aggid:ranges} Everywhere, $0\le\bar C^k\le H$, $2H\le\Delta^k\le\Rb_{h(s)}$, $0\le\widetilde z^k\le\widetilde v^k\le B$, and $\Delta^k\ge\frac23\widetilde v^k$. If $s$ is inactive, then $\widetilde\Phi^k(s)\le -H < 0$.
\item\label{aggid:returns} If $s\in\Sset_h$ is active in period $j\ge 2$, then $Y_h^k\in[0,B]$ and, almost surely,
  \begin{equation}\label{eq:agg_Y_condmean}
    \E\bigl[\ind_k[s,a]\,Y_h^k \mid \Fagg_{k-1}\bigr] = q^k(s)\,\pi^k(a\mid s)\,g^k(s,a).
  \end{equation}
\end{enumerate}
\end{lemma}

\begin{proof}
\emph{Part \ref{aggid:decomp}.} Work conditional on $\Fagg_{k-1}$ and omit superscript $k$. Let $\check z(s,a) := Q^\pi(s,a;c-\widetilde r)$ be the action-value function of $\pi$ under immediate costs $c-\widetilde r$, and set $\check v := \check z + \widetilde r$. Writing $\mathcal R(s,a)$ for the expected sum under $(p,\pi)$ of $\sum_{h'=h(s)+1}^H \widetilde r(s_{h'},a_{h'})$ given $(s_{h(s)},a_{h(s)})=(s,a)$, we have $\check z = Q - \widetilde r - \mathcal R$. Backward induction on $h=H,H-1,\dots,1$ shows $\mathcal R = \Gamma$. Both vanish on $\Sset_H$, and for $s\in\Sset_h$ ($h<H$) on $\Sset_{h+1}$ we have $\widetilde r+\mathcal R = \omega(\Rb_{h+1}-\Gamma)+\Gamma = \omega\Rb_{h+1}+(1-\omega)\Gamma$, which matches the bracket in \cref{eq:agg_Gamma}. Hence $\check v = Q - \Gamma$. Applying \cref{eq:pdl} to $c-\widetilde r$ gives $V^\pi(s_1;c-\widetilde r)-V^{\pi^\star}(s_1;c-\widetilde r)=\sum_{s\in\Sset}q^\star(s)\ip{\pi(\cdot\mid s)-\pi^\star(\cdot\mid s)}{\check z(s,\cdot)}$. Adding $V^\pi(s_1;c)-V^\pi(s_1;c-\widetilde r)=\sum_{s\in\Sset}q^\pi(s)\ip{\pi(\cdot\mid s)}{\widetilde r(s,\cdot)}$ and subtracting $V^{\pi^\star}(s_1;c)-V^{\pi^\star}(s_1;c-\widetilde r)=\sum_{s\in\Sset}q^\star(s)\ip{\pi^\star(\cdot\mid s)}{\widetilde r(s,\cdot)}$ gives
\[
  V^\pi(s_1;c) - V^{\pi^\star}(s_1;c)
  = \sum_{s\in\Sset}q^\star(s)\bigl(\ip{\pi(\cdot\mid s)}{\check z(s,\cdot)} - \ip{\pi^\star(\cdot\mid s)}{\check v(s,\cdot)}\bigr)
  + \sum_{s\in\Sset}q^\pi(s)\ip{\pi(\cdot\mid s)}{\widetilde r(s,\cdot)}.
\]
Adding the action-independent offset $\pre(s)+\Rb_{h(s)}$ to both $\check z(s,\cdot)$ and $\check v(s,\cdot)$ leaves $\ip{\pi}{\check z}-\ip{\pi^\star}{\check v}$ unchanged and turns $(\check v,\check z)$ into $(\widetilde v,\widetilde z)$.

\emph{Part \ref{aggid:ranges}.} For $s\in\Sset_h$, $\pre^k(s)\in[0,h-1]$ and $Q^k(s,a)\in[0,H-h+1]$, so $\bar C^k(s,a)\in[0,H]$. By \ref{aggQ:margin}, $\omega^k(s,a)\in[0,1]$. Backward induction on \cref{eq:agg_Gamma} with $\Rb_{h+2}\le\Rb_{h+1}$ gives $\Gamma^k(s,a)\in[0,\Rb_{h+1}]$, so $\Delta^k(s,a)\in[\Rb_h-\Rb_{h+1},\Rb_h]=[2H,\Rb_h]$. Thus $\widetilde v^k = \bar C^k+\Delta^k \le H+\Rb_h \le B$, $0\le\widetilde z^k\le\widetilde v^k$, $\bar C^k\le H\le\frac12\Delta^k$, $\widetilde v^k\le \frac32\Delta^k$, and $\widetilde\Phi^k(s)\le (H+\Rb_h)-2H < B$. If $s\in\Sset_h$ is inactive, $\omega^k(s,\cdot)\equiv 1$, so $\widetilde z^k(s,\cdot)=\bar C^k(s,\cdot)$ and $\widetilde\Phi^k(s) = \ip{\pi^k-\pi^\star}{Q^k(s,\cdot)} - \ip{\pi^\star}{\Delta^k(s,\cdot)} \le (H-h+1) - 2H = -(h-1)-H \le -H < 0$.

\emph{Part \ref{aggid:returns}.} Since $\sigma_h^k\in\{h+1,\dots,H+1\}$ and $\Rb$ is nonincreasing, we have $\Rb_h-\Rb_{\sigma_h^k}\in[2H,\Rb_h]$. Together with $C^k\in[0,H]$ and $f_j(s_h^k)\in[\frac34,1)$, this gives $Y_h^k\in[2f_j(s_h^k)H,\,B]\subseteq[0,B]$ pathwise. Now work on the $\Fagg_{k-1}$-measurable event that $s\in\Sset_h$ is active in period $j$. If $q^k(s)=0$, then $\ind\{s_h^k=s,a_h^k=a\}=0$ almost surely and both sides of \cref{eq:agg_Y_condmean} vanish. If $q^k(s)>0$, then $\Prob(s_h^k=s,a_h^k=a\mid\Fagg_{k-1})=q^k(s)\pi^k(a\mid s)>0$. Conditioned on $\Fagg_{k-1}$ and $\{s_h^k=s,a_h^k=a\}$, the pre-visit costs have mean $\pre^k(s)$ independently of $a$ and the suffix costs have mean $Q^k(s,a)$. Backward induction on the post-$h$ coin rollout gives $\E[\Rb_{\sigma_h^k}\mid\Fagg_{k-1},s_h^k=s,a_h^k=a]=\Gamma^k(s,a)$, which establishes \cref{eq:agg_Y_condmean}.
\end{proof}

\begin{lemma}[Self-adjusting active relative margins]\label{lem:agg_margins}
Assume $\kagg\le 1/4$, and let $s\in\Sset$ be active in the period $j\ge 2$ containing episode $k$. Write $\epsilon=\epsilon_j(s)$, $f=f_j(s)=1-\epsilon$, $\gamma=\gamma_j(s)$, $\pi=\pi^k(\cdot\mid s)$, and $\chi(a)=\pi(a)/(\pi(a)+\gamma)$, and set $m^{(k)} := \sum_{a\in\Aset}\pi(a)\chi(a)g^k(s,a)$. Everywhere on $\Omagg$, we have $m^{(k)}\in[\frac98 H,B]$ and
\begin{equation}\label{eq:agg_lrn_margin}
  \ip{\pi}{\widetilde z^k(s,\cdot)} \le \Bigl(1 - \frac14\epsilon\Bigr)m^{(k)}.
\end{equation}
In addition, every $\bar\pi\in\Delta(\Aset)$ satisfies
\begin{equation}\label{eq:agg_cmp_margin}
  \ip{\bar\pi}{g^k(s,\cdot)} \le \Bigl(1 - \frac23\epsilon\Bigr)\ip{\bar\pi}{\widetilde v^k(s,\cdot)}.
\end{equation}
\end{lemma}

\begin{proof}
Since $s$ is active, $1-\omega^k(s,a)=f^2\chi(a)$, so $\widetilde z^k(s,a)=\bar C^k(s,a)+f^2\chi(a)\Delta^k(s,a)$, while $\chi(a)g^k(s,a)=\chi(a)\bar C^k(s,a)+f\chi(a)\Delta^k(s,a)$. Subtracting and using $f^2-f = -f\epsilon$ gives
\[
  \ip{\pi}{\widetilde z^k(s,\cdot)} - m^{(k)}
  = \sum_{a\in\Aset}\pi(a)\bigl(1-\chi(a)\bigr)\bar C^k(s,a) - f\epsilon\sum_{a\in\Aset}\pi(a)\chi(a)\Delta^k(s,a).
\]
By \cref{lem:agg_identity}\ref{aggid:ranges}, $\bar C^k(s,a)\le H$, so $\sum_a\pi(a)(1-\chi(a))\bar C^k(s,a) \le HA\gamma$. Since $g^k = \bar C^k + f\Delta^k \le \frac32\Delta^k$, we have $m^{(k)}\le \frac32\sum_a\pi(a)\chi(a)\Delta^k(s,a)$. In addition, $\Delta^k\ge 2H$, $f\ge\frac34$, and $\sum_a\pi(a)\chi(a)\ge 1-A\gamma\ge\frac34$ give $m^{(k)}\ge \frac98 H$. Combining $H\le\frac89 m^{(k)}$, $A\gamma\le\frac14\epsilon$, and $f\ge\frac34$ gives $\ip{\pi}{\widetilde z^k(s,\cdot)} - m^{(k)} \le (\frac89 A\gamma - \frac23 f\epsilon)m^{(k)} \le -\frac{5}{18}\epsilon m^{(k)} \le -\frac14\epsilon m^{(k)}$. Finally, $g^k = \widetilde v^k - \epsilon\Delta^k \le (1-\frac23\epsilon)\widetilde v^k$ since $\Delta^k\ge\frac23\widetilde v^k$, which proves \cref{eq:agg_cmp_margin}.
\end{proof}

\paragraph{Pathwise bounds and drift under aggregate feedback.}
Assume $\kagg\le 1/4$. For any complete active block $b$ of $s\in\Sset_h$ in period $j\ge 2$, let $N_b(a):=\sum_{k\in b}\ind\{s_h^k=s,a_h^k=a\}$. Since $0\le Y_h^k \le B$, we have
\begin{equation}\label{eq:agg_pathwise}
  0 \le \pi_b(a)\widehat\ell_b(a) \le \Lambda_b B\,\frac{N_b(a)}{N_b},\qquad
  0 \le \widehat L_b := \ip{\pi_b}{\widehat\ell_b} \le \Lambda_b B,\qquad
  \sum_{a\in\Aset}\pi_b(a)^2\widehat\ell_b(a)^2 \le \Lambda_b B\,\widehat L_b.
\end{equation}
By \ref{aggQ:stab}, $\eta\widehat L_b\le \tau_b = \eta B\Lambda_b < 1/32$, so \cref{lem:barrier} bounds each block update's log-probability shift by $6\eta B\Lambda_b$ pathwise. Dyadic nesting (\cref{lem:dyadic}) then bounds log-policy changes across $\Sset\times\Aset$ by $6\eta B\Lambda$ within any complete active block of length $\Lambda$, and by $12\eta B\Lmax$ across two consecutive periods.

\begin{lemma}[Block and two-period drift under aggregate feedback]\label{lem:agg_drift}
Assume $\kagg\le 1/4$ and recall $\etail=e^{-20J}$.
\begin{enumerate}[label=(\alph*),leftmargin=*]
\item\label{aggdrift:block} Suppose $\eta \ge 1/(BK)$, fix $s\in\Sset_h$ and any deterministic interval $b=\{t_0,\dots,t_0+\Lambda-1\}\subseteq\{\Lmax+1,\dots,K\}$ in period $j\ge 2$ of length $\Lambda\in[M,\Lmax]$, and set $\tilde\varrho_b := 161\eta B D\Lambda < 0.011$. There exist $\Fagg_{t_0-1}$-measurable reference targets $g^{\circ,k}(s,a)\in[0,B]$ ($k\in b$, $a\in\Aset$) such that whenever $b$ is a complete active block of $s$ with $\bar q := q^{t_0}(s)\ge M/(3\Lambda)$, every $k\in b$ and $a\in\Aset$ satisfy
\begin{equation}\label{eq:agg_blockdrift}
  e^{-\tilde\varrho_b}\bar q \le q^k(s) \le e^{\tilde\varrho_b}\bar q,\qquad
  e^{-\tilde\varrho_b} g^{\circ,k}(s,a) - \etail B \le g^k(s,a) \le e^{\tilde\varrho_b} g^{\circ,k}(s,a) + \etail B.
\end{equation}
The target bounds extend to $\sum_{a\in\Aset}\alpha_a g^k(s,a)$ for weights $\alpha_a\ge 0$ with $\sum_a\alpha_a\le 1$.
\item\label{aggdrift:two} For any $s\in\Sset$, period $j\ge 2$ where $t$ begins period $j-1$, and episode $k$ in period $j-1$ or $j$,
\begin{equation}\label{eq:agg_twoperiod}
  e^{-\rho_P} q^t(s) - \etail \le q^k(s) \le e^{\rho_P} q^t(s) + \etail,\qquad \rho_P := 76\eta B D\Lmax < 0.02.
\end{equation}
\end{enumerate}
\end{lemma}

\begin{proof}
Fix $k\in b$ and set $\tau:=\eta B\Lambda$, $\bar\tau:=\eta B D\Lambda\le\frac{1}{16000}$, $\vartheta:=12\tau<1$, and $y:=22J$. In \cref{lem:pathlaw}, we have $\rho = \frac{576\bar\tau^2}{J}+12\sqrt{44}\,\bar\tau < 80\bar\tau =: \rho_b^0$ when $D=\sqrt{HJ}\le H$, and the zero-tail radius satisfies $H\vartheta=12\bar\tau<\rho_b^0$ when $D=H<\sqrt{HJ}$. Applying \cref{lem:pathlaw} to $F=\ind\{s_h=s\}$ gives $e^{-\rho_b^0}\bar q - e^{-22J}\le q^k(s)\le e^{\rho_b^0}\bar q + e^{-22J}$. Since $\bar q\ge M/(3K)$ and $\eta\ge 1/(BK)$, we have $4e^{-22J}/\bar q < \bar\tau$, which gives $e^{-\varrho_b'}\bar q\le q^k(s)\le e^{\varrho_b'}\bar q$ with $\varrho_b':=81\bar\tau$ (so $\rho_b^0+\varrho_b'=\tilde\varrho_b$).

Split $g^k(s,a)=\pre^k(s)+\mathrm{fw}^k(s,a)$ with $\mathrm{fw}^k(s,a):=Q^k(s,a)+f_j(s)(\Rb_h-\Gamma^k(s,a))$, and define $\pre^{\circ,k}(s)$ and $\mathrm{fw}^{\circ,k}(s,a)$ under the frozen policy $\pi^{t_0}$. Applying \cref{lem:pathlaw} to $F_{\mathrm{pre}}=\ind\{s_h=s\}\sum_{m=1}^{h-1}c^k(s_m,a_m)\in[0,H]$ gives $e^{-\rho_b^0}\bar q\,\pre^{\circ,k}(s)-H e^{-22J} \le q^k(s)\pre^k(s) \le e^{\rho_b^0}\bar q\,\pre^{\circ,k}(s)+H e^{-22J}$. Dividing by $q^k(s)\in[e^{-\varrho_b'}\bar q,e^{\varrho_b'}\bar q]$ and applying \ref{aggQ:tiny} gives $e^{-\tilde\varrho_b}\pre^{\circ,k}(s)-\frac12\etail B \le \pre^k(s)\le e^{\tilde\varrho_b}\pre^{\circ,k}(s)+\frac12\etail B$. For the forward part, consider the post-$h$ rollout that alternates transitions with action-coin pairs $(a_{h'},\varsigma_{h'})$. Because $t_0$ and $k$ belong to the same period $j$, every downstream state shares the same active status and $(f_j,\gamma_j)$ parameters at $t_0$ and $k$. The augmented action-coin log-sensitivity bound of \cref{app:drift} therefore keeps per-step log-ratios bounded by $2\cdot 6\tau = \vartheta$. Applying \cref{lem:pathlaw} to $F_{\mathrm{fw}} := \sum_{h'=h}^H c^k(s_{h'},a_{h'}) + f_j(s)(\Rb_h-\Rb_{\sigma_h})\in[0,B]$ gives $e^{-\rho_b^0}\mathrm{fw}^{\circ,k}(s,a)-\frac12\etail B \le \mathrm{fw}^k(s,a)\le e^{\rho_b^0}\mathrm{fw}^{\circ,k}(s,a)+\frac12\etail B$. Adding the prefix and forward bounds proves \ref{aggdrift:block} with $g^{\circ,k}:=\pre^{\circ,k}+\mathrm{fw}^{\circ,k}$, and \ref{aggdrift:two} follows identically with $\vartheta_P=12\eta B\Lmax$ and $y=20J$.
\end{proof}

\subsection{Concentration and Aggregate Regret Proof}\label{sec:agg_proof_finish}

Assume $\kagg\le 1/4$ and fix $s\in\Sset_h$ with $\bar\pi^\star:=(1-\frac1K)\pi^\star(\cdot\mid s)+\frac1K\Unif(\Aset)$. Let $n_s<K$ be the number of complete active blocks of $s$ in periods $j\ge 2$. Let $T_i$ be the episode preceding the $i$th complete active block ($T_i:=K$ for $i>n_s$), and set $\Gagg_i:=\Fagg_{T_i}$. For $i\le n_s$ ($b_i=\{T_i+1,\dots,T_i+\Lambda_i\}$), define
\[
\begin{gathered}
  V_i(a) := \sum_{k\in b_i}\widetilde v^k(s,a),\qquad
  \ell_i^\lrn := \sum_{k\in b_i}\ip{\pi_i}{\widetilde z^k(s,\cdot)},\qquad
  G_i(a) := \sum_{k\in b_i}g^k(s,a),\\
  m_i := \sum_{a\in\Aset}\pi_i(a)\chi_i(a)G_i(a),\qquad
  m_i^\star := \ip{\bar\pi^\star}{V_i},\qquad
  \widehat G_i^{\SN}(a) := \frac{\Lambda_i}{N_i}\frac{\Sigma_i(a)}{\pi_i(a)},\\
  W_i := \ell_i^\lrn-(1-2\tau_i)\widehat L_i,\qquad
  e_i := \Lambda_i B(3\etail+e^{-M/3}).
\end{gathered}
\]
Here $\widehat G_i^{\SN}\equiv 0$ when $N_i=0$, and for $i>n_s$ we set $\Lambda_i=0$ and all block sums and estimators to $0$. Block quantities are indexed as in \cref{app:events} but take their values at scale $B$: $\pi_i,N_i,\Sigma_i,\widehat\ell_i$ and $\widehat L_i=\ip{\pi_i}{\widehat\ell_i}$ are those of block $b_i$, and
\[
  \epsilon_i:=\bar a\Lambda_i,\qquad f_i:=1-\epsilon_i,\qquad\gamma_i:=\gamma_0\Lambda_i,\qquad\chi_i(a):=\frac{\pi_i(a)}{\pi_i(a)+\gamma_i},\qquad\tau_i:=\eta B\Lambda_i,\qquad\beta_i:=805\,\eta BD\Lambda_i ,
\]
all equal to $0$ (and $f_i=1$) for $i>n_s$. We write $\E_i[\cdot]:=\E[\cdot\mid\Gagg_i]$.

\begin{lemma}[Block-scaled implicit exploration at scale $B$]\label{lem:agg_ix}
Fix $s\in\Sset_h$, a deterministic $\bar\pi\in\Delta(\Aset)$, and $\{0,1\}$-valued $\Gagg_i$-measurable random variables $I_i$ ($i\in[K]$). Then
\[
  \Prob\Bigl(\sum_{i=1}^K I_i\bigl(\ip{\bar\pi}{\widehat\ell_i} - \E[\ip{\bar\pi}{\widehat G_i^{\SN}}\mid\Gagg_i]\bigr) > \frac{BJ}{2\gamma_0}\Bigr) \le e^{-J}.
\]
\end{lemma}
\begin{proof}
Repeating the proof of \cref{lem:ix} with $y_k:=Y_h^k/B\in[0,1]$ and without the outer factor $f_i$ gives $\exp(\frac{2\gamma_0}{B}\ip{\bar\pi}{\widehat\ell_i})\le 1+\frac{2\gamma_0}{B}\ip{\bar\pi}{\widehat G_i^{\SN}}$. Thus $\E[\exp(\frac{2\gamma_0}{B}I_i(\ip{\bar\pi}{\widehat\ell_i}-\E[\ip{\bar\pi}{\widehat G_i^{\SN}}\mid\Gagg_i]))\mid\Gagg_i]\le 1$, and Markov's inequality at $e^J$ concludes the proof.
\end{proof}

Repeating the proof of \cref{lem:classification} with \cref{eq:agg_twoperiod} shows that the classification event $\Eclagg$ across all $s\in\Sset$ and periods $j\ge 2$ fails with probability at most $2SKe^{-J}$. For each $s\in\Sset$ and $i\in[K]$, define the $\Gagg_i$-measurable indicator $I_i:=\ind\{T_i<K,\ q^{T_i+1}(s)\Lambda_i\ge M/3\}$. Since $\ell_i^\lrn,(1-2\tau_i)\widehat L_i,m_i^\star\in[0,\Lambda_i B]$ and $2\tau_i<1$ by \ref{aggQ:stab}, \cref{lem:freedman} applies at $\lambda=\eta$ to both $W_i-\E[W_i\mid\Gagg_i]$ and $\E[m_i^\star\mid\Gagg_i]-m_i^\star$. Intersecting $\Eclagg$ with these two Freedman events and the event of \cref{lem:agg_ix} (with $\bar\pi=\bar\pi^\star$ and the indicators $I_i$ just defined) across all $s\in\Sset$ defines a good event $\Eagg$. A union bound gives $1-\Prob(\Eagg)\le 2SKe^{-J}+3Se^{-J}\le 5SKe^{-J}<\delta$.

\begin{lemma}[Local regret under aggregate feedback]\label{lem:agg_local}
Assume $\kagg\le 1/4$. On $\Eagg$, every $s\in\Sset$ satisfies
\begin{equation}\label{eq:agg_local_bound}
  \sum_{k=1}^K \widetilde\Phi^k(s) \le \frac{A\log K + 2J}{\eta} + \frac{BJ}{2\gamma_0} + B\Lmax + 2B.
\end{equation}
\end{lemma}

\begin{proof}
Fix $s\in\Sset_h$ and work on $\Eagg$. If $\eta<1/(BK)$, then $\widetilde\Phi^k(s)<B$ (\cref{lem:agg_identity}\ref{aggid:ranges}) and $2J>1$ give $\sum_{k=1}^K\widetilde\Phi^k(s)\le BK < (A\log K+2J)/\eta$ deterministically. Assume now that $\eta\ge 1/(BK)$. The argument follows Steps~1--5 of \cref{app:local}, with three differences: the estimator \eqref{eq:agg_estimator} has no outer factor $f_b$, the learner's block loss $\ell_i^\lrn$ is not a multiple of $m_i$, and the two margins come from \cref{lem:agg_margins}. \cref{tab:agg_budget} lists the error terms.

\emph{Bias bounds.} Run the per-state uniform coupling of \cref{app:coupling} on an extension of $\Omagg$, with the coin vectors included in the history of each episode, and apply \cref{lem:relative_sn} with $H$ replaced by $B$ (\cref{rem:sn_range}) (setting $Y_h^k:=0$ when $s_h^k$ is inactive, so that the returns are defined on the whole space). On $\{I_i=1\}$, hypothesis \ref{H:size} holds by definition of $I_i$, and \ref{H:occ}, \ref{H:target} hold by \cref{lem:agg_drift}\ref{aggdrift:block} and \cref{lem:agg_identity}\ref{aggid:returns} with $\varrho=\tilde\varrho_b$. The choice $R_k=\ind\{a_h^k=a\}Y_h^k/\pi_i(a)$ gives $\widehat G_i^{\SN}(a)$ with target $G_i(a)$, and the choice $R_k=\chi_i(a_h^k)Y_h^k$ gives $\widehat L_i$ with target $m_i$. Since $\beta_i\ge e^{4\tilde\varrho_b}-1$ by \ref{aggQ:radius}, on $\{I_i=1\}$
\begin{equation}\label{eq:agg_bias}
  \bigl|\E_i[\widehat G_i^{\SN}(a)-G_i(a)]\bigr|\le\beta_i\,\E_i[G_i(a)]+e_i,\qquad
  \bigl|\E_i[\widehat L_i-m_i]\bigr|\le\beta_i\,\E_i[m_i]+e_i .
\end{equation}
On $\Eclagg\supseteq\Eagg$ we have $I_i=\ind\{i\le n_s\}$, and for $i>n_s$ all quantities vanish, so \eqref{eq:agg_bias} holds for every $i$. As in \cref{app:events}, $\Eclagg$ is used only to certify that the sample point lies in the $\Gagg_i$-measurable event $\{I_i=1\}$.

\emph{Step 1 (episodes outside complete active blocks).} If $s$ is inactive in episode $k$, then $\widetilde\Phi^k(s)\le -H\le 0$ by \cref{lem:agg_identity}\ref{aggid:ranges}. The at most one incomplete active block of $s$ contributes less than $B\Lmax$.

\emph{Step 2 (OMD ledger).} On $b_i$, $\sum_{k\in b_i}\widetilde\Phi^k(s)=\ell_i^\lrn-\ip{\pi^\star(\cdot\mid s)}{V_i}\le\ell_i^\lrn-m_i^\star+\Lambda_iB/K$, because $0\le V_i\le\Lambda_iB$; the last terms sum to at most $B$ over all blocks. Since $\eta\widehat L_i\le\tau_i<1/32$ by \ref{aggQ:stab}, \cref{lem:barrier} with comparator $\bar\pi^\star$ and \cref{eq:agg_pathwise} give $\sum_i(1-2\tau_i)\widehat L_i\le\sum_i\ip{\bar\pi^\star}{\widehat\ell_i}+\frac{A\log K}{\eta}$. Hence
\[
  \sum_{i\le n_s}\sum_{k\in b_i}\widetilde\Phi^k(s)\le\frac{A\log K}{\eta}+\sum_{i=1}^K W_i
  +\sum_{i=1}^K\bigl(\ip{\bar\pi^\star}{\widehat\ell_i}-m_i^\star\bigr)+B .
\]

\emph{Step 3 (learner side).} By \eqref{eq:agg_bias}, $(1-\beta_i)\E_i[m_i]-e_i\le\E_i[\widehat L_i]\le(1+\beta_i)\E_i[m_i]+e_i$. With $\ell_i^\lrn\le(1-\tfrac14\epsilon_i)m_i$ from \cref{eq:agg_lrn_margin},
\[
  \E_i[W_i]\le\bigl((1-\tfrac14\epsilon_i)-(1-2\tau_i)(1-\beta_i)\bigr)\E_i[m_i]+e_i\le\bigl(-\tfrac14\epsilon_i+\beta_i+2\tau_i\bigr)\E_i[m_i]+e_i .
\]
Since $\ell_i^\lrn$ and $(1-2\tau_i)\widehat L_i$ lie in $[0,\Lambda_iB]$, $W_i^2\le\Lambda_iB\bigl(\ell_i^\lrn+\widehat L_i\bigr)$ and hence $\E_i[W_i^2]\le\Lambda_iB\bigl(3\,\E_i[m_i]+e_i\bigr)$. On the Freedman event for $W_i-\E_i[W_i]$ (\cref{lem:freedman} with $\lambda=\eta$, applicable because $\eta|W_i-\E_i[W_i]|\le 2\tau_i<1$) and using $\eta\Lambda_iB=\tau_i\le 1$,
\[
  \sum_i W_i\le\frac J\eta+\sum_i\bigl(-\tfrac14\epsilon_i+\beta_i+5\tau_i\bigr)\E_i[m_i]+2\sum_ie_i
  \le\frac J\eta+2\sum_ie_i ,
\]
where the coefficient is negative by \ref{aggQ:slack}.

\emph{Step 4 (comparator side).} Decompose
\[
  \ip{\bar\pi^\star}{\widehat\ell_i}-m_i^\star
  =\bigl(\ip{\bar\pi^\star}{\widehat\ell_i}-\E_i\ip{\bar\pi^\star}{\widehat G_i^{\SN}}\bigr)
  +\bigl(\E_i\ip{\bar\pi^\star}{\widehat G_i^{\SN}}-\E_i[m_i^\star]\bigr)+\bigl(\E_i[m_i^\star]-m_i^\star\bigr).
\]
The first terms sum to at most $BJ/(2\gamma_0)$ on the event of \cref{lem:agg_ix}. For the second, \eqref{eq:agg_bias} averaged under $\bar\pi^\star$ and \cref{eq:agg_cmp_margin}, which gives $\ip{\bar\pi^\star}{G_i}\le(1-\tfrac23\epsilon_i)m_i^\star$, yield the bound $(1+\beta_i)(1-\tfrac23\epsilon_i)\E_i[m_i^\star]+e_i-\E_i[m_i^\star]\le(-\tfrac23\epsilon_i+\beta_i)\E_i[m_i^\star]+e_i$. Since $m_i^\star\in[0,\Lambda_iB]$, $\Var(m_i^\star\mid\Gagg_i)\le\Lambda_iB\,\E_i[m_i^\star]$, and on the Freedman event for $\E_i[m_i^\star]-m_i^\star$ the third terms sum to at most $\sum_i\tau_i\E_i[m_i^\star]+J/\eta$. As $-\tfrac23\epsilon_i+\beta_i+\tau_i<0$ by \ref{aggQ:slack},
\[
  \sum_i\bigl(\ip{\bar\pi^\star}{\widehat\ell_i}-m_i^\star\bigr)\le\frac{BJ}{2\gamma_0}+\frac J\eta+\sum_ie_i .
\]

\emph{Step 5 (remainders).} The blocks of $s$ are disjoint, so $\sum_i\Lambda_i\le K$ and $3\sum_ie_i\le3KB(3\etail+e^{-M/3})\le12K\etail B\le B$ by \ref{aggQ:tiny}. Combining Steps~1--5 proves \cref{eq:agg_local_bound}.
\end{proof}

\begin{table}[ht]
\caption{Error budget for the local regret of a state under aggregate feedback, on block $i$. All multiplicative terms are paid from the learner margin $\frac14\epsilon_i\E_i[m_i]$ or the comparator margin $\frac23\epsilon_i\E_i[m_i^\star]$ of \cref{lem:agg_margins}.}
\label{tab:agg_budget}
\begin{center}
\begin{footnotesize}
\setlength{\tabcolsep}{3.5pt}
\begin{tabular}{llll}
\toprule
Error term & Source & Bound & How it is paid\\
\midrule
implicit-exploration shrinkage of $\bar C^k$ & \cref{lem:agg_margins} & $HA\gamma_i$ per episode & inside the learner margin\\
self-normalization bias (learner) & \cref{eq:agg_bias} & $\beta_i\E_i[m_i]$ ($+e_i$) & learner margin ($+e_i$: Step 5)\\
log-barrier OMD stability & \cref{lem:barrier} & $2\tau_i\E_i[m_i]$ & learner margin\\
martingale variance (learner) & \cref{lem:freedman} & $3\tau_i\E_i[m_i]$ ($+e_i$) & learner margin ($+e_i$: Step 5)\\
self-normalization bias (comparator) & \cref{eq:agg_bias} & $\beta_i\E_i[m_i^\star]$ ($+e_i$) & comparator margin ($+e_i$: Step 5)\\
martingale variance (comparator) & \cref{lem:freedman} & $\tau_i\E_i[m_i^\star]$ & comparator margin\\
Freedman tails ($J/\eta$ each) & \cref{lem:freedman} & $2J/\eta$ total & additive, once per state\\
implicit-exploration tail & \cref{lem:agg_ix} & $BJ/(2\gamma_0)$ total & additive, once per state\\
log-barrier OMD penalty & \cref{lem:barrier} & $A\log K/\eta$ total & additive, once per state\\
comparator smoothing & Step 2 & $\Lambda_iB/K$ & additive, $\le B$ total\\
exponentially small remainders $e_i$ & Step 5 & $\Lambda_iB(3\etail+e^{-M/3})$ & additive, $\le B$ total\\
final incomplete active block & Step 1 & $B\Lmax$ & additive, once per state\\
\bottomrule
\end{tabular}
\end{footnotesize}
\end{center}
\end{table}

\begin{proposition}[Aggregate master inequality]\label{prop:agg_master}
Let $\eta,\gamma_0>0$ satisfy $\kagg = a_0 + 5A\gamma_0 \le 1/4$ with $a_0=\tilde C_0\eta B D$. On $\Eagg$,
\begin{equation}\label{eq:agg_master}
  \Reg \le \frac{\taeta}{\eta} + \tbeeta\,\eta + \frac{\tagam}{\gamma_0} + \tbgam\,\gamma_0 + 2HB.
\end{equation}
\end{proposition}
\begin{proof}
Summing \cref{eq:agg_local_bound} against $\sum_{s\in\Sset}q^\star(s)=H$ in \cref{eq:agg_decomp} gives $H(\frac{A\log K+2J}{\eta}+\frac{BJ}{2\gamma_0}+B\Lmax+2B)$. Consider next the optimism sum $\sum_{k=1}^K\sum_{s\in\Sset}q^k(s)\ip{\pi^k(\cdot\mid s)}{\widetilde r^k(s,\cdot)}$. In period~$1$, $\omega^k\equiv 1$ gives $\Gamma^k=\Rb_{h+1}$ and $\widetilde r^k\equiv 2H$, which contributes at most $2H^2\Lmax$. In periods $j\ge 2$, active states satisfy $\ip{\pi^k(\cdot\mid s)}{\omega^k(s,\cdot)}\le 2\epsilon_j(s)+A\gamma_j(s)\le 2\kagg\Lambda_j(s)$ with $q^k(s)\Lambda_j(s)\le 5M$ on $\Eclagg$, while inactive states satisfy $q^k(s)\le 3M/\Lmax < 24M\kagg$. Since $\Delta^k(s,a)\le\Rb_h\le 2H^2$, every state $s\in\Sset_h$ contributes at most $24M\kagg\Rb_h\le 48 M H^2\kagg = \tilde C_1 M H^2\kagg$ per episode on $\Eclagg$. Summing over $s\in\Sset$ and $K$ episodes bounds the period-$j\ge 2$ optimism sum by $\tilde C_1 M S H^2 K\kagg = \tbeeta\,\eta + \tbgam\,\gamma_0$. Finally, $(HB+2H^2)\Lmax < \frac{H+1}{\eta}$ by \ref{aggQ:stab} and $H(A\log K+2J)+H+1 < 2HAJ=\taeta$, which proves \cref{eq:agg_master}.
\end{proof}

\begin{proof}[Proof of \cref{thm:agg_main}]
Choose the tuned rates $\eta=\sqrt{\taeta/\tbeeta}$ and $\gamma_0=\sqrt{\tagam/\tbgam}$ from \cref{tab:agg_params} (which have the orders stated in \cref{sec:aggregate}), and set $\widetilde{\mathcal B}:=2\sqrt{\taeta\tbeeta}+2\sqrt{\tagam\tbgam}+2HB$. If $\kagg\le 1/4$, then $\Prob(\Eagg)\ge 1-\delta$ and \cref{prop:agg_master} implies $\Reg\le\widetilde{\mathcal B}$ on $\Eagg$. If instead $\kagg>1/4$, the learner plays $\Unif(\Aset)$ throughout, and $\widetilde{\mathcal B}\ge\tbeeta\eta+\tbgam\gamma_0 = \tilde C_1 M S H^2 K\kagg > 12 M S H^2 K \ge HK\ge\Reg$ deterministically since $S\ge H\ge 1$. Substituting $B=2H^2+H\le 3H^2$ and $M=O(J)$ gives $\sqrt{\taeta\tbeeta}=O(H^{5/2}\sqrt{D}\,J\sqrt{SAK})$, $\sqrt{\tagam\tbgam}=O(H^{5/2}J\sqrt{SAK})$, and $2HB\le 6H^{5/2}\sqrt{S}=O(\sqrt{\taeta\tbeeta})$. Lazy initialization and min-heap block scheduling achieve $O(HA)$ amortized time per episode and $O(SA)$ space as in \cref{app:extensions}.
\end{proof}

\subsection{Two Limits of the Construction}\label{sec:agg_barriers}

The two propositions of this subsection explain design choices of \aggalg{}. They are statements about estimators and surcharge schedules of the kind used here and are not lower bounds on the regret of other algorithms.

\begin{proposition}[Aggregate feedback is invariant to cost transfers between layers]\label{prop:agg_transfer}
Fix $k\in[K]$ and past costs $c^1,\dots,c^{k-1}$, and let $\mathfrak h_{k-1}$ denote the pre-episode-$k$ history generating $\Fagg_{k-1}$. Let $\Psi = \Psi(\mathfrak h_{k-1},\tau^k,C^k,U_{\mathrm{int}}^k)$ be a fixed integrable measurable map, where the conditional law of the internal randomness $U_{\mathrm{int}}^k$ given $(\Fagg_{k-1},\tau^k,C^k)$ does not depend on $c^k$. For any $(s,\tilde a)\in\Sset_h\times\Aset$ with $\Prob(s_h^k=s,a_h^k=\tilde a\mid\Fagg_{k-1})>0$, set $\mu(c^k;\tilde a) := \E[\Psi\mid\Fagg_{k-1},s_h^k=s,a_h^k=\tilde a]$. Then $\mu(c^k;\tilde a)$ is invariant under any step transfer that adds $\delta_{\mathrm{tr}}\in\R$ on $\Sset_{h_1}\times\Aset$ and subtracts $\delta_{\mathrm{tr}}$ on $\Sset_{h_2}\times\Aset$ ($h_1\ne h_2$) while keeping $c^k\in[0,1]^{\Sset\times\Aset}$. Consider the multiplicative-discount targets
\[
  v^{\mathrm{mult},k}(s,a) := c^k(s,a) + \E_{\pi^k}\!\left[\sum_{h'=h+1}^H\Bigl(\prod_{m=h+1}^{h'} w^k(s_m,a_m)\Bigr)c^k(s_{h'},a_{h'})\,\middle|\, s_h=s,a_h=a\right].
\]
For $h'>h$, set $W_{h,h'}^k(s,\tilde a) := \E_{\pi^k}[\prod_{m=h+1}^{h'}w^k(s_m,a_m)\mid s_h=s,a_h=\tilde a]$. Suppose $a\ne a'$ satisfy $\Prob(s_h^k=s,a_h^k=\tilde a\mid\Fagg_{k-1})>0$ for both $\tilde a\in\{a,a'\}$ and $W_{h,h'}^k(s,a)\ne W_{h,h'}^k(s,a')$ for some $h'>h$. Then no such $\Psi$ matches $v^{\mathrm{mult},k}(s,\cdot)$ up to an action-independent offset across $c^k\in[\frac14,\frac34]^{\Sset\times\Aset}$.
\end{proposition}
\begin{proof}
Every trajectory $\tau^k$ visits each layer once, and its conditional law given $\Fagg_{k-1}$ does not depend on $c^k$. Any step transfer leaves $C^k(\tau^k)$ unchanged on every path, so $(\mathfrak h_{k-1},\tau^k,C^k,U_{\mathrm{int}}^k)$ has the same conditional law given $(\Fagg_{k-1},s_h^k=s,a_h^k=\tilde a)$ and $\mu(c^k;\tilde a)$ is unchanged for each $\tilde a\in\{a,a'\}$. Starting from $c^k\equiv\frac12$, add $\delta_{\mathrm{tr}}=\frac14$ on $\Sset_{h'}\times\Aset$ and subtract $\frac14$ on $\Sset_h\times\Aset$, which keeps costs in $[\frac14,\frac34]^{\Sset\times\Aset}$. The layer-$h$ shift cancels in $v^{\mathrm{mult},k}(s,a)-v^{\mathrm{mult},k}(s,a')$, whereas layer $h'$ shifts this difference by $\frac14(W_{h,h'}^k(s,a)-W_{h,h'}^k(s,a'))\ne 0$. Because $\mu(\cdot;a)-\mu(\cdot;a')$ stays constant, $\mu(c^k;\cdot)$ cannot match $v^{\mathrm{mult},k}(s,\cdot)$ up to an action-independent offset on both cost vectors.
\end{proof}

\cref{prop:agg_transfer} excludes single-episode statistics whose conditional mean equals the multiplicatively discounted target exactly, up to an action-independent offset; this is why \aggalg{} keeps $C^k$ at unit weight and discounts only an additive term that it controls.

\begin{proposition}[Surcharge scale needed for nonpositive local regret at inactive states]\label{prop:agg_barrier}
Consider any deterministic first-death surcharge schedule $\Rb_1,\dots,\Rb_H\ge 0$ with $\Rb_{H+1}=0$ in \cref{eq:agg_Gamma,eq:agg_targets}, where inactive states set $\omega^k(s,\cdot)\equiv 1$ and $\widetilde\Phi^k(s)$ is evaluated against $\pi^\star\in\argmin_\pi\sum_{k'=1}^K V^\pi(s_1;c^{k'})$. Suppose every inactive state $s$ satisfies $\widetilde\Phi^k(s)\le 0$ for every episode $k\in[K]$ across all layered MDPs with $|\Sset_m|\le 2$, oblivious cost sequences $(c^1,\dots,c^K)$ in $[0,1]^{\Sset\times\Aset}$, full-support policies $\pi^k$, and active/inactive assignments. Then $\Rb_h - \Rb_{h+1} \ge H-h+1$ for every $h\in[H]$, and hence $\Rb_1\ge \frac12 H(H+1)$ and $\sum_{h=1}^H\Rb_h \ge \frac16 H(H+1)(H+2)$.
\end{proposition}
\begin{proof}
Fix $h\in[H]$ and construct a layered MDP with $\Sset_1=\{s_{1,+}\}$ and $\Sset_m=\{s_{m,+},s_{m,-}\}$ for $2\le m\le H$. Take identical costs $c^1=\dots=c^K=c$. For every $2\le m\le H$ and $a\in\Aset$, state $s_{m,-}$ incurs cost $1$, with $p(s_{m+1,-}\mid s_{m,-},a)=1$ when $m<H$. For every $m\in[H]\setminus\{h\}$ and $a\in\Aset$, state $s_{m,+}$ incurs cost $0$, with $p(s_{m+1,+}\mid s_{m,+},a)=1$ when $m<H$. At $s_{h,+}$, one action $a^\star\in\Aset$ incurs cost $0$ (and sets $p(s_{h+1,+}\mid s_{h,+},a^\star)=1$ if $h<H$), whereas every $a\ne a^\star$ incurs cost $1$ (and sets $p(s_{h+1,-}\mid s_{h,+},a)=1$ if $h<H$). Every policy $\pi$ then has $V^\pi(s_{1,+};c)=(1-\pi(a^\star\mid s_{h,+}))(H-h+1)$, so every optimal comparator $\pi^\star$ satisfies $\pi^\star(a^\star\mid s_{h,+})=1$. Mark $s_{h,+}$ (and $s_{h+1,+}$ if $h<H$) inactive, which gives $\Gamma^k(s_{h,+},a^\star)=\Rb_{h+1}$ and $\widetilde\Phi^k(s_{h,+}) = (1-\pi^k(a^\star\mid s_{h,+}))(H-h+1) - (\Rb_h-\Rb_{h+1})$. Letting $\pi^k(a^\star\mid s_{h,+})\downarrow 0$ under $\widetilde\Phi^k(s_{h,+})\le 0$ forces $\Rb_h-\Rb_{h+1}\ge H-h+1$, and summing over layers gives the bounds on $\Rb_1$ and $\sum_{h=1}^H\Rb_h$.
\end{proof}

\cref{prop:agg_barrier} gives a necessary condition for the property $\widetilde\Phi^k(s)\le 0$ at inactive states, which is the sufficient condition that our proof uses, within deterministic layer-indexed surcharge schedules. The schedule of \aggalg{} uses the uniform increment $2H$ per layer, larger than the necessary $H-h+1$, because \cref{lem:agg_margins} needs $\Delta^k\ge 2H\ge 2\bar C^k$: the term $\bar C^k$ contains the prefix cost and can be as large as $H$ at every layer.

%% file: appendix/M_linear_block.tex
\section{Analysis for Linear AMDPs}\label{app:linear_block}

This appendix defines linear AMDPs (\cref{def:linear}), specifies \linblkalg{} in full (\cref{alg:linear_block}), states and proves the exchangeability identity (\cref{lem:exch_matrix_sn}), and proves \cref{thm:linear_block} and \cref{prop:lin_agg_barrier}. Periods are indexed by $r$ and shells by $j$.

\subsection{The Algorithm}\label{sec:lin_alg_full}

\begin{definition}[Linear AMDP]\label{def:linear}
An AMDP is \emph{linear} with dimension $d$ if there is a known feature map $\phi:\Sset\times\Aset\to\R^d$ with $\|\phi(s,a)\|_2\le 1$ such that, for every layer $h$ and all $(s,a)\in\Sset_h\times\Aset$, $p(s'\mid s,a)=\ip{\phi(s,a)}{\psi_h(s')}$ and $c^k(s,a)=\ip{\phi(s,a)}{\zeta_h^k}\in[0,1]$, for unknown $\psi_h:\Sset_{h+1}\to\R^d$ and $\zeta_h^k\in\R^d$ with $\bigl\|\sum_{s'\in\Sset_{h+1}}|\psi_h(s')|\bigr\|_2\le\sqrt d$ (absolute values taken coordinatewise) and $\|\zeta_h^k\|_2\le\sqrt d$.
\end{definition}

\paragraph{Parameters.}
Set $\iota := 4\log(64KdHA/\delta) > 22$, $\lambda := (KdH)^{-20}$, $\rho := 2d\log(1+1/\lambda)\ge 2$, $W_\star := d\sqrt{(H^{1/2}+\sqrt{d/H})K}$, $\Lmax := 2^{\lfloor\log_2\lceil W_\star\rceil\rfloor}$, $J_0 := \lceil\log_2(2\Lmax)\rceil$, and $d_{\mathrm{VC}} := \lceil 2(d+1)(d+2)(J_0+1)\log_2(J_0+2)\rceil$. With the constants $C_{\mathrm{lin}} := 400$, $C_\eta := 2\times 10^4$ and $c_{\mathrm{blk}} := 4\times 10^6$, set
\begin{equation}\label{eq:lin_blk_params}
\begin{aligned}
  m_0 &:= \lceil C_{\mathrm{lin}}\rho\iota\rceil,\qquad
  \eta := \frac{1}{C_\eta\sqrt{\rho}\,H^{3/2}\sqrt{\iota}\,\Lmax},\qquad
  \beta_{\mathrm{blk}} := c_{\mathrm{blk}}\,\eta\rho m_0^2 H^{5/2}\sqrt{\iota},\\
  \theta &:= \max\Bigl\{4\beta_{\mathrm{blk}},\ \frac{C_{\mathrm{lin}}}{\Lmax}\bigl(\rho(d_{\mathrm{VC}}\iota+\iota^2)+2m_0\bigr)\Bigr\},
\end{aligned}
\end{equation}
so that $m_0=\tTheta(d)$, $\beta_{\mathrm{blk}}=\tTheta(d^{5/2}H/\Lmax)$ and $\theta=\tTheta((d^{5/2}H+d^3)/\Lmax)$. Write $\etail^{\mathrm{lin}} := e^{-4\iota} < (64KdHA)^{-16}$.

\paragraph{Shells.}
At the start of period $r\ge 2$ and for each layer $h$, \cref{alg:regions} (the region construction of \citealt{kulkarni2026introducing}, with window $W=\Lmax$) is applied to the features $\{\phi_h^k\}_k$ observed at layer $h$ during period $r-1$. It returns $J_{r,h}\le J_0$ disjoint shells $\cC_{r,h,j}\subset\R^d$ with empirical masses $\phat_{r,h,j}>\theta/2$ and truncated empirical covariances $\Sigmahat_{r,h,j}$ such that $\|\phi\|_{\Sigmahat_{r,h,j}^\dagger}^2\le 2\rho$ for all $\phi\in\cC_{r,h,j}$ (\cref{lem:lin_peeling_det}). We write $\Pi_{r,h,j}:=\Sigmahat_{r,h,j}\Sigmahat_{r,h,j}^\dagger$ for the projection onto the range of $\Sigmahat_{r,h,j}$ and $\Sigmahat^{-1/2}:=(\Sigmahat^\dagger)^{1/2}$. Shell $(r,h,j)$ has the dyadic block length $\Lambda_{r,h,j}:=\min\{2^i: i\ge 0,\ 2^i\phat_{r,h,j}\ge m_0\}$. The indicator $u_r(s,a):=\ind\{\phi(s,a)\notin\bigcup_j\cC_{r,h(s),j}\}$ marks pairs outside all shells, and pairs with $\phi(s,a)\in\cC_{r,h(s),j}$ receive the bonus $b_r(s,a):=\beta_{\mathrm{blk}}/\phat_{r,h(s),j}\in[0,\frac12]$. In the first period, $u_1\equiv 1$ and $b_1\equiv 0$.

\paragraph{Shifted costs and returns.}
To cut the returns outside the shells and subtract the bonus on the shells without creating negative returns, we add $1$ to the cost of every non-terminal step: $\tilde c^k(s,a) := c^k(s,a) + \ind\{h(s)\le H-1\}$. This adds $H-1$ to $V^\pi(s_1;c^k)$ for every policy $\pi$ and leaves $\Reg$ unchanged. By \cref{def:linear}, for $(s,a)\in\Sset_h\times\Aset$ with $h<H$,
\begin{equation}\label{eq:lin_unit}
  1 = \sum_{s'\in\Sset_{h+1}} p(s'\mid s,a) = \Bigl\langle\phi(s,a),\sum_{s'\in\Sset_{h+1}}\psi_h(s')\Bigr\rangle,
  \qquad \Bigl\|\sum_{s'\in\Sset_{h+1}}\psi_h(s')\Bigr\|_2 \le \sqrt{d},
\end{equation}
so $\tilde c^k(s,a) = \ip{\phi(s,a)}{\tilde\zeta_h^k}$ is still linear, with $\|\tilde\zeta_h^k\|_2\le 2\sqrt{d}$. In period $r$, the returns are computed backward from $Z_{H+1}^{\mathrm{ret},k}:=0$ by
\[
  Y_h^k := \tilde c^k(s_h^k,a_h^k) + Z_{h+1}^{\mathrm{ret},k},\qquad
  Z_h^{\mathrm{ret},k} := \bigl(1 - u_r(s_h^k,a_h^k)\bigr)\bigl(Y_h^k - b_r(s_h^k,a_h^k)\bigr),
\]
which keeps $Z_h^{\mathrm{ret},k}\ge 0$ for $h\le H-1$ and $0\le Y_h^k\le 2H$ on every sample path (\cref{lem:lin_decomp}).

\paragraph{Block estimates and policies.}
For a block $b$ of shell $(r,h,j)$ with length $\Lambda=\Lambda_{r,h,j}$, let $X_k:=\ind\{\phi_h^k\in\cC_{r,h,j}\}$, $u_{k,h}:=\Sigmahat_{r,h,j}^{-1/2}\phi_h^k$, $G_b:=\sum_{k\in b}X_ku_{k,h}u_{k,h}^\top$, $y_b:=\sum_{k\in b}X_ku_{k,h}Y_h^k$ and $N_b:=\sum_{k\in b}X_k$. The block is accepted if $\acc_b:=\ind\{G_b\succeq\frac{m_0}{16}\Pi_{r,h,j},\ N_b\le 24m_0\}$ equals one, and its estimate is the least-squares vector
\begin{equation}\label{eq:lin_ls}
  \what_b:=\acc_b\,\Lambda\,G_b^\dagger y_b=\acc_b\,\Lambda\,G_b^\dagger\sum_{k\in b}X_k\,u_{k,h}\,Y_h^k ,
\end{equation}
in which the Gram matrix $G_b$ of the block takes the place of the visit count of \cref{eq:estimator}. For $\phi(s,a)\in\cC_{r,h,j}$ the block loss is
\[
  \Qhat_b(s,a) := \bigl(\Sigmahat_{r,h,j}^{-1/2}\phi(s,a)\bigr)^\top\what_b - \frac{\Lambda}{2}\,b_r(s,a) - \Lambda K^{-3},
\]
and the loss passed to the local learner at the last episode $k$ of the block is $g_k(s,a):=\Qhat_b(s,a)-\eta\Qhat_b(s,a)^2$, with the second-order correction of \citet{bartlett2008high} and \citet{zimmert2022return} that \citet{kulkarni2026introducing} also use. For pairs on a shell, $g_k(s,a):=0$ in all other episodes of the period. For pairs outside all shells, $g_k(s,a):=-K^{-3}-\eta K^{-6}$ in every episode of a period $r\ge 2$. In the first period, $g_k\equiv 0$. The policy is the exponential-weights distribution $\pi^k(a\mid s)\propto\exp(-\eta\sum_{t<k}g_t(s,a))$, which is evaluated only at the visited states. This requires storing, for every past period $r'$ and layer $h$, the shell data $(M_{r',h,j}^{-1},\Sigmahat_{r',h,j}^{-1/2},\phat_{r',h,j},\Lambda_{r',h,j})_j$ and the vectors $\what_b$ of all blocks, which takes memory polynomial in $d,H,K$. Throughout we assume that $\Sset$ is a standard Borel space and that all functions of $(s,a)$ below are measurable, so that regular conditional distributions and maximal couplings exist; sums over $\Sset_h$ are read as integrals. Ties in \cref{alg:regions} are broken by smallest index.

\begin{algorithm}[ht]
\caption{\linblkalg{} for linear AMDPs with bandit feedback}
\label{alg:linear_block}
\begin{small}
\begin{algorithmic}[1]
\STATE \textbf{Input:} $\Lmax$, $\lambda$, $\rho$, $m_0$, $\eta$, $\theta$, $\beta_{\mathrm{blk}}$ as above. If $\Lmax < 200\iota$, $\Lmax > K/4$, or $\theta > 1/2$, play $\Unif(\Aset)$ in all episodes and \textbf{return}.
\FOR{period $r=1,2,\dots,\lceil K/\Lmax\rceil$}
  \IF{$r\ge 2$}
    \STATE For each $h\in[H]$, run \cref{alg:regions} on $\{\phi_h^k\}_{k\in\text{period }r-1}$ to obtain $(\cC_{r,h,j},\phat_{r,h,j},\Sigmahat_{r,h,j})_{j=1}^{J_{r,h}}$; set $\Lambda_{r,h,j}$, $u_r$, $b_r$ as above and $G_{r,h,j},y_{r,h,j},N_{r,h,j}\leftarrow 0$.
  \ENDIF
  \FOR{episode $k=(r-1)\Lmax+1,\dots,\min\{r\Lmax,K\}$}
    \STATE For $h=1,\dots,H$: compute $\pi^k(\cdot\mid s_h^k)\propto\exp(-\eta\sum_{t<k} g_{t}(s_h^k,\cdot))$, draw $a_h^k$, observe $c^k(s_h^k,a_h^k)$ and $s_{h+1}^k$; let $\phi_h^k:=\phi(s_h^k,a_h^k)$.
    \IF{$r\ge 2$}
      \STATE Compute $Y_H^k,\dots,Y_1^k$. For each $h$ with $\phi_h^k\in\cC_{r,h,j}$: $u_{k,h}\leftarrow\Sigmahat_{r,h,j}^{-1/2}\phi_h^k$, $G_{r,h,j}\leftarrow G_{r,h,j}+u_{k,h} u_{k,h}^\top$, $y_{r,h,j}\leftarrow y_{r,h,j}+u_{k,h} Y_h^k$, $N_{r,h,j}\leftarrow N_{r,h,j}+1$.
      \FORALL{shells $(h,j)$ with $k\bmod\Lambda_{r,h,j}=0$}
        \STATE $(G_b,y_b,N_b)\leftarrow(G_{r,h,j},y_{r,h,j},N_{r,h,j})$; store $\what_b\leftarrow \acc_b\,\Lambda_{r,h,j}\,G_b^\dagger y_b$, which defines $g_k$ on $\cC_{r,h,j}$; reset $G_{r,h,j},y_{r,h,j},N_{r,h,j}\leftarrow 0$.
      \ENDFOR
    \ENDIF
  \ENDFOR
\ENDFOR
\end{algorithmic}
\end{small}
\end{algorithm}

\paragraph{Boundary parameter regimes.}
Note that $\frac12 W_\star \le \Lmax \le 2W_\star$. If $\Lmax < 200\iota$, then $W_\star < 400\iota$, so $d H^{1/4}\sqrt{K} < 400\iota$ and $\Reg \le HK < 400\iota\,H^{3/4}\sqrt{K}$. If $\Lmax > K/4$, then $\sqrt{K} < 8 d(H^{1/4}+d^{1/4}H^{-1/4})$, so $\Reg \le HK = O\bigl((d H^{5/4}+d^{5/4}H^{3/4})\sqrt{K}\bigr)$. If $\theta > 1/2$, then $\theta\Lmax = \tO(d^{5/2}H + d^3)$ and $\Lmax\ge\frac12 W_\star$ give $\Reg \le HK < 2(\theta\Lmax)\frac{HK}{\Lmax} = \tO\bigl((d^{3/2}H^{7/4}+d^{7/4}H^{5/4})\sqrt{K}\bigr)$. Thus \cref{thm:linear_block} holds deterministically whenever \cref{alg:linear_block} returns at Line~1. Assume $200\iota\le\Lmax\le K/4$ and $\theta\le 1/2$ henceforth. The constants in \eqref{eq:lin_blk_params} were chosen for a transparent proof and are very large, so this regime is reached only for extremely large $K$: for instance, with $d=H=5$, $A=2$ and $\delta=0.1$, \cref{alg:linear_block} leaves the uniform policy only for $K$ above about $2\cdot 10^{49}$. \cref{thm:linear_block} is a statement about the dependence of the regret on $d$, $H$ and $K$.

\subsection{Asymmetric R-Max Decomposition and Peeling Properties}\label{sec:lin_decomp_bounds}

For any episode $k$ with $\F_{k-1}$-measurable indicators $u_{(k)}:\Sset\times\Aset\to\{0,1\}$ and active bonuses $b_{(k)}:\Sset\times\Aset\to[0,1/2]$ satisfying $b_{(k)}(s,a)=0$ whenever $u_{(k)}(s,a)=1$ (in \linblkalg{}, $(u_{(k)},b_{(k)})=(u_r,b_r)$ for $k$ in period $r$), define backward from $V_{k,H+1}^z\equiv 0$, for $h=H,\dots,1$ and $(s,a)\in\Sset_h\times\Aset$:
\begin{equation}\label{eq:lin_vz}
\begin{gathered}
  v_{k,h}(s,a) := \tilde c^k(s,a) + \ind\{h<H\}\sum_{s'\in\Sset_{h+1}} p(s'\mid s,a)\,V_{k,h+1}^z(s'),\\
  z_{k,h}(s,a) := \bigl(1 - u_{(k)}(s,a)\bigr)\bigl(v_{k,h}(s,a) - b_{(k)}(s,a)\bigr),\qquad
  V_{k,h}^z(s) := \sum_{a\in\Aset}\pi^k(a\mid s)\,z_{k,h}(s,a),
\end{gathered}
\end{equation}
and set $r_{k,h}(s,a) := v_{k,h}(s,a) - z_{k,h}(s,a)$.

\begin{lemma}[Asymmetric R-max decomposition and pathwise $O(H)$ returns]\label{lem:lin_decomp}
Suppose $0\le b_{(k)}(s,a)\le\frac12$ for all $(s,a)\in\Sset\times\Aset$, and suppose $\pi^k,u_{(k)},b_{(k)}$ are $\F_{k-1}$-measurable. Then:
\begin{enumerate}[label=(\alph*),leftmargin=*]
\item\label{linD:bounds} On every sample path, $0\le Y_h^k\le 2H$ for all $h\in[H]$, and $0\le v_{k,h}(s,a)\le 2H$ and $-\frac12\ind\{h=H\}\le z_{k,h}(s,a)\le v_{k,h}(s,a)$ for all $h\in[H]$ and $(s,a)\in\Sset_h\times\Aset$.
\item\label{linD:linear} For all $h\in[H]$ and $(s,a)\in\Sset_h\times\Aset$, $v_{k,h}(s,a) = \ip{\phi(s,a)}{\xi_{k,h}}$ for an $\F_{k-1}$-measurable vector $\xi_{k,h}\in\R^d$ with $\|\xi_{k,h}\|_2\le (2H+2)\sqrt{d}$, and $\E[Y_h^k\mid\F_{k-1},s_h^k,a_h^k] = v_{k,h}(s_h^k,a_h^k)$ almost surely.
\item\label{linD:regret} Single-episode regret satisfies
\[
\begin{aligned}
  V^k(s_1) - V^{\pi^\star}(s_1;c^k)
  &= \sum_{h=1}^H\sum_{s\in\Sset_h} q^\star(s)\bigl(\ip{\pi^k(\cdot\mid s)}{z_{k,h}(s,\cdot)} - \ip{\pi^\star(\cdot\mid s)}{v_{k,h}(s,\cdot)}\bigr)\\
  &\quad + \sum_{h=1}^H\sum_{s\in\Sset_h,a\in\Aset} q^k(s,a)\,r_{k,h}(s,a).
\end{aligned}
\]
\end{enumerate}
\end{lemma}

\begin{proof}
If $H=1$, then $Y_1^k = c^k(s_1^k,a_1^k)\in[0,1]$. If $H\ge 2$, at $h=H$ we have $Y_H^k\in[0,1]$ and $Z_H^{\mathrm{ret},k}\in[-\frac12,1]$. At $h=H-1$, $\tilde c^k\ge 1$ gives $Y_{H-1}^k \ge 1 - \frac12 = \frac12 \ge b_{(k)}(s_{H-1}^k,a_{H-1}^k)$, so $Z_{H-1}^{\mathrm{ret},k}\ge 0$. Backward induction yields $Z_h^{\mathrm{ret},k}\ge 0$ for all $h\le H-1$ and $0\le Y_h^k\le \sum_{h'=h}^H\tilde c^k(s_{h'}^k,a_{h'}^k)\le 2(H-h)+1\le 2H$ for all $h\in[H]$. The same backward induction on \cref{eq:lin_vz} gives $0\le v_{k,h}\le 2H$ and $|V_{k,h+1}^z(s')|\le 2H$.

For part~\ref{linD:linear}, backward induction under the Markov kernel gives $\E[Y_h^k\mid\F_{k-1},s_h^k,a_h^k]=v_{k,h}(s_h^k,a_h^k)$ almost surely. By \cref{def:linear} and \cref{eq:lin_unit}, $v_{k,h}(s,a)=\ip{\phi(s,a)}{\xi_{k,h}}$ across all $(s,a)\in\Sset_h\times\Aset$ with $\xi_{k,h}=\zeta_h^k+\ind\{h<H\}\bigl(\sum_{s'\in\Sset_{h+1}}\psi_h(s')+\sum_{s'\in\Sset_{h+1}}\psi_h(s')V_{k,h+1}^z(s')\bigr)$ and $\|\xi_{k,h}\|_2\le(2+2H)\sqrt{d}$. For part~\ref{linD:regret}, $V^\pi(s_1;c^k)-V^{\pi^\star}(s_1;c^k) = V^\pi(s_1;\tilde c^k)-V^{\pi^\star}(s_1;\tilde c^k)$, and $z_{k,h}$ is the exact $Q$-function of $\pi^k$ under $\tilde c^k - r_{k,h}$; applying \cref{eq:pdl} yields the identity.
\end{proof}

\begin{algorithm}[!t]
\caption{Ellipsoid Feature-Region Peeling \citep{kulkarni2026introducing}}
\label{alg:regions}
\begin{algorithmic}[1]
\STATE \textbf{Input:} Features $\phi_1,\dots,\phi_W\in\R^d$, $\lambda>0$, $\theta>0$, $\rho=2d\log(1+1/\lambda)$.
\STATE Initialize $R\leftarrow\{1,\dots,W\}$ and $j\leftarrow 0$.
\WHILE{$|R| > W\theta$}
  \STATE Set $j\leftarrow j+1$, $m_j\leftarrow |R|$, $R_j\leftarrow R$, and $M_j\leftarrow \lambda I + \frac{1}{m_j}\sum_{i\in R_j}\phi_i\phi_i^\top$.
  \WHILE{some $i\in R_j$ satisfies $\|\phi_i\|_{M_j^{-1}}^2 > \rho$}
    \STATE Remove one such $i$ from $R_j$ and update $M_j\leftarrow \lambda I + \frac{1}{m_j}\sum_{i\in R_j}\phi_i\phi_i^\top$ (keeping $m_j$ fixed).
  \ENDWHILE
  \STATE Set $\cC_j\leftarrow \{\phi\in\R^d : \|\phi\|_{M_j^{-1}}^2\le\rho\}\setminus(\cC_1\cup\dots\cup\cC_{j-1})$ and $\phat_j\leftarrow |R_j|/W$.
  \STATE Set $\Sigmahat_j\leftarrow \frac{1}{|R_j|}\sum_{i\in R_j}\phi_i\phi_i^\top$ with eigenvalues strictly below $\lambda m_j/|R_j|$ zeroed out, and $R\leftarrow R\setminus R_j$.
\ENDWHILE
\STATE \textbf{Return} $(\cC_i,\phat_i,\Sigmahat_i)_{i=1}^j$ (storing $M_i^{-1}$ for each $\cC_i$).
\end{algorithmic}
\end{algorithm}

\begin{lemma}[Deterministic properties of ellipsoid peeling]\label{lem:lin_peeling_det}
On any $W$ vectors $\phi_1,\dots,\phi_W\in\R^d$ with $\|\phi_i\|_2\le 1$, \cref{alg:regions} terminates in at most $J_0=\lceil\log_2(2W)\rceil$ shells. Each shell $j$ satisfies $R_j=\{i\in[W]:\phi_i\in\cC_j\}$, $\phat_j = |R_j|/W > \theta/2$, $\|\phi\|_{\Sigmahat_j^\dagger}^2\le 2\rho$, and $\|(I-\Sigmahat_j\Sigmahat_j^\dagger)\phi\|_2\le \sqrt{2\rho\lambda}$ for all $\phi\in\cC_j$.
\end{lemma}

\begin{proof}
Following \citet[Lemma~B.1]{kulkarni2026introducing}, as points are removed in the inner loop, $M_j$ decreases in Loewner order so $M_j^{-1}$ increases monotonically. Every previously removed point therefore keeps $\|\phi_i\|_{M_j^{-1}}^2>\rho$, and $R_j=\{i\in[W]:\phi_i\in\cC_j\}$. Removing $\phi_i$ with $\phi_i^\top M_j^{-1}\phi_i>\rho$ multiplies $\det(M_j)$ by $1-\phi_i^\top M_j^{-1}\phi_i/m_j < e^{-\rho/m_j}$. Since $\det(\lambda I + \frac{1}{m_j}\sum_{i\in R}\phi_i\phi_i^\top)/\det(\lambda I)\le (1+1/\lambda)^d = e^{\rho/2}$, fewer than $m_j/2$ points are removed, giving $|R_j|>m_j/2 > W\theta/2$. Consequently $m_{j+1}=m_j-|R_j|<m_j/2$ with $m_1=W$ and $m_j\ge 1$, so the number of shells is at most $J_0$. Because $M_j = \lambda I + \frac{|R_j|}{m_j}\bigl(\frac{1}{|R_j|}\sum_{i\in R_j}\phi_i\phi_i^\top\bigr)$ shares its eigenbasis with $\Sigmahat_j$, every retained eigenvalue $\mu\ge\lambda m_j/|R_j|$ satisfies $\lambda+\frac{|R_j|}{m_j}\mu\le 2\mu$ (so $\Sigmahat_j^\dagger\preceq 2M_j^{-1}$), while every zeroed eigenvalue $\mu<\lambda m_j/|R_j|$ satisfies $\lambda+\frac{|R_j|}{m_j}\mu < 2\lambda$ (so $I-\Sigmahat_j\Sigmahat_j^\dagger \preceq 2\lambda M_j^{-1}$). For every $\phi\in\cC_j$, this yields $\phi^\top\Sigmahat_j^\dagger\phi\le 2\phi^\top M_j^{-1}\phi\le 2\rho$ and $\|(I-\Sigmahat_j\Sigmahat_j^\dagger)\phi\|_2^2\le 2\lambda\phi^\top M_j^{-1}\phi\le 2\rho\lambda$.
\end{proof}

\subsection{\texorpdfstring{$G_b$}{G\_b}-Norm Pathwise Bounds and Period Covariance Concentration}\label{sec:lin_blk_drift_conc}

\begin{lemma}[Block pathwise bounds and dyadic log-policy drift]\label{lem:lin_blk_stab}
In \cref{alg:linear_block}, on every sample path in every period $r\ge 2$:
\begin{enumerate}[label=(\alph*),leftmargin=*]
\item\label{lblk:path} Every $(s,a)$ satisfies $0\le b_r(s,a)\le\frac12$, and every shell $(r,h,j)$ satisfies $1\le\Lambda_{r,h,j}\le\Lmax$, $\Lambda_{r,h,j}\mid\Lmax$, and $m_0\le\Lambda_{r,h,j}\phat_{r,h,j}<2m_0$. On every complete block $b$ of length $\Lambda=\Lambda_{r,h,j}$, setting $\widehat L_b(s,a):=(\Sigmahat_{r,h,j}^{-1/2}\phi(s,a))^\top\widehat w_b$ and $\Qtil_b := \Qhat_b - \eta\Qhat_b^2$, we have $|\widehat L_b(s,a)|\le 56\sqrt{\rho}\,H\Lambda$, $|\Qhat_b(s,a)|\le 57\sqrt{\rho}\,H\Lambda =: R_b$ with $\eta R_b < \frac18$, and $|\Qtil_b(s,a)|\le 65\sqrt{\rho}\,H\Lambda$.
\item\label{lblk:drift} Let $b=\{t_0,\dots,t_0+\Lambda-1\}$ be an aligned dyadic interval of a period $r\ge 2$, that is, $\Lambda$ is a power of two dividing $\Lmax$ and $t_0-1-(r-1)\Lmax$ is a multiple of $\Lambda$; this includes every complete block of a shell and the whole period ($\Lambda=\Lmax$). Then every $(s',a')\in\Sset\times\Aset$ and $k\in b$ satisfy $|\log\frac{\pi^k(a'\mid s')}{\pi^{t_0}(a'\mid s')}|\le \vartheta_b := 130\eta\sqrt{\rho}\,H\Lambda < \frac{1}{150\sqrt{H\iota}}$, with Hoeffding--Azuma path-law radius (\cref{lem:pathlaw} at $y=4\iota$) $\varrho_b^{\mathrm{lin}} := 4H\vartheta_b^2+\vartheta_b\sqrt{8H\iota} < 370\eta\sqrt{\rho}\,H^{3/2}\sqrt{\iota}\,\Lambda < 0.02$. Across any two consecutive periods $r-1$ and $r$, per-step log-ratios change by at most $2\vartheta_{\Lmax} < \frac{1}{75\sqrt{H\iota}}$ with path-law radius $\varrho_{2P} < 0.04$.
\end{enumerate}
\end{lemma}

\begin{proof}
By \cref{lem:lin_peeling_det} and \cref{eq:lin_blk_params}, $\phat_{r,h,j}>\theta/2\ge \max\{2\beta_{\mathrm{blk}}, m_0/\Lmax\}$, so $0\le b_r(s,a)\le\frac{2\beta_{\mathrm{blk}}}{\theta}\le\frac12$ and $1\le\Lambda_{r,h,j}\le\Lmax$ divides $\Lmax$ with $m_0\le\Lambda_{r,h,j}\phat_{r,h,j}<2m_0$. Fix $\phi(s,a)\in\cC_{r,h,j}$ and write $v:=\Sigmahat_{r,h,j}^{-1/2}\phi(s,a)\in\operatorname{range}(\Sigmahat_{r,h,j})$, so $\|v\|_2^2=\|\phi(s,a)\|_{\Sigmahat_{r,h,j}^\dagger}^2\le 2\rho$. When $\acc_b=0$, $\widehat L_b(s,a)=0$. When $\acc_b=1$, we have $0\preceq G_b^\dagger\preceq\frac{16}{m_0}\Pi_{r,h,j}$, $G_b^\dagger G_b G_b^\dagger = G_b^\dagger$, $N_b\le 24m_0$, and $0\le Y_h^k\le 2H$ (\cref{lem:lin_decomp}\ref{linD:bounds}). Applying Cauchy--Schwarz across the $N_b$ block visits $X_k=\ind\{\phi_h^k\in\cC_{r,h,j}\}$ gives
\[
  \bigl|v^\top G_b^\dagger y_b\bigr|
  \le \Bigl(\sum_{k\in b} X_k\bigl(v^\top G_b^\dagger u_{k,h}\bigr)^2\Bigr)^{1/2}\Bigl(\sum_{k\in b} X_k(Y_h^k)^2\Bigr)^{1/2}
  = \bigl(v^\top G_b^\dagger v\bigr)^{1/2}\Bigl(\sum_{k\in b} X_k(Y_h^k)^2\Bigr)^{1/2}.
\]
Since $v^\top G_b^\dagger v\le \frac{16}{m_0}\|v\|_2^2\le \frac{32\rho}{m_0}$ and $\sum_{k\in b}X_k(Y_h^k)^2\le 24m_0\cdot 4H^2 = 96m_0 H^2$, the factor $m_0$ cancels and $|v^\top G_b^\dagger y_b|\le \sqrt{3072\,\rho}\,H < 56\sqrt{\rho}\,H$. Multiplying by $\Lambda$ yields $|\widehat L_b(s,a)|\le 56\sqrt{\rho}\,H\Lambda$. Since $\sqrt{\rho}\,H\ge\sqrt{2}>1.25$, we get $|\Qhat_b(s,a)|\le 57\sqrt{\rho}\,H\Lambda = R_b$, $\eta R_b < 0.0013 < \frac18$, and $|\Qtil_b(s,a)|\le \frac98 R_b < 65\sqrt{\rho}\,H\Lambda$.

For part~\ref{lblk:drift}, fix $(s',a')\in\Sset_{h'}\times\Aset$ and $k\in b$. If $u_r(s',a')=1$, then $\sum_{t'=t_0}^{k-1}|g_{t'}(s',a')|\le 2\Lambda K^{-3}$. If $u_r(s',a')=0$, then $\phi(s',a')$ belongs to a unique shell $\cC_{r,h',j'}$, and every completed block $b'$ of $(r,h',j')$ ending in $\{t_0,\dots,k-1\}$ satisfies $b'\subsetneq b$ by dyadic nesting, so $\sum_{b'\subsetneq b}\Lambda_{b'}\le\Lambda$ and $\sum_{t'=t_0}^{k-1}|g_{t'}(s',a')|\le 65\sqrt{\rho}\,H\Lambda$. Thus $|\log\frac{\pi^k(a'\mid s')}{\pi^{t_0}(a'\mid s')}|\le 2\eta\max_{\tilde a}\sum_{t'=t_0}^{k-1}|g_{t'}(s',\tilde a)|\le 130\eta\sqrt{\rho}\,H\Lambda = \vartheta_b$. Spanning two consecutive periods bounds total block lengths by $2\Lmax$, giving $2\vartheta_{\Lmax}$, and substituting $\vartheta_b$ and $2\vartheta_{\Lmax}$ into \cref{lem:pathlaw} ($y=4\iota$) gives $\varrho_b^{\mathrm{lin}} < 370\eta\sqrt{\rho}\,H^{3/2}\sqrt{\iota}\,\Lambda < 0.02$ and $\varrho_{2P} < 0.04$.
\end{proof}

\paragraph{Uniform convergence for slowly drifting samples.}
The shells of period $r$ are computed from the trajectories of period $r-1$, which are generated by slowly changing, data-dependent policies and are therefore neither independent nor identically distributed. The next four lemmas show that uniform convergence over the relevant class of sets still holds relative to the policy at the start of period $r-1$. The plan follows \citet[Lemmas~B.4, B.5 and C.4]{kulkarni2026introducing}: a relative deviation bound for i.i.d.\ samples (\cref{lem:lin_vc_iid}), a bound on the VC dimension of the class of shells (\cref{lem:lin_vc_class}), and a coupling of the drifting sample with i.i.d.\ samples. The coupling is where we depart from them. Theirs assumes that the density ratio between the trajectory laws of two policies is bounded on every sample path, whereas \cref{lem:pathlaw} bounds it only outside an event of small probability. \cref{lem:lin_coupling} therefore clips the density ratio and pays for the clipping with a maximal coupling; it is stated for abstract sequential samples and can be read independently of the rest of this appendix. \cref{lem:lin_uc_drift} combines the three ingredients.

\begin{lemma}[Relative uniform convergence, i.i.d.\ case]\label{lem:lin_vc_iid}
Let $\mathcal T$ be a class of measurable subsets of a measurable space with VC dimension at most $V\ge 1$, let $X_1,\dots,X_n$ be i.i.d.\ with law $\mu$ and empirical measure $\widehat\mu$, let $\delta''\in(0,1)$, and set $\epsilon_n:=\frac{9}{2n}\bigl(V\log(2en)+\log(8/\delta'')\bigr)$. With probability at least $1-\delta''$, every $T\in\mathcal T$ satisfies
\[
  \tfrac23\bigl(\mu(T)-\epsilon_n\bigr)\le\widehat\mu(T)\le\tfrac32\,\mu(T)+\epsilon_n,
\]
and the same inequalities hold with $\mu(T),\widehat\mu(T)$ replaced by $\int f\,d\mu,\int f\,d\widehat\mu$ for every measurable $f$ with values in $[0,1]$ whose level sets $\{f>t\}$, $t\in[0,1)$, belong to $\mathcal T$.
\end{lemma}

\begin{proof}
Let $\epsilon':=(V\log(2en)+\log(8/\delta''))/n$. By the relative deviation inequalities of \citet[Theorem~5.1]{boucheron2005theory} and Sauer's lemma, with probability at least $1-\delta''$ every $T\in\mathcal T$ satisfies $\mu(T)-\widehat\mu(T)\le 2\sqrt{\mu(T)\epsilon'}$ and $\widehat\mu(T)-\mu(T)\le 2\sqrt{\widehat\mu(T)\epsilon'}$. Since $2\sqrt{x\epsilon'}=2\sqrt{(x/3)(3\epsilon')}\le x/3+3\epsilon'$, the first inequality gives $\widehat\mu(T)\ge\frac23\mu(T)-3\epsilon'=\frac23(\mu(T)-\epsilon_n)$ and the second gives $\widehat\mu(T)\le\frac32\mu(T)+\frac92\epsilon'=\frac32\mu(T)+\epsilon_n$. For $f$ as in the statement, integrate these inequalities over the level sets, using $\int f\,d\mu=\int_0^1\mu(\{f>t\})\,dt$ and the same identity for $\widehat\mu$.
\end{proof}

\begin{lemma}[The class of shells]\label{lem:lin_vc_class}
Let $\mathcal T$ be the class of all intersections of at most $J_0+1$ subsets of $\R^d$ of the form $\{\phi:\phi^\top Q\phi\le c\}$ or $\{\phi:\phi^\top Q\phi>c\}$, with $Q\in\R^{d\times d}$ symmetric and $c\in\R$. Then $\mathcal T$ has VC dimension at most $d_{\mathrm{VC}}$. For any output of \cref{alg:regions} with at most $J_0$ shells, $\mathcal T$ contains every shell $\cC_j$, the complement $\R^d\setminus\bigcup_{j}\cC_j$ of their union, and every set $\{\phi\in\cC_j:(x^\top\phi)^2>t\}$ with $x\in\R^d$ and $t\in\R$.
\end{lemma}

\begin{proof}
Each set $\{\phi:\phi^\top Q\phi\le c\}$ and its complement is a halfspace in the $d(d+1)/2$ coordinates $(\phi_i\phi_{i'})_{i\le i'}$, so these sets form a class of VC dimension at most $(d+1)(d+2)/2$. By \citet[Lemma~3.2.3]{blumer1989learnability}, intersections of at most $J_0+1$ of them have VC dimension at most $(d+1)(d+2)(J_0+1)\log_2(3(J_0+1))\le 2(d+1)(d+2)(J_0+1)\log_2(J_0+2)\le d_{\mathrm{VC}}$. For the second claim, let $M_j$ be the final matrix of the $j$th iteration of \cref{alg:regions}. Then $\cC_j=\{\phi^\top M_j^{-1}\phi\le\rho\}\cap\bigcap_{j'<j}\{\phi^\top M_{j'}^{-1}\phi>\rho\}$ is an intersection of $j\le J_0$ such sets, the complement of the union of the shells is $\bigcap_{j}\{\phi^\top M_j^{-1}\phi>\rho\}$, and $\{\phi\in\cC_j:(x^\top\phi)^2>t\}=\cC_j\cap\{\phi^\top(xx^\top)\phi>t\}$.
\end{proof}

\begin{lemma}[Coupling a slowly drifting sample with i.i.d.\ samples]\label{lem:lin_coupling}
Let $P_1$ be a probability measure on a standard Borel space, let $W\ge 1$, $\rho_0:=\frac1{50}$ and $\varepsilon\in[0,10^{-5}]$. Let $\tau^{(1)},\dots,\tau^{(W)}$ be random elements of this space such that, for every $i\in[W]$ and every value of $(\tau^{(1)},\dots,\tau^{(i-1)})$, the conditional law $P_i$ of $\tau^{(i)}$ given $(\tau^{(1)},\dots,\tau^{(i-1)})$ has a density $L_i$ with respect to $P_1$ with
\begin{equation}\label{eq:lin_clip_tails}
  P_i\bigl(L_i>e^{\rho_0}\bigr)\le\varepsilon,\qquad P_1\bigl(L_i<e^{-\rho_0}\bigr)\le\varepsilon .
\end{equation}
Then there is a probability space carrying $(\tau^{(i)})_{i\le W}$ with their joint law, i.i.d.\ elements $\tau_1,\dots,\tau_W,\tau_1',\dots,\tau_W'$ with law $P_1$, and i.i.d.\ Bernoulli variables $B_1,\dots,B_W$ with mean $e^{-\rho_0}$ that are independent of $(\tau_i)_{i\le W}$, together with an event $\mathcal C$ of probability at least $1-2W\varepsilon-2e^{-W/100}$ on which:
\begin{enumerate}[label=(\roman*),leftmargin=*,nosep]
\item\label{cpl:inj} there is an injective map from $\{1,\dots,W\}$ into the index set of the family $(\tau_1,\dots,\tau_W,\tau_1',\dots,\tau_W')$ that sends each $i$ to an element equal to $\tau^{(i)}$;
\item\label{cpl:eq} $\tau^{(i)}=\tau_i$ for every $i$ with $B_i=1$;
\item\label{cpl:count} more than $9W/10$ of the variables $B_1,\dots,B_W$ are equal to one.
\end{enumerate}
\end{lemma}

\begin{proof}
\emph{Step 1 (clipped mixture).} Let $\bar L_i:=\min\{\max\{L_i,e^{-\rho_0}\},e^{\rho_0}\}$ and $Z_i:=\E_{P_1}[\bar L_i]-e^{-\rho_0}$. By \eqref{eq:lin_clip_tails} and $\E_{P_1}[L_i]=1$,
\[
  \E_{P_1}|\bar L_i-L_i|\le\E_{P_1}\bigl[L_i\ind\{L_i>e^{\rho_0}\}\bigr]+e^{-\rho_0}P_1\bigl(L_i<e^{-\rho_0}\bigr)\le 2\varepsilon,\qquad \bigl|Z_i-(1-e^{-\rho_0})\bigr|\le 2\varepsilon.
\]
In particular $Z_i>0$, because $1-e^{-\rho_0}>0.0198$. Let $R_i$ be the probability measure with density $(\bar L_i-e^{-\rho_0})/Z_i$ with respect to $P_1$, and let $\widetilde P_i:=e^{-\rho_0}P_1+(1-e^{-\rho_0})R_i$. Then
\[
  \|\widetilde P_i-P_i\|_{\mathrm{TV}}=\tfrac12\,\E_{P_1}\Bigl|e^{-\rho_0}+\tfrac{1-e^{-\rho_0}}{Z_i}\bigl(\bar L_i-e^{-\rho_0}\bigr)-L_i\Bigr|\le\tfrac12\,\E_{P_1}|\bar L_i-L_i|+\tfrac12\bigl|1-e^{-\rho_0}-Z_i\bigr|\le 2\varepsilon.
\]

\emph{Step 2 (construction).} Let $\tau_1,\dots,\tau_W$ and $\tau_1',\tau_2',\dots$ be i.i.d.\ with law $P_1$ and let $B_1,\dots,B_W$ be i.i.d.\ Bernoulli variables with mean $e^{-\rho_0}$, all independent. We generate $\tau^{(1)},\dots,\tau^{(W)}$ in order. For index $i$, given everything generated before it, first form a proposal $\widetilde\tau^{(i)}$: if $B_i=1$, let $\widetilde\tau^{(i)}:=\tau_i$; if $B_i=0$, go through the elements $\tau_m'$ that have not been examined yet, in increasing order of $m$, accept each one independently with probability $(\bar L_i(\tau_m')-e^{-\rho_0})/(e^{\rho_0}-e^{-\rho_0})\in[0,1]$, and let $\widetilde\tau^{(i)}$ be the first accepted one. An element $\tau_m'$ that has not been examined is independent of everything generated before its examination, so by the correctness of rejection sampling the proposal has conditional law $R_i$ when $B_i=0$, and hence conditional law $\widetilde P_i$ overall. Then draw $\tau^{(i)}$ from a maximal coupling of $P_i$ and $\widetilde P_i$ given $\widetilde\tau^{(i)}$. Its conditional law given everything generated before index $i$ is $P_i$ (for $i=1$ this is $P_1$ itself; the densities $L_i$ and the maximal couplings are taken in jointly measurable versions, which exist because the space is standard Borel), so $(\tau^{(i)})_{i\le W}$ has the prescribed joint law, and $\Prob(\tau^{(i)}\ne\widetilde\tau^{(i)})\le 2\varepsilon$ for each $i$. After index $W$, examine the remaining $\tau_m'$ with $m\le W$ in the same way with $\bar L_W$, so that every $\tau_m'$ with $m\le W$ has an acceptance indicator.

\emph{Step 3 (the event $\mathcal C$).} Let $\mathcal S$ be the event that $\tau^{(i)}=\widetilde\tau^{(i)}$ for all $i\le W$; then $\Prob(\mathcal S^c)\le 2W\varepsilon$. Let $\mathcal E_1$ be the event that fewer than $W/10$ of the $B_i$ are zero; since $\Prob(B_i=0)=1-e^{-\rho_0}<0.025$, Hoeffding's inequality gives $\Prob(\mathcal E_1^c)\le e^{-2W(0.075)^2}\le e^{-W/100}$. Let $\mathcal E_2$ be the event that at least $W/10$ of $\tau_1',\dots,\tau_W'$ are accepted. Each of them is accepted with conditional probability $Z_i/(e^{\rho_0}-e^{-\rho_0})\ge(1-e^{-\rho_0}-2\varepsilon)/(e^{\rho_0}-e^{-\rho_0})>0.49$ given everything generated before its examination, so the Azuma--Hoeffding inequality gives $\Prob(\mathcal E_2^c)\le e^{-2W(0.39)^2}\le e^{-W/100}$. Let $\mathcal C:=\mathcal S\cap\mathcal E_1\cap\mathcal E_2$. On $\mathcal E_1\cap\mathcal E_2$, fewer than $W/10$ indices have $B_i=0$, and their proposals are the first accepted elements $\tau_m'$, all with $m\le W$ and with distinct indices $m$. Hence on $\mathcal C$ each $\tau^{(i)}$ equals $\tau_i$ if $B_i=1$ and equals an element $\tau_m'$ with $m\le W$ used by no other index if $B_i=0$, which gives \ref{cpl:inj} and \ref{cpl:eq}, and \ref{cpl:count} holds on $\mathcal E_1$.
\end{proof}

\begin{lemma}[Uniform convergence within a period]\label{lem:lin_uc_drift}
Fix a period $r\ge 2$ and a layer $h$, let $t_i:=t_{\mathrm{ref}}(r)+i-1$ for $i\in[\Lmax]$ be the episodes of period $r-1$, where $t_{\mathrm{ref}}(r):=(r-2)\Lmax+1$, and write $\mu:=\Prob_{\pi^{t_1}}(\phi_h\in\cdot)$ for the law of the layer-$h$ feature under the policy of the first episode of period $r-1$. Let $\bar\epsilon:=\frac{9}{\Lmax}\bigl(d_{\mathrm{VC}}\log(2e\Lmax)+\log(512KH/\delta)\bigr)$. With probability at least $1-\frac{\delta}{16KH}$, every $T$ in the class $\mathcal T$ of \cref{lem:lin_vc_class} satisfies
\begin{equation}\label{eq:lin_uc}
  \frac{\mu(T)-\bar\epsilon}{3}\;\le\;\frac{1}{\Lmax}\sum_{i=1}^{\Lmax}\ind\{\phi_h^{t_i}\in T\}\;\le\;3\,\mu(T)+\bar\epsilon,
\end{equation}
and the same holds with $\mu(T)$ and $\ind\{\phi_h^{t_i}\in T\}$ replaced by $\int f\,d\mu$ and $f(\phi_h^{t_i})$ for every $f:\R^d\to[0,1]$ whose level sets belong to $\mathcal T$.
\end{lemma}

\begin{proof}
Write $W:=\Lmax\ge 200\iota$ and condition on $\F_{t_1-1}$, which fixes $\pi^{t_1}$. Let $\tau^{(i)}$ be the trajectory of episode $t_i$ and let $P_i$ be the law of a trajectory under $\pi^{t_i}$, so that $P_i$ is the conditional law of $\tau^{(i)}$ given $(\tau^{(1)},\dots,\tau^{(i-1)})$ and $P_1$ is fixed. This uses that \linblkalg{} has no internal randomness and that, the costs being fixed in advance, the observed costs are functions of the trajectory, so that $\pi^{t_i}$ is a function of $\F_{t_1-1}$ and of the earlier trajectories of the period.

\emph{Step 1 (density ratios).} All policies have full support, so $P_i\ll P_1$ with density $L_i$ equal to the product of the $H$ ratios $\pi^{t_i}(a_{h'}\mid s_{h'})/\pi^{t_1}(a_{h'}\mid s_{h'})$ along the trajectory. By \cref{lem:lin_blk_stab}\ref{lblk:drift}, each ratio lies in $e^{\pm \vartheta_{\Lmax}}$ with $\vartheta_{\Lmax}<1/(150\sqrt{H\iota})$ on every sample path (the policies do not change at all during the first period). The proof of \cref{lem:pathlaw} with $y=4\iota$ shows that the log-likelihood ratio of two such laws exceeds $4H\vartheta_{\Lmax}^2+\vartheta_{\Lmax}\sqrt{8H\iota}<\rho_0=\frac{1}{50}$ with probability at most $\etail^{\mathrm{lin}}=e^{-4\iota}$ under the law in the numerator. Applied in both directions, this gives \eqref{eq:lin_clip_tails} with $\varepsilon=\etail^{\mathrm{lin}}<10^{-5}$, for every value of the past. \cref{lem:lin_coupling} therefore applies; let $(\tau_i)_{i\le W}$, $(\tau_m')_{m\le W}$, $(B_i)_{i\le W}$ and $\mathcal C$ be as in its statement, with $\Prob(\mathcal C^c)\le 2W\etail^{\mathrm{lin}}+2e^{-W/100}$.

\emph{Step 2 (uniform convergence).} Let $I$ consist of the first $\lceil W/2\rceil$ indices $i$ with $B_i=1$ (completed by indices with $B_i=0$ if there are fewer). The $B_i$ are independent of the $\tau_i$, so given $(B_i)_{i\le W}$ the elements $(\tau_i)_{i\in I}$ are i.i.d.\ with law $P_1$. Apply \cref{lem:lin_vc_iid} with $V=d_{\mathrm{VC}}$ and $\delta'':=\delta/(64KH)$ to the layer-$h$ features of the three i.i.d.\ samples $(\tau_i)_{i\le W}$, $(\tau_m')_{m\le W}$ and $(\tau_i)_{i\in I}$; the corresponding additive terms $\epsilon_W,\epsilon_W,\epsilon_{\lceil W/2\rceil}$ are at most $\bar\epsilon/2,\bar\epsilon/2,\bar\epsilon$, because $\log(8/\delta'')=\log(512KH/\delta)$. On the intersection of the three resulting events with $\mathcal C$, every $T\in\mathcal T$ satisfies
\[
  \frac1W\sum_{i=1}^W\ind\{\phi_h(\tau^{(i)})\in T\}\le\frac1W\sum_{i=1}^W\ind\{\phi_h(\tau_i)\in T\}+\frac1W\sum_{m=1}^W\ind\{\phi_h(\tau_m')\in T\}\le 3\mu(T)+\bar\epsilon,
\]
by \cref{lem:lin_coupling}\ref{cpl:inj}. By \cref{lem:lin_coupling}\ref{cpl:count} more than $9W/10\ge\lceil W/2\rceil$ indices have $B_i=1$, so $\tau^{(i)}=\tau_i$ for all $i\in I$ by \cref{lem:lin_coupling}\ref{cpl:eq}, and
\[
  \frac1W\sum_{i=1}^W\ind\{\phi_h(\tau^{(i)})\in T\}\ge\frac{|I|}{W}\cdot\frac{1}{|I|}\sum_{i\in I}\ind\{\phi_h(\tau_i)\in T\}\ge\frac12\cdot\frac23\bigl(\mu(T)-\bar\epsilon\bigr).
\]
The same two chains of inequalities hold for nonnegative functions $f$ with level sets in $\mathcal T$. The total failure probability is at most $3\delta''+2e^{-W/100}+2W\etail^{\mathrm{lin}}\le\frac{3\delta}{64KH}+2e^{-2\iota}+2Ke^{-4\iota}\le\frac{\delta}{16KH}$, using $W\ge 200\iota$ and $e^{-2\iota}=(64KdHA/\delta)^{-8}$.
\end{proof}

\begin{lemma}[Period covariance concentration]\label{lem:lin_blk_good}
For each period $r\ge 2$, let $t_{\mathrm{ref}}(r):=(r-2)\Lmax+1$ be the first episode of period $r-1$, write $\Sigma_{h,j}^\pi := \E_\pi[\ind\{\phi_h\in\cC_{r,h,j}\}\phi_h\phi_h^\top]$, and let $\mathcal E_{\mathrm{per},r-1}\in\F_{(r-1)\Lmax}$ be the event that, for every $h\in[H]$ and $j\in[J_{r,h}]$,
\begin{gather*}
  \Prob_{\pi^{t_{\mathrm{ref}}(r)}}\Bigl(\phi_h\notin\bigcup_{j'}\cC_{r,h,j'}\Bigr)\le 4\theta,\qquad
  \Prob_{\pi^{t_{\mathrm{ref}}(r)}}(\phi_h\in\cC_{r,h,j})\le 4\phat_{r,h,j},\\
  \tfrac14\phat_{r,h,j}\Sigmahat_{r,h,j}\preceq \Pi_{r,h,j}\Sigma_{h,j}^{\pi^{t_{\mathrm{ref}}(r)}}\Pi_{r,h,j}\preceq 4\phat_{r,h,j}\Sigmahat_{r,h,j}.
\end{gather*}
Set $\mathcal E_{\mathrm{per}} := \bigcap_{r\ge 2}\mathcal E_{\mathrm{per},r-1}$. Then $\Prob(\mathcal E_{\mathrm{per}})\ge 1-\delta/4$, and on $\mathcal E_{\mathrm{per},r-1}$ every episode $k$ of period $r$ satisfies
\[
  \Prob_{\pi^k}\Bigl(\phi_h\notin\bigcup_{j'}\cC_{r,h,j'}\Bigr)\le 5\theta,\qquad
  \Prob_{\pi^k}(\phi_h\in\cC_{r,h,j})\le 4.3\,\phat_{r,h,j},\qquad
  \tfrac15\phat_{r,h,j}\Sigmahat_{r,h,j}\preceq \Pi_{r,h,j}\Sigma_{h,j}^{\pi^k}\Pi_{r,h,j}\preceq 5\phat_{r,h,j}\Sigmahat_{r,h,j}.
\]
\end{lemma}

\begin{proof}
Fix a period $r\ge 2$ and a layer $h$, abbreviate $t_{\mathrm{ref}}=t_{\mathrm{ref}}(r)$, and work on the event of \cref{lem:lin_uc_drift}. The shells $\cC_j=\cC_{r,h,j}$ are computed from the layer-$h$ features of period $r-1$, so by \cref{lem:lin_peeling_det} the empirical frequency of $\cC_j$ in \eqref{eq:lin_uc} is $\phat_j=\phat_{r,h,j}>\theta/2$ and the empirical frequency of the complement of the union of the shells is at most $\theta$. By the definition of $\theta$ in \eqref{eq:lin_blk_params}, $\log(2e\Lmax)\le\iota$ and $\log(512KH/\delta)\le 5\iota\le\iota^2$,
\[
  2\rho\,\bar\epsilon\le\frac{18\rho(d_{\mathrm{VC}}\iota+\iota^2)}{\Lmax}\le\frac{18}{400}\,\theta\le 0.09\,\phat_j ,
\]
and in particular $\bar\epsilon\le 0.05\,\theta\le 0.1\,\phat_j$, since $\rho\ge 2$. Both sets belong to $\mathcal T$ by \cref{lem:lin_vc_class}, so the left inequality of \eqref{eq:lin_uc} gives $\Prob_{\pi^{t_{\mathrm{ref}}}}(\phi_h\in\cC_j)\le 3\phat_j+\bar\epsilon\le 3.1\,\phat_j$ and $\Prob_{\pi^{t_{\mathrm{ref}}}}(\phi_h\notin\bigcup_{j'}\cC_{j'})\le 3\theta+\bar\epsilon\le 3.05\,\theta$.

For the covariance, fix a shell $\cC_j$ and a vector $x$ in the range of $\Sigmahat_j$ with $x^\top\Sigmahat_jx=1$, and let $f(\phi):=\ind\{\phi\in\cC_j\}(x^\top\phi)^2/(2\rho)$. For $\phi\in\cC_j$, the Cauchy--Schwarz inequality and \cref{lem:lin_peeling_det} give $(x^\top\phi)^2=(x^\top\Pi_j\phi)^2\le(x^\top\Sigmahat_jx)(\phi^\top\Sigmahat_j^\dagger\phi)\le 2\rho$, so $f$ takes values in $[0,1]$, and its level sets belong to $\mathcal T$ by \cref{lem:lin_vc_class}. Since $x$ is orthogonal to the eigenvectors whose eigenvalues are set to zero in $\Sigmahat_j$, the empirical mean of $2\rho f$ over period $r-1$ equals $\phat_j\,x^\top\Sigmahat_jx=\phat_j$. Applying \eqref{eq:lin_uc} to $f$ therefore gives
\[
  0.3\,\phat_j\le\frac{\phat_j-2\rho\bar\epsilon}{3}\le x^\top\Sigma_{h,j}^{\pi^{t_{\mathrm{ref}}}}x\le 3\phat_j+2\rho\bar\epsilon\le 3.1\,\phat_j ,
\]
which, by homogeneity in $x$, is the two-sided matrix inequality $0.3\,\phat_j\Sigmahat_j\preceq\Pi_j\Sigma_{h,j}^{\pi^{t_{\mathrm{ref}}}}\Pi_j\preceq 3.1\,\phat_j\Sigmahat_j$. Thus $\mathcal E_{\mathrm{per},r-1}$ holds on the event of \cref{lem:lin_uc_drift} for all $h\in[H]$, and a union bound over the at most $K$ periods and the $H$ layers gives $\Prob(\mathcal E_{\mathrm{per}})\ge 1-\delta/16\ge 1-\delta/4$.

It remains to pass from $\pi^{t_{\mathrm{ref}}}$ to the policy of an episode $k$ in period $r$, starting from the constants in the definition of $\mathcal E_{\mathrm{per},r-1}$. By \cref{lem:lin_blk_stab}\ref{lblk:drift}, the per-step log-ratios between $\pi^k$ and $\pi^{t_{\mathrm{ref}}}$ are at most $2\vartheta_{\Lmax}$, and \cref{lem:pathlaw} with $y=4\iota$ gives $\E_{\pi^k}[F]\le e^{0.04}\E_{\pi^{t_{\mathrm{ref}}}}[F]+F_{\max}\etail^{\mathrm{lin}}$ and $\E_{\pi^k}[F]\ge e^{-0.04}(\E_{\pi^{t_{\mathrm{ref}}}}[F]-F_{\max}\etail^{\mathrm{lin}})$ for every function $F$ of the trajectory with values in $[0,F_{\max}]$. Taking $F$ to be the indicator of $\{\phi_h\notin\bigcup_{j'}\cC_{j'}\}$, the indicator of $\{\phi_h\in\cC_j\}$, and $\ind\{\phi_h\in\cC_j\}(x^\top\phi_h)^2\in[0,2\rho]$ for $x$ as above, and using $e^{0.04}<1.041$, $e^{-0.04}>0.96$ and $2\rho\,\etail^{\mathrm{lin}}\le 10^{-20}\theta\le 10^{-19}\phat_j$, we obtain the bounds $5\theta$, $4.3\,\phat_j$, and $\frac15\phat_j\Sigmahat_j\preceq\Pi_j\Sigma_{h,j}^{\pi^k}\Pi_j\preceq 5\phat_j\Sigmahat_j$.
\end{proof}

\subsection{Exchangeable Matrix Self-Normalization and Main Linear Regret Proof}\label{sec:lin_blk_assembly}

\begin{lemma}[Exchangeable matrix self-normalization]\label{lem:exch_matrix_sn}
Let $V\subseteq\R^d$ be a subspace with orthogonal projection $\Pi_V$, and let $(X_k,u_k)_{k=1}^\Lambda$ be an exchangeable sequence in $\{0,1\}\times V$ with $\|X_ku_k\|_2\le R$ almost surely for a constant $R$. Let $G:=\sum_{k=1}^\Lambda X_ku_ku_k^\top$, and let $\acc\in\{0,1\}$ be a measurable function of the sequence that is invariant under permutations and such that $G\succeq\alpha\Pi_V$ whenever $\acc=1$, for some $\alpha>0$. Then, for all fixed $\xi_1,\dots,\xi_\Lambda\in V$,
\[
  \E\Bigl[\acc\cdot\Lambda\,G^\dagger\sum_{k=1}^\Lambda X_ku_k\ip{u_k}{\xi_k}\Bigr]=\Prob(\acc=1)\sum_{k=1}^\Lambda \xi_k.
\]
\end{lemma}

\begin{proof}
For each $k\in[\Lambda]$, define the random matrix $T_k := \acc\,G^\dagger X_k u_ku_k^\top$. On $\{\acc=1\}$ we have $G^\dagger\preceq\alpha^{-1}\Pi_V$, so $\|T_k\|_{\mathrm{op}}\le R^2/\alpha$ almost surely and $\E[T_k]$ is finite. Since $(X_{k'},u_{k'})_{k'=1}^\Lambda$ is exchangeable and both $G$ and $\acc$ are invariant under permutations of this sequence, $\E[T_k]$ does not depend on $k$. Summing over $k$ on every sample path gives $\sum_{k=1}^\Lambda T_k = \acc\,G^\dagger G = \acc\,\Pi_V$: every $u_k$ lies in $V$, so the range of $G$ is contained in $V$, and $G\succeq\alpha\Pi_V$ on $\{\acc=1\}$ makes it equal to $V$. Taking expectations, $\Lambda\,\E[T_k]=\Prob(\acc=1)\,\Pi_V$ for every $k$. Finally,
\[
  \E\Bigl[\acc\cdot\Lambda\,G^\dagger\sum_{k=1}^\Lambda X_ku_k\ip{u_k}{\xi_k}\Bigr]=\sum_{k=1}^\Lambda\Lambda\,\E[T_k]\,\xi_k=\Prob(\acc=1)\sum_{k=1}^\Lambda\Pi_V\xi_k=\Prob(\acc=1)\sum_{k=1}^\Lambda \xi_k,
\]
because each $\xi_k$ is fixed and lies in $V$.
\end{proof}

\begin{lemma}[Within-block matrix self-normalization bias under drift]\label{lem:lin_blk_bias}
Fix period $r\ge 2$, shell $\cC=\cC_{r,h,j}$, and complete block $b=\{t_0,\dots,t_0+\Lambda-1\}$ ($\Lambda=\Lambda_{r,h,j}$). Note that $\mathcal E_{\mathrm{per},r-1}\in\F_{(r-1)\Lmax}\subseteq\F_{t_0-1}$. Almost surely on $\mathcal E_{\mathrm{per},r-1}$, every $(s,a)\in\Sset_h\times\Aset$ with $\phi(s,a)\in\cC$ satisfies
\[
  \Bigl|\E\bigl[\widehat L_b(s,a)\mid\F_{t_0-1}\bigr] - \sum_{k\in b}v_{k,h}(s,a)\Bigr|
  \le 2019\sqrt{\rho}\,m_0 H\Lambda\varrho_b^{\mathrm{lin}} + \frac12\Lambda K^{-3}.
\]
Here $\E[\widehat L_b(s,a)\mid\F_{t_0-1}]$ denotes the version $(\Sigmahat_{r,h,j}^{-1/2}\phi(s,a))^\top\E[\what_b\mid\F_{t_0-1}]$, which is defined for all pairs simultaneously and is $\F_{t_0-1}$-measurable, whereas the sum is a random variable that depends on the policies played inside the block. The inequality is between these two random variables: it holds for every realization of the episodes of the block.
\end{lemma}

\begin{proof}
Condition on $\F_{t_0-1}$ on $\mathcal E_{\mathrm{per},r-1}$, and construct an i.i.d.\ reference stream $(\tau^{\circ,k})_{k\in b}\stackrel{\mathrm{i.i.d.}}{\sim}\Prob_{\pi^{t_0}}$ sequentially: at each $k\in b$, given $\F_{k-1}$ and $\tau^{\circ,<k}$, draw $\tau^{\circ,k}\sim\Prob_{\pi^{t_0}}$ independently and draw $\tau^k\sim\Prob_{\pi^k}$ from the maximal coupling of $(\Prob_{\pi^{t_0}},\Prob_{\pi^k})$ given $\tau^{\circ,k}$. Let $\widehat L_b^\circ(s,a)$ and $\acc_b^\circ$ be the block quantities computed on $(\tau^{\circ,k})_{k\in b}$ using period $r$'s fixed $(u_r,b_r)$, and set $A_\cC:=\{\tau:\phi_h(\tau)\in\cC\}$. Because $(u_r,b_r)$ are fixed throughout period $r$, $Y_h^k(\tau^k) = Y_h^{\circ,k}(\tau^{\circ,k})$ whenever $\tau^k=\tau^{\circ,k}$, so $\widehat L_b(s,a)=\widehat L_b^\circ(s,a)$ identically on $\mathcal G_{\mathrm{cpl},b} := \{\forall k\in b:\tau^k=\tau^{\circ,k}\text{ or }(\tau^k\notin A_\cC\text{ and }\tau^{\circ,k}\notin A_\cC)\}$. Under the maximal coupling, $\Prob(\tau^k\ne\tau^{\circ,k}\text{ and }(\tau^k\in A_\cC\text{ or }\tau^{\circ,k}\in A_\cC)\mid\F_{k-1})\le \int_{A_\cC}|d\Prob_{\pi^k}-d\Prob_{\pi^{t_0}}|$. By \cref{lem:lin_blk_stab}\ref{lblk:drift}, \cref{lem:pathlaw}, and $\Prob_{\pi^{t_0}}(\phi_h\in\cC)\le 4.3\phat_{r,h,j}$, this integral is at most $(e^{\varrho_b^{\mathrm{lin}}}-e^{-\varrho_b^{\mathrm{lin}}})(4.3\phat_{r,h,j})+2\etail^{\mathrm{lin}} \le 9\varrho_b^{\mathrm{lin}}\phat_{r,h,j}+2\etail^{\mathrm{lin}}$. Summing over $k\in b$ ($\Lambda<2m_0/\phat_{r,h,j}$) gives $\Prob(\mathcal G_{\mathrm{cpl},b}^c\mid\F_{t_0-1})\le 18m_0\varrho_b^{\mathrm{lin}}+2\Lambda\etail^{\mathrm{lin}}$. Since $|\widehat L_b-\widehat L_b^\circ|\le 112\sqrt{\rho}\,H\Lambda$ by \cref{lem:lin_blk_stab}\ref{lblk:path}, we obtain $|\E[\widehat L_b-\widehat L_b^\circ\mid\F_{t_0-1}]|\le 2016\sqrt{\rho}\,m_0 H\Lambda\varrho_b^{\mathrm{lin}}+\frac18\Lambda K^{-3}$.

Next, let $v_{k,h}^\circ(s,a)=\ip{\phi(s,a)}{\xi_{k,h}^\circ}$ ($\xi_{k,h}^\circ\in\F_{t_0-1}$) evaluate \cref{eq:lin_vz} under $\pi^{t_0}$. Given $(\phi_h^{\circ,k'})_{k'\in b}$, we have $\E[Y_h^{\circ,k}\mid\F_{t_0-1},(\phi_h^{\circ,k'})_{k'\in b}] = \ip{\Pi\phi_h^{\circ,k}}{\xi_{k,h}^\circ} + \ip{(I-\Pi)\phi_h^{\circ,k}}{\xi_{k,h}^\circ}$. Apply \cref{lem:exch_matrix_sn} conditionally on $\F_{t_0-1}$, with $V=\operatorname{range}(\Sigmahat_{r,h,j})$, the fixed vectors $\xi_k:=\Sigmahat_{r,h,j}^{1/2}\xi_{k,h}^\circ$ and $\acc=\acc_b^\circ$, to the projected part and \cref{lem:lin_peeling_det} ($\|(I-\Pi)\phi\|_2\le\sqrt{2\rho\lambda}$) to the residual. By Matrix Chernoff \citep[Theorem~1.1]{tropp2012user} ($\sum_{k\in b}\E[X_k^\circ u_k^\circ(u_k^\circ)^\top\mid\F_{t_0-1}]\succeq\frac{m_0}{5}\Pi$, $X_k^\circ u_k^\circ(u_k^\circ)^\top\preceq 2\rho\Pi$) and \cref{eq:binom_tails} applied at $5\iota$ ($\mu=\Lambda\Prob_{\pi^{t_0}}(\phi_h\in\cC)<8.6m_0$, so $2\mu+15\iota<18m_0<24m_0$), we have $\Prob(\acc_b^\circ=0\mid\F_{t_0-1})\le d e^{-m_0/(80\rho)}+e^{-5\iota}\le(d+1)e^{-5\iota}$, giving $|\E[\widehat L_b^\circ(s,a)\mid\F_{t_0-1}]-\sum_{k\in b}v_{k,h}^\circ(s,a)|\le \frac14\Lambda K^{-3}$. Finally, applying \cref{lem:pathlaw} to the suffix rollout in $[0,2H]$ gives $|v_{k,h}(s,a)-v_{k,h}^\circ(s,a)|\le (e^{\varrho_b^{\mathrm{lin}}}-1)(2H)+2H\etail^{\mathrm{lin}}\le 3\varrho_b^{\mathrm{lin}} H+\frac18 K^{-3}$ for each $k\in b$, completing the proof.
The additive terms of order $\Lambda K^{-3}$ and $K^{-3}$ in this proof absorb quantities of order $\Lambda\,\poly(d,H)\bigl(\etail^{\mathrm{lin}}+\sqrt\lambda+(d+1)e^{-5\iota}\bigr)$, which is legitimate in the standing regime $200\iota\le\Lmax\le K/4$ because $\sqrt\lambda=(KdH)^{-10}$, $\etail^{\mathrm{lin}}<(64KdHA)^{-16}$, $\rho\le 42\,d\log(KdH)$ and $\Lambda\le K$.
\end{proof}

\paragraph{Blocks and their filtration.}
Let $K' := \lfloor K/\Lmax\rfloor\Lmax$ and let $\cK:=\{\Lmax+1,\dots,K'\}$ be the episodes of the complete periods $r\ge 2$. In these periods every block length divides $\Lmax$, so every block of every shell is complete. For a layer $h$ and a shell index $j\in[J_0]$, let $\mathcal B_{h,j}$ be the sequence of blocks of the shells $\cC_{r,h,j}$ over the complete periods $r\ge 2$ with $j\le J_{r,h}$, in chronological order. For a block $b$ we write $r(b)$ for its period, $\cC_b$ for its shell, $\phat_b$ for the mass of the shell, $\Lambda_b$ for its length, $t_0(b)$ and $t_1(b)$ for its first and last episodes, and $\mathcal P_b:=\{(s,a)\in\Sset_h\times\Aset:\phi(s,a)\in\cC_b\}$. The blocks of $\mathcal B_{h,j}$ are disjoint, and the start of a block is determined at the beginning of its period, so the blocks are indexed by stopping times exactly as in \cref{app:events}: the statistics of the $i$th block are measurable with respect to the history up to its last episode, which precedes the first episode of the $(i+1)$st block.

\begin{lemma}[Block supermartingales]\label{lem:lin_supermart}
Fix $h\in[H]$, $j\in[J_0]$ and $\delta'\in(0,1)$. For every block $b\in\mathcal B_{h,j}$ let $\mu_b$ be a measure on $\Sset_h\times\Aset$ with total mass at most one that is supported on $\mathcal P_b$ and determined by $\F_{t_0(b)-1}$. With probability at least $1-2\delta'$, both of the following hold simultaneously for every initial segment $\mathcal B'$ of $\mathcal B_{h,j}$, where $\E_b[\cdot]:=\E[\cdot\mid\F_{t_0(b)-1}]$ and $\Qtil_b=\Qhat_b-\eta\Qhat_b^2$:
\begin{align}
  \sum_{b\in\mathcal B'}\int\Qtil_b\,d\mu_b &\le \sum_{b\in\mathcal B'}\int\E_b[\Qhat_b]\,d\mu_b+\frac{\log(1/\delta')}{\eta},\label{eq:lin_sm_cmp}\\
  -\sum_{b\in\mathcal B'}\int\Qtil_b\,d\mu_b &\le \sum_{b\in\mathcal B'}\int\bigl(-\E_b[\Qhat_b]+2\eta\,\E_b[\Qhat_b^2]\bigr)\,d\mu_b+\frac{\log(1/\delta')}{\eta}.\label{eq:lin_sm_lrn}
\end{align}
\end{lemma}

\begin{proof}
By \cref{lem:lin_blk_stab}\ref{lblk:path}, $u:=\eta\Qhat_b(s,a)$ satisfies $|u|\le\frac18$ for all $(s,a)\in\mathcal P_b$. For such $u$ we have $e^{u-u^2}\le 1+u$ and $e^{-u+u^2}\le 1-u+2u^2$. Let $\|\mu_b\|\le 1$ denote the total mass of $\mu_b$. Since $\exp$ is convex and $\mu_b$ together with a point mass $1-\|\mu_b\|$ at zero is a probability measure, Jensen's inequality gives
\[
  \exp\Bigl(\eta\int\Qtil_b\,d\mu_b\Bigr)\le\int e^{\eta\Qhat_b-\eta^2\Qhat_b^2}\,d\mu_b+\bigl(1-\|\mu_b\|\bigr)\le 1+\eta\int\Qhat_b\,d\mu_b .
\]
Taking $\E_b$, exchanging it with the integral ($\Qhat_b$ is bounded and $\mu_b$ is determined by $\F_{t_0(b)-1}$), and using $1+x\le e^x$ shows that $\E_b\bigl[\exp\bigl(\eta\int(\Qtil_b-\E_b[\Qhat_b])\,d\mu_b\bigr)\bigr]\le 1$. The partial products of these exponentials along $\mathcal B_{h,j}$ therefore form a nonnegative supermartingale with respect to the block filtration, and Ville's inequality gives \eqref{eq:lin_sm_cmp} with probability at least $1-\delta'$. The same argument with $e^{-u+u^2}\le 1-u+2u^2$ gives $\E_b\bigl[\exp\bigl(-\eta\int\Qtil_b\,d\mu_b\bigr)\bigr]\le\exp\bigl(\eta\int(-\E_b[\Qhat_b]+2\eta\E_b[\Qhat_b^2])\,d\mu_b\bigr)$ and hence \eqref{eq:lin_sm_lrn}.
\end{proof}

\begin{proof}[Proof of \cref{thm:linear_block}]
If the algorithm returns at line~1 the theorem holds by the paragraph on boundary regimes, so assume $200\iota\le\Lmax\le K/4$ and $\theta\le\frac12$.

\emph{Step 0 (episodes outside $\cK$).} The first period and a possibly incomplete last period contain fewer than $2\Lmax$ episodes, each with regret at most $H$. They contribute at most $2H\Lmax$ to $\Reg$.

\emph{Step 1 (decomposition).} For $k\in\cK$ and $s\in\Sset_h$ let $\Phi_{k,h}(s):=\ip{\pi^k(\cdot\mid s)}{z_{k,h}(s,\cdot)}-\ip{\pi^\star(\cdot\mid s)}{v_{k,h}(s,\cdot)}$. By \cref{lem:lin_decomp}\ref{linD:regret},
\begin{equation}\label{eq:lin_proof_decomp}
  \sum_{k\in\cK}\bigl(V^k(s_1)-V^{\pi^\star}(s_1;c^k)\bigr)=\sum_{h=1}^H\sum_{s\in\Sset_h}q^\star(s)\sum_{k\in\cK}\Phi_{k,h}(s)+\sum_{k\in\cK}\sum_{h=1}^H\sum_{s\in\Sset_h,a}q^k(s,a)\,r_{k,h}(s,a).
\end{equation}

\emph{Step 2 (exponential weights).} Fix $h$ and $s\in\Sset_h$, and write all inner products at $s$. For every $k$,
\[
  \Phi_{k,h}(s)=\ip{\pi^k-\pi^\star}{g_k}+\ip{\pi^k}{z_{k,h}-g_k}+\ip{\pi^\star}{g_k-v_{k,h}} .
\]
By \cref{lem:lin_blk_stab}\ref{lblk:path}, $\eta|g_k(s,a)|\le 65\eta\sqrt\rho\,H\Lmax<1$. The policy $\pi^k(\cdot\mid s)$ is the exponential-weights distribution for the losses $g_1(s,\cdot),g_2(s,\cdot),\dots$, and $g_k\equiv 0$ in the first period, so the standard second-order bound for exponential weights (using $e^{-x}\le 1-x+x^2$ for $x\ge -1$) gives $\sum_{k\in\cK}\ip{\pi^k-\pi^\star}{g_k}\le\frac{\log A}{\eta}+\eta\sum_{k\in\cK}\sum_a\pi^k(a\mid s)g_k(s,a)^2$. Hence
\begin{equation}\label{eq:lin_proof_local}
  \sum_{k\in\cK}\Phi_{k,h}(s)\le\frac{\log A}{\eta}+\sum_{k\in\cK}\sum_{a\in\Aset}\Bigl[\pi^k(a\mid s)\bigl(z_{k,h}-g_k+\eta g_k^2\bigr)(s,a)+\pi^\star(a\mid s)\bigl(g_k-v_{k,h}\bigr)(s,a)\Bigr].
\end{equation}

\emph{Step 3 (pairs outside the shells).} Let $k\in\cK$ belong to period $r$ and let $u_r(s,a)=1$. Then $z_{k,h}(s,a)=0$, $g_k(s,a)=-K^{-3}-\eta K^{-6}$ and $v_{k,h}(s,a)\ge 0$, so the comparator term in \eqref{eq:lin_proof_local} is nonpositive and the learner term is at most $\pi^k(a\mid s)\bigl(K^{-3}+\eta K^{-6}+4\eta K^{-6}\bigr)\le 2K^{-3}\pi^k(a\mid s)$. Summed over $a$ and $k\in\cK$, these pairs contribute at most $2K^{-2}$ to \eqref{eq:lin_proof_local}.

\emph{Step 4 (pairs on a shell).} Let $b\in\mathcal B_{h,j}$ with $t_0=t_0(b)$, $t_1=t_1(b)$, $\Lambda=\Lambda_b$, period $r$, and let $(s,a)\in\mathcal P_b$. On the block, $z_{k,h}(s,a)=v_{k,h}(s,a)-b_r(s,a)\in[-\frac12,2H]$, $g_k(s,a)=0$ for $k\ne t_1$ and $g_{t_1}(s,a)=\Qtil_b(s,a)$. By \cref{lem:lin_blk_stab}\ref{lblk:drift}, $|\pi^k(a\mid s)-\pi^{t_0}(a\mid s)|\le(e^{\vartheta_b}-1)\pi^{t_0}(a\mid s)\le 2\vartheta_b\,\pi^{t_0}(a\mid s)$ for all $k\in b$, and by \cref{lem:lin_blk_stab}\ref{lblk:path}, $|\Qtil_b(s,a)|\le 65\sqrt\rho\,H\Lambda$. Therefore the contribution of $(s,a)$ and the episodes of $b$ to the sum in \eqref{eq:lin_proof_local} is at most
\[
  \pi^{t_0}(a\mid s)\Bigl[\sum_{k\in b}\bigl(v_{k,h}(s,a)-b_r(s,a)\bigr)-\Qtil_b(s,a)+\mathsf e_b\Bigr]+\pi^\star(a\mid s)\Bigl[\Qtil_b(s,a)-\sum_{k\in b}v_{k,h}(s,a)\Bigr],
\]
where $\mathsf e_b:=2\vartheta_b\bigl(2H\Lambda+65\sqrt\rho\,H\Lambda\bigr)+(1+2\vartheta_b)\,\eta\,(65\sqrt\rho\,H\Lambda)^2\le 21704\,\eta\rho H^2\Lambda^2$ collects the drift of $\pi^k(a\mid s)$ within the block and the second-order term of exponential weights; we used $\vartheta_b=130\eta\sqrt\rho\,H\Lambda<\frac1{150}$ and $2+65\sqrt\rho\le 67\sqrt\rho$.

\emph{Step 5 (concentration).} For $b\in\mathcal B_{h,j}$ define the measures $\nu_b(s,a):=q^\star(s)\pi^{t_0(b)}(a\mid s)\ind\{(s,a)\in\mathcal P_b\}$ and $\mu_b^\star(s,a):=q^\star(s)\pi^\star(a\mid s)\ind\{(s,a)\in\mathcal P_b\}$ on $\Sset_h\times\Aset$. Both have total mass at most $\sum_{s\in\Sset_h}q^\star(s)=1$ and are determined by $\F_{t_0(b)-1}$, because $q^\star$ and $\pi^\star$ are deterministic and the shell is fixed at the start of the period. Apply \cref{lem:lin_supermart} with $\delta':=\delta/(16HJ_0)$ to $(\mu_b^\star)_b$, using \eqref{eq:lin_sm_cmp}, and to $(\nu_b)_b$, using \eqref{eq:lin_sm_lrn}, for every $(h,j)\in[H]\times[J_0]$. By a union bound, with probability at least $1-\delta/4$, for all $(h,j)$,
\begin{align*}
  \sum_{b\in\mathcal B_{h,j}}\int\Qtil_b\,d\mu_b^\star&\le\sum_{b\in\mathcal B_{h,j}}\int\E_b[\Qhat_b]\,d\mu_b^\star+\frac{\iota}{\eta},\\
  -\sum_{b\in\mathcal B_{h,j}}\int\Qtil_b\,d\nu_b&\le\sum_{b\in\mathcal B_{h,j}}\int\bigl(-\E_b[\Qhat_b]+6498\,\eta\rho H^2\Lambda_b^2\bigr)\,d\nu_b+\frac{\iota}{\eta},
\end{align*}
where we used $\log(16HJ_0/\delta)\le\iota$ and $\E_b[\Qhat_b^2]\le(57\sqrt\rho\,H\Lambda_b)^2$.

\emph{Step 6 (bias and the bonus).} Work on $\mathcal E_{\mathrm{per}}$. For $b\in\mathcal B_{h,j}$ in period $r$ and $(s,a)\in\mathcal P_b$, the definition of $\Qhat_b$ and \cref{lem:lin_blk_bias} give
\[
  \Bigl|\E_b[\Qhat_b(s,a)]+\frac{\Lambda_b}{2}b_r(s,a)+\Lambda_bK^{-3}-\sum_{k\in b}v_{k,h}(s,a)\Bigr|\le\mathsf B_b:=2019\sqrt\rho\,m_0H\Lambda_b\varrho_b^{\mathrm{lin}}+\frac12\Lambda_bK^{-3}.
\]
Integrating the bound of Step~4 against $q^\star$, summing over the blocks of $\mathcal B_{h,j}$, and inserting the two inequalities of Step~5 and this bias bound, the total contribution of the shells of index $j$ at layer $h$ to $\sum_{s\in\Sset_h}q^\star(s)\sum_{k\in\cK}\Phi_{k,h}(s)$ is at most $\frac{2\iota}{\eta}$ plus
\begin{equation}\label{eq:lin_proof_bracket}
  \sum_{b\in\mathcal B_{h,j}}\biggl\{\int\Bigl(-\frac{\Lambda_b}{2}b_r+\mathsf B_b+\Lambda_bK^{-3}+28202\,\eta\rho H^2\Lambda_b^2\Bigr)d\nu_b+\int\Bigl(-\frac{\Lambda_b}{2}b_r+\mathsf B_b-\Lambda_bK^{-3}\Bigr)d\mu_b^\star\biggr\}.
\end{equation}
Both integrands are nonpositive. Indeed, $b_r=\beta_{\mathrm{blk}}/\phat_b$ on $\mathcal P_b$, $\varrho_b^{\mathrm{lin}}<370\,\eta\sqrt\rho\,H^{3/2}\sqrt\iota\,\Lambda_b$ by \cref{lem:lin_blk_stab}\ref{lblk:drift}, and $\Lambda_b<2m_0/\phat_b$, so
\[
  2019\sqrt\rho\,m_0H\Lambda_b\varrho_b^{\mathrm{lin}}+28202\,\eta\rho H^2\Lambda_b^2\le\bigl(1.4941\times 10^6+5.7\times 10^4\bigr)\frac{\eta\rho\,m_0^2H^{5/2}\sqrt\iota}{\phat_b}\,\Lambda_b<0.39\,\Lambda_b b_r ,
\]
by the choice $c_{\mathrm{blk}}=4\times 10^6$ in \eqref{eq:lin_blk_params}, while $\frac32\Lambda_bK^{-3}<0.05\,\Lambda_bb_r$ because $b_r\ge\beta_{\mathrm{blk}}=200\sqrt\rho\,m_0^2H/\Lmax>30K^{-3}$.

\emph{Step 7 (local regret).} Summing Steps~2--6 over $j\in[J_0]$ and $h\in[H]$, and using $\sum_{s\in\Sset_h}q^\star(s)=1$ for the terms $\frac{\log A}{\eta}$ and $2K^{-2}$, we obtain on $\mathcal E_{\mathrm{per}}$ and the event of Step~5
\[
  \sum_{h=1}^H\sum_{s\in\Sset_h}q^\star(s)\sum_{k\in\cK}\Phi_{k,h}(s)\le\frac{H\log A+2HJ_0\iota}{\eta}+2HK^{-2}.
\]

\emph{Step 8 (cost of optimism).} Let $k\in\cK$ be in period $r$. On a shell, $r_{k,h}=b_r=\beta_{\mathrm{blk}}/\phat_{r,h,j}$, and outside the shells $r_{k,h}=v_{k,h}\le 2H$. By \cref{lem:lin_blk_good}, on $\mathcal E_{\mathrm{per}}$,
\[
  \sum_{s\in\Sset_h,a}q^k(s,a)\,r_{k,h}(s,a)\le\sum_{j=1}^{J_{r,h}}4.3\,\phat_{r,h,j}\cdot\frac{\beta_{\mathrm{blk}}}{\phat_{r,h,j}}+2H\cdot 5\theta\le 5J_0\beta_{\mathrm{blk}}+10H\theta ,
\]
so the second sum in \eqref{eq:lin_proof_decomp} is at most $HK(5J_0\beta_{\mathrm{blk}}+10H\theta)$.

\emph{Step 9 (conclusion).} The events $\mathcal E_{\mathrm{per}}$ and that of Step~5 hold simultaneously with probability at least $1-\delta/2$. On them, Steps~0, 7 and~8 give
\[
  \Reg\le 2H\Lmax+\frac{H\log A+2HJ_0\iota}{\eta}+2HK^{-2}+HK\bigl(5J_0\beta_{\mathrm{blk}}+10H\theta\bigr).
\]
With $\eta^{-1}=\tO(\sqrt d\,H^{3/2}\Lmax)$, $\beta_{\mathrm{blk}}=\tO(d^{5/2}H/\Lmax)$ and $\theta=\tO((d^{5/2}H+d^3)/\Lmax)$ from \eqref{eq:lin_blk_params}, this is
\begin{align*}
  \Reg &= \tO\!\left(\sqrt{d}\,H^{5/2}\Lmax + \frac{(d^{5/2}H^3 + d^3 H^2)K}{\Lmax}\right)
        = \tO\!\Bigl(\bigl(d^{3/2}H^{11/4} + d^{7/4}H^{9/4}\bigr)\sqrt{K}\Bigr),
\end{align*}
where the last step uses $\Lmax=\Theta\bigl(d\sqrt{(H^{1/2}+\sqrt{d/H})K}\bigr)$ and $\sqrt{H^{1/2}+\sqrt{d/H}}\le H^{1/4}+d^{1/4}H^{-1/4}$.

\emph{Running time.} \cref{alg:regions} takes $O(\Lmax^2 d^2+\Lmax d^3)$ time per period and layer, each block end computes $G_b^\dagger y_b\in\R^d$ in $O(d^3)$ time, and the policy at a visited state is evaluated from the stored shells and vectors $\what_b$ in time polynomial in $d,H,A,K$. No optimization oracle is called.
\end{proof}

\subsection{Comparison with the Earlier Linear Bound}\label{app:lin_compare}

The $\tO(d^3H^4\sqrt K)$ bound of \citet{kulkarni2026introducing} is stated for regions built from the features of the last $\sqrt K$ episodes, and their proof gives $\tO((d^2H^4+d^3H^3)\sqrt K)$. The length of this window is not optimized there. With a window of $W$ episodes, their step size must satisfy $1/\eta=\tOmega(dH^3W)$ for the policies to stay close over the window, and their threshold scales as $\tTheta(d^3/W)$ through uniform convergence over $W$ samples, so the two leading terms of their bound become $\tO(d^2H^4W)$ and $\tO(d^3H^3K/W)$. The choice $W=\sqrt{dK/H}$ would then give $\tO(d^{5/2}H^{7/2}\sqrt K)$. This is our reading of their proof: the conditions of their Eq.~(65) and the steps that use the window scale as stated when $W$ replaces $\sqrt K$, and the remaining terms of their final display become $\tO(dH^2K/W+H^2W)$, which are of lower order. They do not state this bound, and we have not re-derived their lemmas. Against it, the bound of \cref{thm:linear_block} gains a factor $dH^{3/4}$ when $H\ge\sqrt d$ and $d^{3/4}H^{5/4}$ otherwise. Our proof performs the corresponding balance through $\Lmax$. At equal window lengths the differences are these. The pathwise bound $\tO(\sqrt d\,H\Lambda)$ of the block estimate (\cref{lem:lin_blk_stab}) and the trajectory bound of \cref{lem:pathlaw} allow $1/\eta=\tTheta(\sqrt d\,H^{3/2}\Lmax)$ instead of $\tTheta(dH^3W)$; the blocks of a shell are disjoint, so the concentration terms cost $\tO(H/\eta)$ instead of $\tO(dH/\eta)$; and the cost of optimism is $\tO((d^{5/2}H^3+d^3H^2)K/\Lmax)$ instead of $\tO(d^3H^3K/W)$.

\subsection{Aggregate Feedback in Linear AMDPs}\label{sec:lin_agg_proof}

In tabular AMDPs, \aggalg{} maintains one estimator per state, and the prefix cost $\pre^\pi(s):=\E_{p,\pi}[\sum_{m<h(s)}c(s_m,a_m)\mid s_{h(s)}=s]$ cancels. In linear AMDPs, $\pre^\pi(s)$ depends on how $s$ is reached, that is, on $\psi_{h-1}(s)$, and not on $\phi(s,a)$, so the conditional mean of the episode total is not linear in the features. The following proposition shows that layerwise least squares on episode totals can then be misled under every policy.

\begin{proposition}[Prefix costs break linearity under aggregate feedback]\label{prop:lin_agg_barrier}
There is a linear AMDP (\cref{def:linear}) with $d=2$, $H=3$, $A=2$, costs in $[0,1]$ that do not change over episodes, and a state $s_\star\in\Sset_3$ such that for \emph{every} Markov policy $\pi$: $q^\pi(s)=\frac13$ for each $s\in\Sset_3$; the function $\bar C^\pi(s,a):=\pre^\pi(s)+c(s,a)$ on $\Sset_3\times\Aset$ is not a function of $\phi(s,a)$; $\E_{p,\pi}[\phi_3\phi_3^\top]\succeq\frac13 I_2$; and the population least-squares vector $w_{\mathrm{LS}}^\pi := (\E_{p,\pi}[\phi_3\phi_3^\top])^{-1}\E_{p,\pi}[\phi_3\,C]$ of the episode total $C$ on the layer-$3$ feature satisfies
\[
  Q^\pi(s_\star,a_1) - Q^\pi(s_\star,a_2) = +\tfrac14
  \qquad\text{while}\qquad
  \ip{\phi(s_\star,a_1)-\phi(s_\star,a_2)}{w_{\mathrm{LS}}^\pi} \le -\tfrac14.
\]
Consequently, let $(\pi^k)_{k=1}^K$ be Markov policies and $(\widehat w_k)_{k=1}^K$ vectors with $\sum_{k=1}^K\|\widehat w_k - w_{\mathrm{LS}}^{\pi^k}\|_2 = o(K)$. If the learner at $s_\star$ has sublinear regret with respect to the estimated losses against action $a_1$, that is, $\sum_{k=1}^K\ip{\pi^k(\cdot\mid s_\star)-e_{a_1}}{\ip{\phi(s_\star,\cdot)}{\widehat w_k}} \le o(K)$, then $\Reg = \frac{K}{12}-o(K)$. The same conclusion holds if $\widehat w_k$ approximates $w_{\mathrm{LS}}^{\pi^k}+c_k(1,1)^\top$ for arbitrary scalars $c_k$, which is the least-squares vector obtained when the constant $c_k$ is added to $C$ before the regression.
\end{proposition}

The proposition concerns estimators that regress the episode total on the features of a single layer. Centering the features by $\phi(s,a)-\sum_{a'}\pi^k(a'\mid s)\phi(s,a')$ removes the prefix cost but requires a common block scale across the actions of a state, which conflicts with shell-dependent block lengths. Methods that work with the expected features of whole trajectories, $\E_{p,\pi}[\sum_{h=1}^H\phi(s_h,a_h)]$, are not affected \citep{liu2026computationally}.

\begin{proof}[Proof of \cref{prop:lin_agg_barrier}]
Set $d=2$, $H=3$, $\Aset=\{a_1,a_2\}$, $\Sset_1=\{s_1\}$, $\Sset_2=\{s_2^{(A)},s_2^{(B)}\}$, and $\Sset_3=\{s^{(1)},s^{(2)},s_\star\}$. At layer $1$, set $\phi(s_1,a)=(1,0)^\top$ for both $a\in\Aset$, $\zeta_1=(0,0)^\top$, and $\psi_1(s_2^{(A)})=\psi_1(s_2^{(B)})=(\frac12,0)^\top$, so every policy transitions uniformly to $\{s_2^{(A)},s_2^{(B)}\}$ at zero cost. At layer $2$, set $\phi(s_2^{(A)},a)=(1,0)^\top$ and $\phi(s_2^{(B)},a)=(0,1)^\top$ for both $a\in\Aset$, $\zeta_2=(0,1)^\top$ (so $c(s_2^{(A)},a)=0$ and $c(s_2^{(B)},a)=1$), and define $\psi_2:\Sset_3\to\R^2$ by $\psi_2(s^{(1)})=(\frac23,0)^\top$, $\psi_2(s^{(2)})=(0,\frac23)^\top$, and $\psi_2(s_\star)=(\frac13,\frac13)^\top$ ($\|\sum_{s'\in\Sset_3}|\psi_2(s')|\|_2=\|(1,1)^\top\|_2=\sqrt{2}=\sqrt{d}$). At terminal layer $3$, set $\zeta_3=(\frac14,0)^\top$, $\phi(s^{(1)},a)=(1,0)^\top$ and $\phi(s^{(2)},a)=(0,1)^\top$ for both $a\in\Aset$, and at $s_\star$ set $\phi(s_\star,a_1)=(1,0)^\top$ and $\phi(s_\star,a_2)=(0,1)^\top$. Every condition of \cref{def:linear} holds.

Because transitions and costs at layers $1$ and $2$ do not depend on actions, every Markov policy $\pi$ induces $q^\pi(s^{(1)})=q^\pi(s^{(2)})=q^\pi(s_\star)=\frac13$, with $\pre^\pi(s^{(1)})=0$, $\pre^\pi(s^{(2)})=1$, and $\pre^\pi(s_\star)=\frac12$. Hence $\bar C^\pi(s^{(1)},a_1)=\frac14\ne\frac34=\bar C^\pi(s_\star,a_1)$ despite $\phi(s^{(1)},a_1)=\phi(s_\star,a_1)=(1,0)^\top$ (and $\bar C^\pi(s^{(2)},a_2)=1\ne\frac12=\bar C^\pi(s_\star,a_2)$ despite $\phi(s^{(2)},a_2)=\phi(s_\star,a_2)=(0,1)^\top$). Writing $\alpha_\star:=\pi(a_2\mid s_\star)\in[0,1]$, we have $\E_{p,\pi}[\phi_3\phi_3^\top]=\operatorname{diag}(\frac{2-\alpha_\star}{3},\frac{1+\alpha_\star}{3})\succeq\frac13 I_2$, and
\[
  (w_{\mathrm{LS}}^\pi)_1 = \frac{\frac13\cdot\frac14 + \frac13(1-\alpha_\star)\cdot\frac34}{\frac13 + \frac13(1-\alpha_\star)} = \frac{1 + 3(1-\alpha_\star)}{4(2-\alpha_\star)} \le \frac12,\qquad
  (w_{\mathrm{LS}}^\pi)_2 = \frac{\frac13\cdot 1 + \frac13 \alpha_\star\cdot\frac12}{\frac13 + \frac13 \alpha_\star} = \frac{2+\alpha_\star}{2(1+\alpha_\star)} \ge \frac34.
\]
Thus $\ip{\phi(s_\star,a_1)-\phi(s_\star,a_2)}{w_{\mathrm{LS}}^\pi} = (w_{\mathrm{LS}}^\pi)_1 - (w_{\mathrm{LS}}^\pi)_2 \le -\frac14$ for \emph{every} $\pi$, whereas $Q^\pi(s_\star,a_1)-Q^\pi(s_\star,a_2)=+\frac14$ ($V^\pi(s_1)=\frac{7}{12}+\frac{1}{12}(1-\alpha_\star)$ is minimized at $\pi^\star(a_2\mid s_\star)=1$). Note that adding a constant $c$ to $C^k$ shifts $(w_{\mathrm{LS}}^\pi)_1$ and $(w_{\mathrm{LS}}^\pi)_2$ equally by $+c$ on orthogonal $\{e_1,e_2\}$, and regressing $C^k+\tilde b(\phi_3)$ onto $\phi_3\in\{e_1,e_2\}$ then subtracting $\tilde b(\phi_3)$ recovers $w_{\mathrm{LS}}^\pi$ identically. Finally, if $\sum_{k=1}^K\|\widehat w_k-w_{\mathrm{LS}}^{\pi^k}\|_2=o(K)$ and $\sum_{k=1}^K\ip{\pi^k(\cdot\mid s_\star)-e_{a_1}}{\ip{\phi(s_\star,\cdot)}{\widehat w_k}}\le o(K)$, then
\[
  \frac14\sum_{k=1}^K\pi^k(a_2\mid s_\star)
  \le \sum_{k=1}^K\pi^k(a_2\mid s_\star)\ip{\phi(s_\star,a_2)-\phi(s_\star,a_1)}{w_{\mathrm{LS}}^{\pi^k}}
  \le o(K) + \sqrt{2}\sum_{k=1}^K\|\widehat w_k-w_{\mathrm{LS}}^{\pi^k}\|_2 = o(K),
\]
so $\Reg = \frac{1}{12}\sum_{k=1}^K(1-\pi^k(a_2\mid s_\star)) = \frac{K}{12}-o(K)=\Omega(K)$.
\end{proof}